\documentclass[notitlepage,12pt]{jedm-template/jedm}
\usepackage[table]{xcolor}
\usepackage{hyperref}
\usepackage{graphicx}
\usepackage{amsmath}
\usepackage{amsfonts}
\usepackage{todonotes}
\usepackage{float}
\usepackage{comment}
\usepackage{longtable}
\usepackage{booktabs}
\usepackage{array,multirow}
\usepackage[utf8x]{inputenc} 
\usepackage{enumitem}
\usepackage[flushleft]{threeparttable}
\setlist[itemize]{nolistsep,itemsep=0.05cm}

\hypersetup{
  colorlinks   = true, 
  urlcolor     = blue, 
  linkcolor    = blue, 
  citecolor   = blue 
}

\newcommand\bs[1]{\boldsymbol{#1}}
\newcommand\surl[1]{\small{\url{#1}}}

\newcommand{\red}[1]{{#1}}

\begin{document}

\title{\red{Empirical Evaluation of Deep Learning\\ Models for Knowledge Tracing: \\Of Hyperparameters and Metrics on\\ Performance and Replicability}}


\date{}

\author{
     {\large Sami Sarsa}\\Aalto University\\sami.sarsa@aalto.fi \and 
     {\large Juho Leinonen}\\Aalto University\\juho.2.leinonen@aalto.fi \and 
     {\large Arto Hellas}\\Aalto University\\arto.hellas@aalto.fi 
}

\maketitle

\begin{abstract}

\red{
We review and evaluate a body of deep learning knowledge tracing (DLKT) models with openly available and widely-used data sets, and with a novel data set of students learning to program. The evaluated knowledge tracing models include Vanilla-DKT, two Long Short-Term Memory Deep Knowledge Tracing (LSTM-DKT) variants, two Dynamic Key-Value Memory Network (DKVMN) variants, and Self-Attentive Knowledge Tracing (SAKT). As baselines, we evaluate simple non-learning models, logistic regression and Bayesian Knowledge Tracing (BKT). 
To evaluate how different aspects of DLKT models influence model performance, we test input and output layer variations found in the compared models that are independent of the main architectures. We study maximum attempt count options, including filtering out long attempt sequences, that have been implicitly and explicitly used in prior studies. We contrast the observed performance variations against variations from non-model properties such as randomness and hardware. Performance of models is assessed using multiple metrics, whereby we also contrast the impact of the choice of metric on model performance.
The key contributions of this work are the following: Evidence that DLKT models generally outperform more traditional models, but not necessarily by much and not always; Evidence that even simple baselines with little to no predictive value may outperform DLKT models, especially in terms of accuracy -- highlighting importance of selecting proper baselines for comparison; Disambiguation of properties that lead to better performance in DLKT models including metric choice, input and output layer variations, common hyperparameters, random seeding and hardware; Discussion of issues in replicability when evaluating DLKT models, including discrepancies in prior reported results and methodology. Model implementations, evaluation code, and data are published as a part of this work.
}

{\parindent0pt
\textbf{Keywords:} \red{Knowledge Tracing, Deep Learning, Memory Networks, Attention-Based Models, Hyperparameter Optimization, Evaluation Metrics, Replicability}
}

\end{abstract}

\newpage

\section{Introduction \label{sec:introduction}}



Knowledge tracing (KT) is a student modeling task where knowledge or skill is estimated based on a trace of interactions with learning activities, such as course exercises. While new knowledge tracing models and modifications of existing ones are presented regularly~\cite{yudelson2013individualized,Piech2015,zhang2017dynamic,yeung2018addressing,Abdelrahman2019,Nakagawa:2019:GKT:3350546.3352513,pu2020deep}, it is not always clear what the actual factors that contribute to the performance of the proposed models are~\cite{wilson2016estimating,lipton2018troubling}. To what extent do the results depend on the proposed model algorithms and the used data, or are optimizations the actual key to the results?

When working with deep learning models -- machine learning models that use multiple processing layers for transforming an input to an output -- the resulting models are often not transparent and hence difficult to interpret. Substantial performance differences can be observed already due to the choice of hyperparameters that are used to train the models~\cite{bouthillier2021accounting}. Even the used hardware and software can influence the outcomes; for example, the documentation of PyTorch -- an open source machine learning library -- warns that ``\emph{Completely reproducible results are not guaranteed across PyTorch releases, individual commits, or different platforms. Furthermore, results may not be reproducible between CPU and GPU executions, even when using identical seeds}''\footnote{\surl{https://pytorch.org/docs/stable/notes/randomness.html}, accessed 2020-12-01}. 
\red{Deeper insight into the factors that contribute to differences in performance between models can be gained through reproduction, replication and ablation. Since the terms reproduction and replication are often used interchangeably, in this work by \emph{reproducibility} we denote using the same data and methodology -- including code -- by different researchers, and by \emph{replicability} we denote using the same methodology but re-implementing the work and with different data and researchers, as is done in previous work~\cite{patil2016statistical,stevens2017replicability}.  Although the need for replication and reproduction has been highlighted within the educational data mining domain~\cite{Ihantola2015,gardner2019modeling}, such studies are still scarce. }

\red{In our work, we replicate earlier studies, reimplementing models ourselves following descriptions available in the original articles. We review and evaluate a body of KT model algorithms with a focus on KT algorithms that utilize Deep Learning Knowledge Tracing (hereafter DLKT). As baselines, we use both non-learning models and Bayesian Knowledge Tracing and a recent Logistic Regression -based model. We triangulate factors that contribute to the performance of the models, including metric choice, input and output variations in model structure, non-model properties (e.g.\@ maximum input sequence lengths, random seeding and hardware) and commonly used hyperparameters. 
The evaluations are conducted with seven datasets -- six open ones and a novel dataset made openly available as a part of this study -- and seven metrics, with the purpose of identifying and discussing performance differences between the models, the datasets, and the metrics. Our research questions for this study are as follows.
}

\begin{itemize}

    \item[\bf{RQ1}] How do DLKT models compare to naive baselines and non deep-learning KT models?

    \item[\bf{RQ2}] How do DLKT models perform on the same and different datasets as originally evaluated with?

    \item[\bf{RQ3}] What is the impact of variations in architecture and hyperparameters on DLKT models' performance?

\end{itemize}

\section{Background}
\label{sec:related}


\subsection{Intelligent Tutoring Systems and Student Modeling}

Intelligent tutoring systems are software systems designed to provide personalized tutoring to students, supporting learning also in contexts where students have little to no access to human tutors~\cite{anderson1985intelligent,nwana1990intelligent,vanlehn2011relative,ma2014intelligent}. Classic intelligent tutoring systems have four main components; (1) a task environment, (2) a domain knowledge module, (3) a student model, and (4) a pedagogical module~\cite{corbett1997intelligent}.

Students work within the task environment, where their actions are evaluated through the domain knowledge module, which contains a structured representation of concepts, rules, and strategies that the student is expected to learn. The student model contains details of the students' actions within the task environment, an estimate of students' current knowledge, and possibly also other information about the students such as background and demographic information. Finally, based on the information from the domain knowledge module and the student model, the pedagogical module is used to decide on appropriate instructional interventions that will be fed to the task environment. These interventions may include, among other things, providing hints and suggestions for the current task, as well as guiding the students to appropriate tasks. 

Reviews that compare intelligent tutoring systems and other environments have suggested that intelligent tutoring systems outperform teacher-led large-group instruction, computer-based instruction, and the use of textbooks or workbooks~\cite{ma2014intelligent}, and that intelligent tutoring systems can be nearly as effective as human tutors~\cite{vanlehn2011relative,ma2014intelligent}. 

Students who use intelligent tutoring systems learn mainly through solving problems within the task environment. The problems are given based on the student model, which estimates what the student knows based on students' actions and other collected information~\cite{chrysafiadi2013student}. There are multiple ways for estimating students' knowledge, including model tracing and knowledge tracing. In model tracing, the domain model is used to interpret students' actions and follow the students step-by-step through the problem space, making adjustments if needed. In knowledge tracing, the system attempts to monitor students' changing knowledge as the student works on problems~\cite{corbett1994knowledge}, and, e.g., provides content related to a new topic when the student has mastered the current topic. In this article, our focus is on knowledge tracing.

\subsection{Knowledge Tracing}

Knowledge tracing~\cite{corbett1994knowledge} is an approach for student modeling in which students' actions within the task environment are used for estimating their current knowledge with regard to each individual knowledge component\footnote{While the original article~\cite{corbett1994knowledge} uses the term \emph{rule} for describing the concepts and objectives that the student is expected to learn, we refer here to the term as \emph{knowledge component} as it is more commonly used in contemporary research on the topic.} that the intelligent tutoring system is teaching. The estimates of current knowledge are typically probabilistic, where mastery of the current topic is assumed when the models posit at least a 95\% probability that the student has mastered the topic.

In practice, most if not all of the knowledge tracing algorithms are used to create regression models, which are used for estimating relationships between a dependent variable (in our context, often knowledge) and one or more independent variables (in our context, often variables describing students' behavior or outcomes from the task environment). One particular family of regression models that has been used in educational data mining is logistic regression models, which are often used for estimating the probability of a given class (e.g.\@ has mastered the topic / has not mastered the topic) based on a set of independent variables.

Here, we revisit models for knowledge tracing. The topics include probabilistic graphical models, factor analysis models, and deep learning models. 

\subsubsection{Probabilistic Graphical Models}

Probabilistic graphical models are probabilistic models that use graphs for presenting the conditional dependencies between the studied variables. The most prominent probabilistic graphical model for knowledge tracing is Bayesian Knowledge Tracing~\cite{corbett1994knowledge} (BKT\footnote{The original article uses the term Knowledge Tracing for the approach. Over time, the approach has been relabeled as Bayesian Knowledge Tracing by the scientific community.}). BKT assumes that knowledge is binary, i.e.\@ that a student has either mastered or not mastered the topic, and furthermore, that once a topic has been mastered, it cannot be unlearned or forgotten. Interactions with the task environment are related to individual knowledge components, and the interactions may produce an output, which is observable evidence that indicates whether the student's action was correct or incorrect. 

Whenever the system produces an output based on a student's actions, which represents either correct or incorrect knowledge, the student's knowledge of the knowledge component is updated. In BKT, the update is based on four parameters (described more formally in~\ref{subsec:bkt}), which define the probability that the student had already learned the knowledge component (prior learning), the probability that the student's action led to the student learning the knowledge component (transition), the probability that the student's interaction with the task environment produced a correct output by accident (guess), and the probability that the student's interaction with the task environment produced an incorrect output by accident (slip). In the original article, data from all students were used to estimate optimal values for the four parameters.

A range of BKT variations has been published since the original article. For example, researchers have added information on whether and how the student was helped during the process~\cite{chang2006does,lin2016intervention}, adjusted the probability of guessing or slipping using additional factors~\cite{d2008more}, adjusted the prior learning and transition parameters based on data from each individual student instead of all students~\cite{yudelson2013individualized}, and included information on the task difficulty to the process~\cite{pardos2011kt}.

In addition to the BKT variants, there exists other probabilistic graphical model -based approaches for KT. For example, Partially Observable Markov Decision Processes~\cite{rafferty2011faster} have been used to model student behavior. Similarly, methods such as (Dynamic) Bayesian Networks~\cite{pearl1985bayesian,pearl1988probabilistic,ghahramani1997learning} have been applied, e.g., for estimating student's knowledge~\cite{rowe2010modeling}, for extending BKT to allow multiple knowledge components~\cite{pardos2008composition}, for modeling dependencies between knowledge components~\cite{kaser2014beyond}, and for adjusting the BKT prior learning parameter based on additional factors~\cite{pardos2010modeling}.

\subsubsection{Factor Analysis Based Models}

Factor analysis is a method for reducing the amount of observed variables into a potentially lower number of unobserved variables called factors. The methods briefly described next, Item Response Theory, Learning Factors analysis, and Performance Factors Analysis all use existing data to estimate one or more latent (unobserved) variables.

Item Response Theory (IRT)~\cite{hambleton1985item}, from the field of psychometrics, is used for assessing student's ability and item difficulty based on students' responses to, e.g., questionnaire items. As IRT models can be used to produce estimates of item difficulty and student ability, they have been used for estimating the probability with which students will answer questions correctly~\cite{johns2006estimating}. However, the basic IRT model does not provide means for dynamically updating students' knowledge, and thus, the model has been incorporated with knowledge tracing, resulting in models that outperform BKT~\cite{khajah2014integrating,khajah2014integratingA,gonzalez2014general}. In addition, recent work on logistic regression models incorporating more aspects from students' behavior has been shown to also outperform DLKT models~\cite{gervet2020deep}.

Other variants purposed for the knowledge tracing task are Learning Factors Analysis~\cite{Cen2006} (LFA) and Performance Factors Analysis~\cite{PavlikJr2009} (PFA). LFA is comprised of three parts, which are (1) a parameter that quantifies student's ability, (2) difficulty of the knowledge components defined through a single parameter for each knowledge component, and (3) learning rate or benefit of frequency of prior practice for each knowledge component. Similar to the basic IRT model, LFA has downsides in how it can be used for adaptive environments as it does not account for the correct or incorrect outputs from the task environment, i.e.\@ correct and incorrect responses by the student. PFA accounts for this by reconfiguring the model by making it more sensitive to student's performance, and providing better means to adapt to the student's performance. Literature has provided evidence of PFA outperforming both LFA and BKT~\cite{PavlikJr2009}, although there also exists evidence on BKT performing on par with PFA~\cite{gong2010comparing}. In addition, it has been observed that, in particular tasks, PFA performs almost on par with a DLKT model~\cite{xiong2016going}, which we discuss next.

\subsubsection{Deep Learning Models for Knowledge Tracing}
\label{subsec:related-dlkt-models}

Deep learning models are a set of machine learning models that are based on (artificial) neural networks. Neural networks are networks of ``neurons'' organized in layers that are used as information processing systems for e.g.\@ classification tasks. Neurons in the neural networks are mathematical functions that take a fixed sized set, i.e.\@ a vector, of inputs, have a weight (learnable parameter) for each input, and produce an output as a weighted sum of the inputs. The output is then processed using an activation function (e.g.\@ a sigmoid function).

The first layer of a neural network is its input, i.e.\@ the data that is fed to the network. A single layer may contain an arbitrary amount of neurons and a neural network can consist of one or more layers of neurons without an upper bound. Complex neural networks, i.e.\@, ones that are deeply layered, are often called deep learning models or deep networks. In contrast, neural networks with a relatively simple layer structure with few layers are sometimes referred to as shallow networks \cite{bianchi2014shallow}. The most common type of layer in a neural network, one that is used commonly in the models explored in this paper, is the fully connected layer (FCL), wherein each neuron (and weight) of the layer is connected to each of its inputs.

Training a neural network consists of providing training data to the network and adjusting the weights of the neurons so that the overall network learns to produce a desired output. During learning, the performance of the model is estimated using an error function that compares the output of the model with the expected output, i.e.\@ true outcome. Neural networks often require extensive amounts of training data in order to produce generalizable results. This is especially true when the training data is complex, such as when it contains sequential relations, e.g.\@ temporality or word order. Thus, adjustments to neural networks have been made to better conform to certain kinds of data relations. The Recurrent Neural Network (RNN)~\cite{hochreiter1997long} is a result of one such adjustment that has been widely adopted. In an RNN, connections between neurons are constructed as a directed graph that can represent a temporal sequence, and where the output of each neuron can be fed back to the network (similar to recursion). 
RNNs have been shown to work well in various domains, including natural language processing~\cite{mikolov2011ext_rnn}, stock market prediction~\cite{kamijo1990stockrnn,saad1998comparative}, and protein sequence prediction~\cite{saha2006prediction}

RNNs were first used for the Knowledge Tracing task in a study by Piech et al.~\cite{Piech2015}.
They showed that ``Deep Knowledge Tracing'' (DKT) outperformed BKT, despite modeling all the skills jointly instead of building a separate model for each skill.
Since then, DKT has been shown to perform well in a range of studies~\cite{wilson2016estimating,lin2017comparisons,mao2018deep,montero2018does,ding2019deep}. This has also led to a range of DLKT-based approaches that have been proposed and evaluated for the Knowledge Tracing task.
\red{
Their contributions include both proposing new model algorithms by introducing techniques from the broader machine learning domain and also how to leverage different sorts of input into the models.
For example,
Zhang et al.~\cite{zhang2017dynamic} introduce DKVMN, a memory network that utilizes next skill input when predicting next attempt correctness, that is not present in the original DKT model, and Abderahman et al.~\cite{Abdelrahman2019} further propose use of memory network architecture in tandem with LSTMs in their SKVMN model.
As a more input oriented example, The models by Su et al.~\cite{su2018exercise} (EERNN) and Liu et al.~\cite{liu2019ekt} (EKT) utilize textual content of exercises.
Pandey et al.~\cite{p2019selfattentive} started a line of research on self-attentive models known as transformers for knowledge tracing with their SAKT model.
Further attention-based models include RKT~\cite{pandey2020rkt}, context-aware AKT~\cite{ghosh2020context} that fuses DLKT with IRT,
SAINT~\cite{choi2020towards}
with an encoder-decoder architecture,
time-aware encoder-decoder by Pu et al.~\cite{pu2020deep},
and a time-aware version of SAINT called SAINT+~\cite{shin2021saint+}.
Another line of development has concentrated on graph neural networks with models such as  GKT~\cite{Nakagawa:2019:GKT:3350546.3352513} that models learning of skills as a network and  JKT~\cite{song2021jkt} which takes a step further by model by modeling both skill and exercise relations.
Many of these models have been introduced since this study is started, which highlights the rapid development and popularity of the field, but also the need for high quality methodology and replication work for fair comparisons and disambiguation of causes for performance differences.
}


\subsection{Replicating Work from Others}

Discussions on the importance of replicating work from others have stemmed from observations that published results do not always hold up under scrutiny~\cite{Ioannidis2005,Moonesinghe2007}.
Researchers from various disciplines have voiced out their concerns about the state of replicability in their fields, which include, for example, biology~\cite{begley2012}, computing education~\cite{Ahadi:2016}, educational data mining~\cite{gardner2019modeling}, health care~\cite{Ioannidis2005-2}, political science~\cite{golden1995} and computational science~\cite{peng2011}.
This concern was highlighted also in a multi-disciplinary survey of over 1500 researchers, where about three quarters of the respondents considered that they could only trust \emph{at least half} of the papers in their field~\cite{Baker2016}.
Whether published results hold may depend also on the field of science; for example, one study from psychology replicated 100 experimental and correlational studies from three journals and found that only about one-third to one-half of the original findings could be replicated~\cite{open2015}, raising discussion about how to conduct replication studies~\cite{gilbert2016,anderson2016}.

Despite the need for replication studies, studies that attempt to replicate earlier findings remain relatively rare~\cite{muma1993,makel2012}. This rarity has been linked with valuing innovation and original research~\cite{Mackey2012,asendorpf2013,Ahadi:2016} and the concern that replications may not be publishable~\cite{fanelli2011,spellman2012,Ahadi:2016}. Both of these issues may steer researchers away from conducting and publishing replication studies. In the survey of over 1500 researchers, where relative lack of trust towards earlier results was highlighted as one of the issues, over half of the surveyed researchers had been unable to \red{replicate} an experiment from others~\cite{Baker2016}, suggesting that there is a dire need of sufficient details for replication as well as a need to maintain quality. Indeed, calls for more detailed reporting of experimental designs, requiring publication of data and research materials, and rewarding replication attempts have been made~\cite{asendorpf2013,ioannidis2014}.

With replication studies, one could -- for example -- identify issues with how methodological details are presented and consequently also highlight the importance of replication to the community~\cite{Ihantola2015,gardner2019modeling}. By reimplementing algorithms as a part of the replication studies, one may also reveal bugs in existing research code, that could lead researchers into faulty conclusions~\cite{bhandari2019characterization}. Although DLKT models often outperform older models, the relative performance of evaluated models may be influenced, among other factors, by the context, the task, the data, the hyperparameter tuning and the chosen metrics~\cite{xiong2016going,lalwani2017few,mao2018deep,ding2019deep}. The relative impact of these factors could be revealed through replication studies.

\subsection{\red{Effects of Metrics on Reported Performance}}

\red{
Metrics for evaluating model performance have been extensively studied and developed, and different metrics are used to evaluate different aspects of model performance. Relying on any single metric alone is prone to provide misleading information on model performance~\cite{national2005thinking}.
Plenty of discussion has concentrated on the goodness of accuracy and (ROC-)AUC (Receiving Operator Characteristic Area Under Curve) as metrics,
where some scrutinize the former~\cite{ling2003auc,halimu2019empirical} as misleading and others scrutinize the latter~\cite{jeni2013facing,muschelli2020roc} as potentially masking bad performance.
Both of these are popular metrics in the Knowledge Tracing field as well as other 
domains~\cite{pahikkala2009matrix,huang2019patient}.}

\red{Within the field of educational data mining, some studies have been conducted on the effect of metric choice on model performance, although on different metrics and not so much on accuracy and AUC.
As an example, Pelanek at al. \cite{pelanek2015metrics} compare MAE (Mean Average Error, RMSE (Root Mean Squared Error, LL (Log-likelihood aka binary cross-entropy and AUC on BKT, PFA and Elo rating system adaptation~\cite{pelanek2014application} models.
A rather recent study by Effenberger et al. \cite{effenberger2020impact} compare the highly similar metrics MAE and RMSE for multiple models for student modeling, including the mentioned Elo adaption, an IRT model and Random Forest, and show that metric choice affects model performance and can even affect model ranking. They also highlight other issues in methodological choices such as filtering choices of data in preprocessing.}

\subsection{Recent Comparisons of Deep Learning Models for Knowledge Tracing}

There are signs that the popularity of replication studies in educational data mining is on the rise. This has been especially visible within the Educational Data Mining community, where recent calls for papers at e.g.\@ the Educational Data Mining conference have included reproducibility of research as a topic of interest\footnote{See e.g.\@ \surl{https://educationaldatamining.org/edm2021/call-for-papers/}, accessed 2020-10-15}.

A recent article by Gervet et al.~\cite{gervet2020deep} compared different models that have been used for KT. Their evaluated models included (1) different DLKT models, i.e.\@ DKT~\cite{Piech2015}, SAKT~\cite{p2019selfattentive} and a feedforward network; (2) regression models, i.e.\@ IRT, PFA~\cite{PavlikJr2009}, DAS3H~\cite{choffin2019das3h} and a logistic regression model (LR); and (3) a variation of BKT called BKT+~\cite{khajah2016deep} that adds individualization, forgetting and discovery of knowledge components. Additionally, they conducted ablation studies to examine which features of different models could explain differences in their performance. The metrics that were used to compare the models were AUC score and RMSE, and the study used nine different datasets. The results showed that DKT and logistic regression performed the best, with DKT being the best model in five datasets and LR being the best in four. SAKT underperformed DKT in all of the datasets, which is contrary to prior results by Pandey and Karypis~\cite{p2019selfattentive}, where SAKT performed significantly better compared to DKT. The results by Gervet et al. also suggest that dataset size affects model performance: LR worked better for smaller datasets and DKT performed better for larger ones. Additionally, DKT seemed to reach peak performance faster when compared to LR, for which the performance continues to improve over a longer time.

In a similar study, Pandey et al.~\cite{pandey2021empirical} compared DLKT models. They included DKT~\cite{Piech2015}, DKVMN~\cite{zhang2017dynamic}, SAKT~\cite{p2019selfattentive} and RKT~\cite{pandey2020rkt} in their comparison. The models were compared using the EdNet dataset~\cite{choi2020ednet} by comparing accuracies and AUCs. The results of Pandey et al. suggest that the self-attention based models (RKT and SAKT) perform better compared to DKT and DKVMN.

Similar to Pandey et al.~\cite{pandey2021empirical}, Mandalapu et al.~\cite{mandalapu2021we} used the EdNet dataset~\cite{choi2020ednet} to compare different knowledge tracing models. They used two versions of the EdNet dataset: the full data with 600,000 students and a pruned dataset with 50,000 students. They evaluated a baseline model (prediction based on probability of correctness in training set), a logistic regression model, DKT~\cite{Piech2015}, and SAKT~\cite{p2019selfattentive}. The models were compared with the AUC metric. Based on the results of Mandalapu et al.~\cite{mandalapu2021we}, the logistic regression model very slightly outperformed DKT and SAKT (0.77 AUC for LR vs 0.76 AUC for DKT and SAKT) in the smaller 50,000 student dataset. In the full dataset, SAKT outperformed DKT (0.76 vs 0.72 AUC), while the logistic regression model could not be trained with the full dataset. Contrary to Gervet et al.~\cite{gervet2020deep}, Mandalapu et al.~\cite{mandalapu2021we} found that DKT's performance was worse with a larger dataset. Mandalapu et al. hypothesized that SAKT benefits from a larger dataset size. Interestingly, out of the two models present in both Pandey et al.'s and Mandalapu et al.'s studies -- SAKT and DKT --, SAKT performed approximately similarly in both Mandalapu et al.'s and Pandey et al.'s studies achieving around 0.76 AUC, while there was a difference in the performance of DKT, for which Mandalapu et al. got an AUC score of 0.72 and Pandey et al. an AUC score of around 0.742. Since the model and the data are the same, the difference of around 0.022 AUC is likely due to differences in hyperparameter tuning or the used hardware; in the present study, we also explore the effects of hyperparameter tuning and hardware.

What is evident from all the comparisons is that the results of comparing different models are affected by many factors. Interestingly, the authors of prior comparison studies mostly relied on AUC scores to evaluate the models: AUC is the only performance metric present in all three of~\cite{gervet2020deep},~\cite{pandey2021empirical}, and~\cite{mandalapu2021we}, despite the possibility of models performing differently with different metrics~\cite{caruana2004data,national2005thinking,gunawardana2009survey,sanyal2020feature}. In addition, the differences between the best few models in the three comparison studies are not particularly large: the AUC scores between the best two models, for example, are typically within 0.02 AUC of each other.

\section{Evaluated Models}
\label{sec:models}

\subsection{\red{Overview}}

We inspect three naive models, two commonly used baseline models, and three
deep learning knowledge tracing (DLKT) models. We also include three variants of the DLKT models that we report as separate models; one of these is a novel variant created for the purposes of this study.
The models and their respective abbreviations are summarized in Table \ref{tbl:models}.

The models included in this study were chosen based on recentness \red{and gained traction} at the beginning of this work in late 2019, where each selected model reported state-of-the-art results and possibly also conflicting results for the previous state-of-the-art; the included DLKT models have all been used also in more recent works that build on existing knowledge tracing models \cite{liu2019exploiting,trifa2019knowledge,oya2021lstm}.
The naive models were selected to provide a more comprehensive analysis of how different results can be achieved even with simpler methods as well as to provide validity of the machine learning models' usefulness, while BKT and a logistic regression model were chosen as baselines to include non-DLKT models for comparison.
We note that modern variants of BKT may provide significantly better results than the baseline we use\footnote{\surl{https://github.com/myudelson/hmm-scalable}, accessed 2020-03-01}, which includes an individualized BKT~\cite{yudelson2013individualized} variant.

For our logistic regression baseline we use the Best-LR\footnote{\surl{https://github.com/theophilee/learner-performance-prediction}, accessed 2021-05-27} (GLR) model that builds on PFA, IRT and DAS3H, and was the best performing logistic regression model in the recent study by Gervet et al.~\cite{gervet2020deep}.

\begin{table}[ht!]
    \caption{Summary of evaluated models}
    \small
	\label{tbl:models}
	\begin{center}
		\begin{tabular}{c | m{5.3cm} m{2.5cm} m{6.7cm} }
		    \toprule
			& Model & Shorthand & Details \\
			\midrule
			\parbox[t]{2mm}{\multirow{3}{*}[-0.5ex]{\rotatebox[origin=c]{90}{Naive}}} 
			& Mean prediction & Mean & A simple statistic\\[0.4ex]
			& Next as previous & NaP & A simple baseline model\\[0.4ex]
			& Next as previous N mean & NaPNM & A slightly less simple baseline model\\[0.4ex]
			\midrule
			\parbox[t]{2mm}{\multirow{2}{*}[1.2ex]{\rotatebox[origin=c]{90}{Non-DLKT}}} 
			& Bayesian Knowledge Tracing & BKT & \red{Commonly used as a baseline}, predecessor \red{to} DLKT models  \cite{yudelson2013individualized} \\[0.4ex]
			& Gervet et al. Logistic Regression & GLR & Logistic regression model with best input feature combination in \cite{gervet2020deep}\\[1.1ex]
			\midrule
			\parbox[t]{2mm}{\multirow{6}{*}[2.0ex]{\rotatebox[origin=b]{90}{Deep Learning Knowledge Tracing (DLKT)}}} 
			& Long Short Term Memory (Recurrent Neural Network) Deep Knowledge Tracing & LSTM-DKT & First DLKT model~\cite{Piech2015}\\[0.4ex]
			& Vanilla (Recurrent Neural Network) Deep Knowledge Tracing & Vanilla-DKT & First DLKT model along with LSTM-DKT~\cite{Piech2015}\\[0.4ex]
			& Dynamic Key-Value Memory Network (MXNet implementation) & DKVMN & First DLKT model \red{with} separate next attempt skills as input~\cite{zhang2017dynamic}\\[0.4ex]
			& Dynamic Key-Value Memory Network (as depicted in its respective paper) & DKVMN-Paper & Same as above \\[0.4ex]
			& Self Attentive Neural Network & SAKT & A DLKT model based on Transformer neural network ~\cite{p2019selfattentive}\\[0.4ex]
			& LSTM-DKT with next skill input & LSTM-DKT-S+ & LSTM-DKT \red{variation} with added skill input as in DKVMN and SAKT, presented in this work \\
			\bottomrule
		\end{tabular}
	\end{center}
\end{table}

All the evaluated models, outlined in Table~\ref{tbl:models}, accept inputs as sequences of exercise attempts per student and output the probability of next attempt correctness for each attempt in the input. An attempt consists of a skill id $s_t$ and correctness $c_t$ at time step $t$, where time is an increasing integer sequence. The common variables used in the subsequent descriptions of the models are summarized in Table \ref{tbl:vars}.

\begin{table}[ht!]
    \caption{Common variables used in the descriptions of the evaluated models}
    \small
	\label{tbl:vars}
	\begin{center}
		\begin{tabular}{ m{3.5cm} m{11.5cm} }
		    \toprule
			Variable & Description \\
			\midrule
			$T\in\mathbb{N}$ & Number of attempts in an attempt sequence\\[0.4ex]
			$t\in\{1..T\}$ & Time step of an attempt sequence\\[0.4ex]
			$S\in\mathbb{N}$ & Number of distinct skills \\[0.4ex]
	        $\bs{s}\in\{1..S\}^{T}$ & Set of skill identifiers in an attempt sequence\\[0.4ex]
	        $\bs{c}\in\{0,1\}^{T}$ & Set of correctness values in an attempt sequence\\[0.4ex]
			$s_t\in\{1..S\}$ & Skill identifier at time step (or attempt number) $t$\\[0.4ex]
			$c_t\in\{0,1\}$ & Attempt correctness at time step $t$ (1 indicates correct and 0 incorrect)\\[0.4ex]
			$y_t\in[0,1]$ & Model output, i.e.\@ prediction of attempt correctness produced at time step $t$ ($y_t$ is an estimate of $c_{t+1}$ produced by the model)\\[0.4ex]
			$\sigma(x)$ & Standard logistic function $\sigma(x) = \frac{1}{1 + e^{-x}}$ \\[0.4ex]
			$\tanh(x)$ & Hyperbolic tangent $\tanh(x) = \frac{e^{2x} - 1}{e^{2x} + 1}$\\
			$\bs{x}\odot\bs{y}$ & Element-wise product between $\bs{x}$ and $\bs{y}$. \\
			\bottomrule
		\end{tabular}
	\end{center}
\end{table}

Next, we explore the architectures of the studied models with some mathematical detail to portray a concrete picture of the differences between the models. We mainly follow the mathematical notation from the original articles. Some differences to the notation are introduced for instance to maintain consistency of variable definitions across the model descriptions in this work; the model descriptions also include a less mathematical overview of how the models work as well as illustrative figures of the DLKT model architectures. The used notation assumes basic understanding of linear algebra and neural networks.
 
\subsection{Naive Baselines \label{subsec:naive-baselines}}

\subsubsection{Mean}
Mean is our simplest model.
It is also the only one of the naive models that is computed using training data. 
The mean model takes the mean of the correctness values in training data and uses the computed mean as prediction output.
\red{To keep the model as simple as possible, the mean is computed over all of the training data values, i.e. not per student or skill.}

For metrics that operate on binary prediction values (correct or incorrect) instead of percentages, the mean predictor model is equivalent to Majority Vote aka ZeroR (Zero Rate) classifier that predicts all values as the most common label in given data. \red{This happens because the mean prediction is rounded in order to compute the metric score. For metrics that are computed from continuous values (e.g. RMSE), the models are not equivalent. }

While the mean model can be considered to have no predictive power, for that reason specifically, it is good for evaluating whether other models are useful in practice or not~\cite{devasena2011effectiveness}.

\subsubsection{Next as Previous}
Another naive model we use is predicting the correctness of a student's answer at time $t + 1$ to be the same as the student's previous answer correctness $c_t$. Formally, NaP$(t+1)=c_t$.
Similarly to mean prediction, NaP has minimal predictive power and thus serves as an additional validity check for more sophisticated models.

\subsubsection{Next as Previous N's Mean}

For our final naive baseline, we extend the Next as Previous model to compute the probability of a student answering an exercise correct as the mean of N (or $t$ if $t<N$) previous correctnesses so that NaPNM$(t+1)=\text{mean}(c_t,..,c_{\max(t-N,1)})$.
We report results for NaP3M and NaP9M.

\subsection{Bayesian Knowledge Tracing \label{subsec:bkt}}
Bayesian Knowledge Tracing (BKT) \cite{corbett1994knowledge}
models student knowledge as a latent, i.e.\@ hidden, variable. It is a special case of the Hidden Markov Model, which is a statistical model containing unobservable variable states where the probability of a state is dependent only of the immediately previous state and no other states. In BKT, student knowledge of a skill $s$ is represented as a binary variable $m_s$ indicating skill mastery (either \textit{true} or \textit{false}). 

The latent student state is determined by four parameters.

The first is prior learning $P(L_0)$, which is the probability that a student has learned a skill before attempting to apply it.
Transition $P(T)$ is the probability of transition from \textit{not mastered} to \textit{mastered} state for a student's knowledge of a skill after an attempt to apply it.
Guess $P(G)$ is the probability of applying a skill correctly by coincidence.
Slip $P(C)$ is the probability of applying a skill incorrectly by coincidence.
BKT models each skill separately and thus sets these parameters individually for each skill. The inner workings of the BKT model is explained in more detail in e.g.\@ \cite{yudelson2013individualized}.

\subsection{Gervet et al. Best-LR \label{subsec:best-lr}}
The Best-LR (GLR) model is a logistic regression model introduced in a recent Deep Learning for Knowledge Tracing study~\cite{gervet2020deep}; the model was the best performing logistic regression model within the study.

The model leverages knowledge tracing input features that are used in previous logistic regression models for knowledge tracing with an additional feature of total success and error counts for previous student attempts.
The authors describe the model as ``DAS3H~\cite{choffin2019das3h}  without  time-window  features  but  augmented  with  total count features, or equivalently PFA~\cite{PavlikJr2009} with rescaled count features, augmented with total counts features and IRT student ability and question difficulty parameters''.
Mathematical details of the GLR model can be found in the original work.

\subsection{Deep Learning Knowledge Tracing}

\subsubsection{RNN-DKT\label{subsec:vanilla-dkt}}

The first DLKT model DKT (Deep Knowledge Tracing) ~\cite{Piech2015} is an RNN, which is a neural network that uses an internal memory, often referred to as kernel, to process sequential inputs such as student attempts at an exercise. 
An RNN for DLKT receives the encoded sequences of previous attempts and skills as input and produces a probability of next attempt correctness as output. The probability of next attempt correctness is output for each part of the input sequence.

The original DKT model was tested with both a ``vanilla'' recurrent kernel, i.e.\@, a simple fully connected layer (FCL) with a hyperbolic tangent ($\tanh(x) = \frac{e^{2x} - 1}{e^{2x} + 1}$) activation function, and a more complex Long Short Term Memory (LSTM) \cite{hochreiter1997long} recurrent kernel.
The main advantage of LSTM over vanilla RNN is that LSTM is specifically designed to not lose information across the recurrence over time for long input sequences by using a gate structure to control computation flow over the recurrent network.
In our experiments, we refer to the DKT models that incorporate an RNN architecture, i.e.
RNN-DKTs, by their kernels, namely Vanilla-DKT and LSTM-DKT.

Mathematically speaking, an RNN kernel is an update equation $f$ for its hidden state $\bs{h}$ at timestep $t$ that takes the current input $\bs{x}_t$ and previous hidden state as parameters, i.e., $\bs{h}_t = f(\bs{x}_t,\bs{h}_{t-1})$. In our context, $\bs{x}_t$ is an embedded vector representation of skill $s_t$ and correctness $c_t$ for timestep $t$.

For our vanilla RNN, we used the SimpleRNN\footnote{\surl{https://keras.io/api/layers/recurrent_layers/simple_rnn/}, accessed 2020-01-15} Keras implementation which consists of multiplying inputs and weights, summing the result to weighted previous outputs, and applying an activation function to the resulting sum.
We use tanh as the activation function and thus, our vanilla RNN kernel is $\bs{h}_t = \tanh(\bs{h}_{t-1}\bs{W}^{h} + \bs{x}_t\bs{W}^{x} + \bs{b})$, where $\bs{W}^x$ and $\bs{W}^h$ are trainable matrices of input and hidden state weights and $\bs{b}$ is a bias vector.
The RNN kernel output is equal to $\bs{h}_t$ for the vanilla RNN.

For the sake of conciseness, we hereafter use $\sigma$ to denote the standard logistic function $\text{f}(x) = \frac{1}{1 + e^{-x}}$ that converts real values into probabilities, i.e.\@ values between 0 and 1. Also, we use $\odot$ to denote element-wise product.
For the LSTM kernel, which has multiple variations \cite{hochreiter1997long,gers2001lstm,graves2013generating}, we use the Keras kernel\footnote{\surl{https://keras.io/api/layers/recurrent_layers/lstm/}, accessed 2020-01-15} which is based on the Hochreiter and Schmidhuber \cite{hochreiter1997lstm} variant. The chosen LSTM kernel is a complex layer that comprises an input gate, a forget gate, an output gate and a memory cell.
The variables $\bs{W}^g$ and $\bs{U}^g$ in the following equations represent trainable input and hidden state weight matrices that are specific for computing $g$, where $g$ represents which part of the kernel the weights relate to. The LSTM kernel parts are defined as follows:
\begin{itemize}
    \item[] input gate: $\bs{i}_t = \sigma(\bs{x}_t\bs{W}^i + \bs{h}_{t-1}\bs{U}^i + \bs{b}^i)$
    \item[] forget gate: $\bs{f}_t = \sigma(\bs{x}_t\bs{W}^f + \bs{h}_{t-1}\bs{U}^f + \bs{b}^f)$
    \item[] output gate: $\bs{o}_t = \tanh(\bs{x}_t\bs{W}^o + \bs{h}_{t-1}\bs{U}^o + \bs{b}^o)$ 
    \item[] memory cell: $\bs{m}_t = \bs{f}_t\odot\bs{m}_{t-1}  + \bs{i}_t \odot\sigma(\bs{x}\bs{W}^m + \bs{h}_{t-1}\bs{U}^m + \bs{b}^m)$
    \item[] output vector:  $\bs{h}_t = \bs{o}_t\odot\tanh(\bs{m}_t)$
\end{itemize}

\begin{figure}
    \centering
    \includegraphics[scale=.17]
    {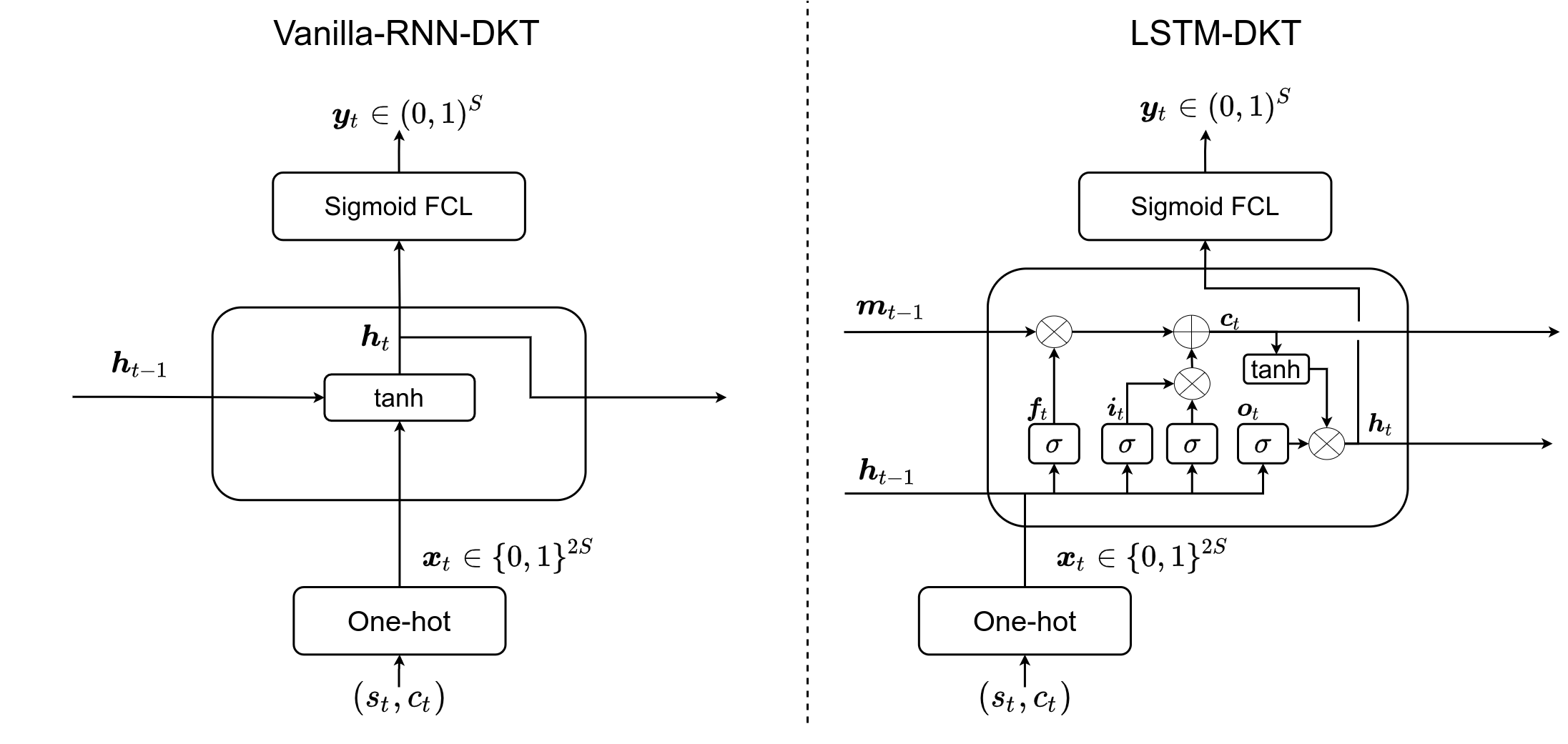}
    \caption{Architectures of Vanilla-\red{DKT} and LSTM-DKT models.}
    \label{fig:rnn-dkts}
\end{figure}

The original RNN-DKT models consist of the recurrent kernel and an output layer that produces the outputs $\bs{y}_t = \sigma(\bs{o}_t\bs{W}^y + \bs{b}^y)$ of the neural net. The RNN-DKT architectures for the two kernels (Vanilla-RNN and LSTM) are depicted in Figure \ref{fig:rnn-dkts}.

The RNN-DKT models predict the probabilities of a student having mastered skills given the student's exercise attempt sequence. For each attempt at time $t$, the attempt sequence contains the resulting correctness of the attempt $c_t\in\{0,1\}$, where $t\in\{1..T\}$, and a skill id $s_t\in\{1..S\}$, where $S$ is the number of distinct skill tags.
Given that $T$ denotes the number of exercise attempts for a student, the neural network takes as input a sequence of student's exercise attempt tuples $(s_t, c_t)$  that are combined into integers $2s_t+c_t$ and then converted into embedded one-hot vectors $\bs{x}_t\in\mathbb{R}^{2S}$. The $T\times2S$ sized sequence is then fed to a standard LSTM recurrent neural network with $T$ outputs $\bs{y}_t\in(0,1)^S$, one per each time step $t$. In the resulting output vector, each vector dimension represents the probability of a student knowing the corresponding skill.
The model is trained by considering only the skill tag at time $t + 1$ with a binary cross entropy loss $L(\bs{y}^T\cdot\delta(s_{t+1}), c_{t+1})$ per student attempt where $L$ denotes the binary cross entropy function and $\delta$ denotes one-hotting. Note that $T$ in the loss function denotes transpose and not the number of attempts. Overall, this means that the model can improve itself for the upcoming skill based on previous skills for each time step.
A successful modification~\cite{yeung2018addressing}, coined DKT+, has been created to improve the current timestep's skill output in addition to the next skill output. This modification has been left out from the current study however, \red{since} the differences in model performance are not reportedly significant. 

\subsubsection{DKVMN \label{subsec:dkvmn}}

Dynamic Key-Value Memory Network (DKVMN) \cite{zhang2017dynamic} is an RNN similar to the previously discussed DLKT models LSTM-DKT and Vanilla-DKT, however, its architecture differs significantly from these two models. DKVMN is built to leverage the additional information of next attempt skill tag as an input in addition to the current skill tag and the current attempt correctness.
The architecture is heavily inspired by 
RNNs that use a read-write memory recurrent kernel that incorporates an attention mechanism \cite{graves2014neural}.
Namely, these RNNs are
the Memory Network~\cite{weston2014memory,sukhbaatar2015end} and the Key-Value Memory Network \cite{miller2016key} models, which were originally designed to solve natural language processing problems. An additional difference is that the DKVMN, as presented originally, replaces the one-hot embedding of inputs used in LSTM-DKT with embedding matrices \cite{gal2016theoretically} to transform the inputs into dense vector representations.

In brief, the DKVMN consists of a recurrent kernel with read and write processes where read denotes the process of generating values for output of the kernel, and the write process involves updating the state of the recurrent kernel. The key inputs are used to compute attention weights via a key memory layer, whereas value inputs are used in the write process. The computed attention weights are applied to both read and write processes of the recurrent kernel. The outputs of the read process of the kernel are then fed to a feed-forward network that produces the model outputs, i.e., the probabilities of exercise correctness.

The DKVMN model involves two embedding layers to process the skill and correctness inputs into fixed-size dense vectors.
A key embedding layer for the next skill id in the input sequence, and a value embedding layer for concatenated current skill and correctness.
We denote the $i$th row of a weight matrix $\bs{A}$ as $\bs{A}(i)$, key embedding size as $K$ and value embedding size as $V$.
Using a key embedding matrix $\bs{E}^k\in{\mathbb{R}^{S\times K}}$, key embeddings $\bs{k}_t$ are obtained from the matrix row that corresponds to the next skill in sequence, i.e.\@ $\bs{k}_t = \bs{E}^k(s_{t+1})$.
Similarly, value embeddings $\bs{v}_t$ are obtained from a 
value embedding matrix $\bs{E}^v\in{\mathbb{R}^{2S\times V}}$
so that $\bs{v}_t = \bs{E}^v(2s_t + c_t)$.

We describe the DKVMN recurrent kernel using two distinct steps. The first step is the read process, where an input of a single timestep is read and an output is produced. The second step is the write process, in which the DKVMN recurrent kernel (value memory) is updated for processing the next timestep. DKVMN has two separate memory layers, one for values and one for keys, both with the same memory size $M$.
The key memory $\bs{M}^k\in{\mathbb{R}^{K\times M}}$, which is non-recurrent, is used to compute
attention weights $\bs{w}_t$ by multiplying each key memory row with the key embedding so that $\bs{w}_t = \left(\text{softmax}(\bs{k}_t\bs{M}^k(1)), .., \text{softmax}(\bs{k}_t\bs{M}^k(M))\right) \in\mathbb{R}^M$, where the softmax function is
$\text{softmax}(\bs{x})_i = \frac{e^{x_i}}{\sum_{j=1}^Ne^{x_j}}$ for all values $i\in\{1..N\}$ in vector $\bs{x}$.

A read value vector $\bs{r}_t$ is computed using the attention weights and current value memory $\bs{M}_t^v\in{\mathbb{R}^{M\times V}}$ as $\sum_{i=1}^M w_t(i)\bs{M}^v(i) \in\mathbb{R}^V$, i.e.\@ the read value is a weighted sum of value memory rows, where key memory provides the weights.
Note that for the first input, the read value is not dependent on correctnesses, but only the skill id $s_{t+1}$ and the initial weights of the value memory.
The read value is then concatenated with the key embeddings and the concatenated vector is fed to a dense, i.e.\@, fully connected layer (FCL) of size $S^\prime$ 
with weights $\bs{W}^s\in\mathbb{R}^{K + V\times S^\prime}$, which is called the summary layer as it is meant to summarize student knowledge state at time $t$. Then, the summary layer output, i.e., the summary vector $\bs{s}_t=\tanh($concat$(\bs{k}_t, \bs{r}_t) \bs{W}^s + \bs{b}^s)\in(-1, 1)^{K + V}$ is fed to a binary, i.e.\@ one-neuron, sigmoid activated output layer with weights $\bs{W}^y\in\mathbb{R}^{S^\prime\times 1}$ resulting in a probability prediction
$y_t=$ $\sigma(\bs{s}_t \bs{W}^y + \bs{b}^y)\in(0,1)$ for exercise correctness at time $t+1$.

The write process of DKVMN involves value memory, attention weights, an addition layer and an erase layer. The addition layer is a FCL with weights $\bs{W}^a\in\mathbb{R}^{V\times V}$ with tanh activation. The erase layer is a FCL with sigmoid activation and weights $\bs{W}^e\in\mathbb{R}^{V\times V}$.
The addition and erase layers both take value embeddings $v_t$ as input and respectively produce an 
addition vector $\bs{a}_t=\text{tanh}(\bs{v}_t\bs{W}^a + \bs{b}^a)\in(-1,1)^V$ and an
erase vector $\bs{e}_t=\sigma(\bs{v}_t\bs{W}^e + \bs{b}^e)\in(0,1)^V$ as output. Each value memory row is then updated by multiplying the value memory row with complemented attention weighted erase vector values $e_t$ summed with the addition vector $a_t$, i.e.\@
the updated value memory $\bs{M}_{t+1}^v(i) = \bs{a}_t + \bs{M}_t^v(i)(1-\bs{w}_t(i)\bs{e}_t)$.

\begin{figure}
    \centering
    \includegraphics[scale=.15]
    {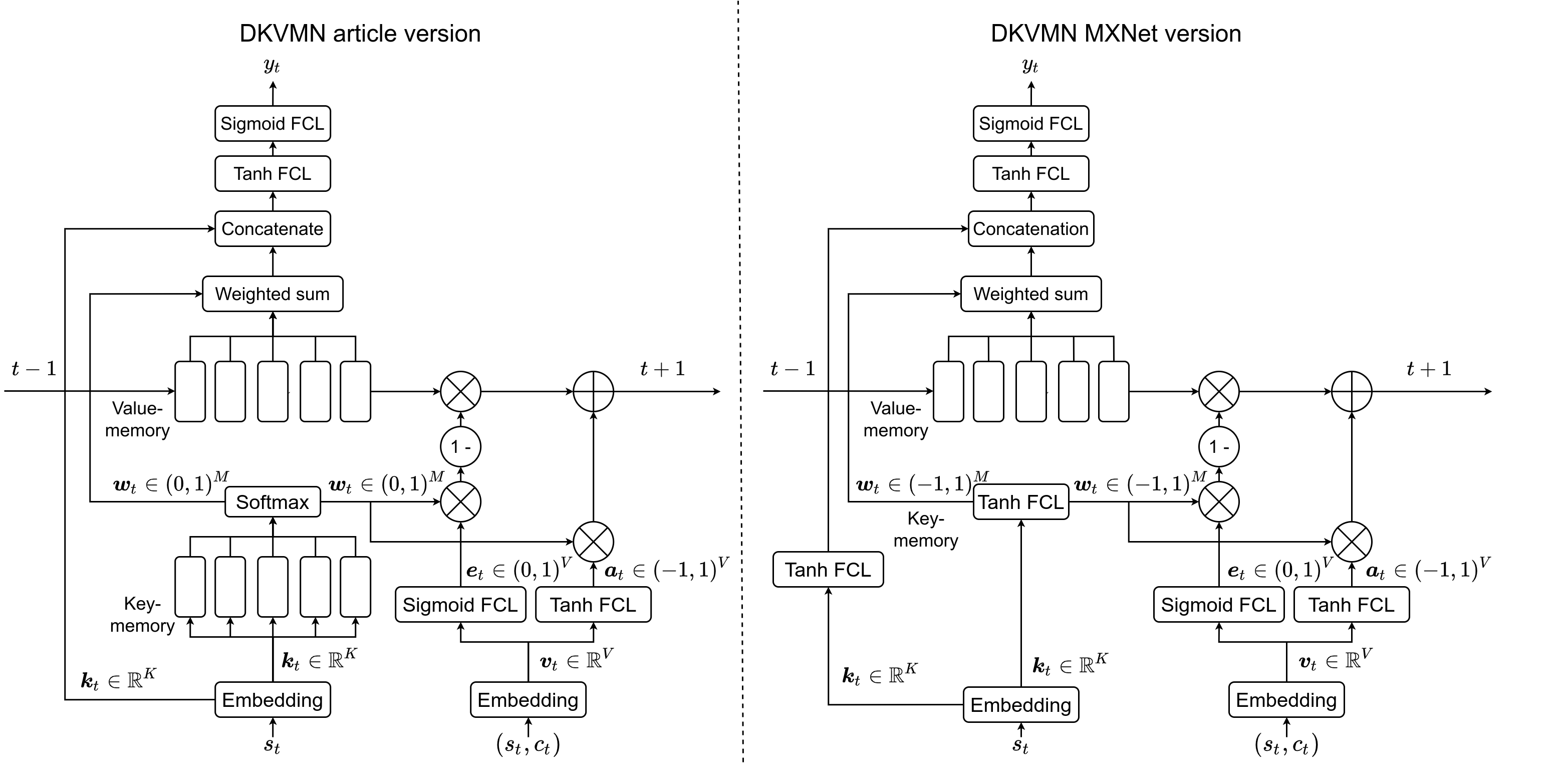}
    \caption{DKVMN architectures for the article and the MXNet implementation based on \url{https://github.com/jennyzhang0215/DKVMN}.}
    \label{fig:dkvmn-versions}
\end{figure}

During reimplementation and verification of DKVMN, we noticed that the MXNet implementation found on the author's GitHub page\footnote{\surl{https://github.com/jennyzhang0215/DKVMN}, accessed 2020-01-01} included a fully connected layer (FCL) with tanh activation function for generating key vectors from key embeddings, which is not mentioned in the original article nor in the GitHub documentation.
Notably, we find that the MXNet implementation replaces the key memory layer described in the article with a regular dense layer with softmax activation.
The differences between the two models (the version presented in the article and the version in the implementation) is shown in the Figure \ref{fig:dkvmn-versions}
that represents the model architectures of the article version and MXNet version.
Both the article version and the MXNet version have been reimplemented for the present evaluation.

\subsubsection{Self-Attentive Knowledge Tracing}

Self-Attentive Knowledge Tracing (SAKT) \cite{p2019selfattentive} is an attentive recurrent neural network that is based on the self-attentive neural network Transformer \cite{vaswani2017attention}.
Similar to DKVMN, SAKT uses embedding layers to create dense vector representations of encoded skill-correctness inputs and next skill key inputs.
The key embeddings and value embeddings in SAKT are created as described in the read process of DKVMN with the addition of positional encoding to value embeddings. The positions are added to value embeddings to add a notion of order in time, because unlike RNNs, attention networks do not operate sequentially. SAKT leverages a self-attention layer to process sequential input without explicit ordering that effectively reveals distance in time.

To be more precise, SAKT incorporates a self-attention layer called multi-head attention that computes attention
for each of its inputs simultaneously.
For preventing the use of future value inputs, SAKT uses a mask on the multi-head attention that constrains the attention computation to only the preceding inputs.
The self-attention layer outputs a vector of real values that is then passed to a feed-forward network that outputs the attempt correctness predictions of the SAKT model.
As opposed to attention in DKVMN, while DKVMN leverages attention for its read values per each timestep, SAKT
applies attention jointly on the whole key and value sequences that precede each attempt.

More formally, as in DKVMN, the inputs for SAKT are key embeddings $\bs{k}_t$ and value embeddings ${\bs{v}}_t$ for each timestep $t$.
To add information of timesteps in SAKT, position embeddings $\bs{p}_t=\bs{W}^p(t)$ are added to value embeddings to create positioned value embeddings $\bs{v}_t^p=\bs{p}_t+\bs{v}_t$.
These embeddings are then passed on to the multi-head attention mechanism.

Multi-head attention first projects its inputs, key and positioned value embeddings, and projects them with linear dense layers $\bs{W}^k\in{\mathbb{R}^{K\times A}}$ and $\bs{W}^v\in{\mathbb{R}^{V\times A}}$  to query vectors $\bs{q}_t^a=\bs{k}_t\bs{W}^k$, key vectors $\bs{k}_t^a=\bs{v}_t^p\bs{W}^v$ and value vectors $\bs{v}^a=\bs{v}_t^p\bs{W}^v$, each of size $A$.
Note that the key embeddings serve as query inputs in multi-head attention terminology, and that the positioned value embeddings serve as both key and value inputs for the multi-head attention.
The projections are fed into a dot-product attention layer \cite{luong2015effective}. Attention$(\bs{q}_t^a, \bs{k}_t^a, \bs{v}_t^a)=\text{softmax}(\frac{\bs{q}_t^a\cdot\bs{k}_t^a}{\sqrt{A}})\cdot\bs{v}^a$, whose outputs are then concatenated and projected once more with a linear fully connected layer.

\begin{figure}
    \centering
    \includegraphics[scale=.17]
    {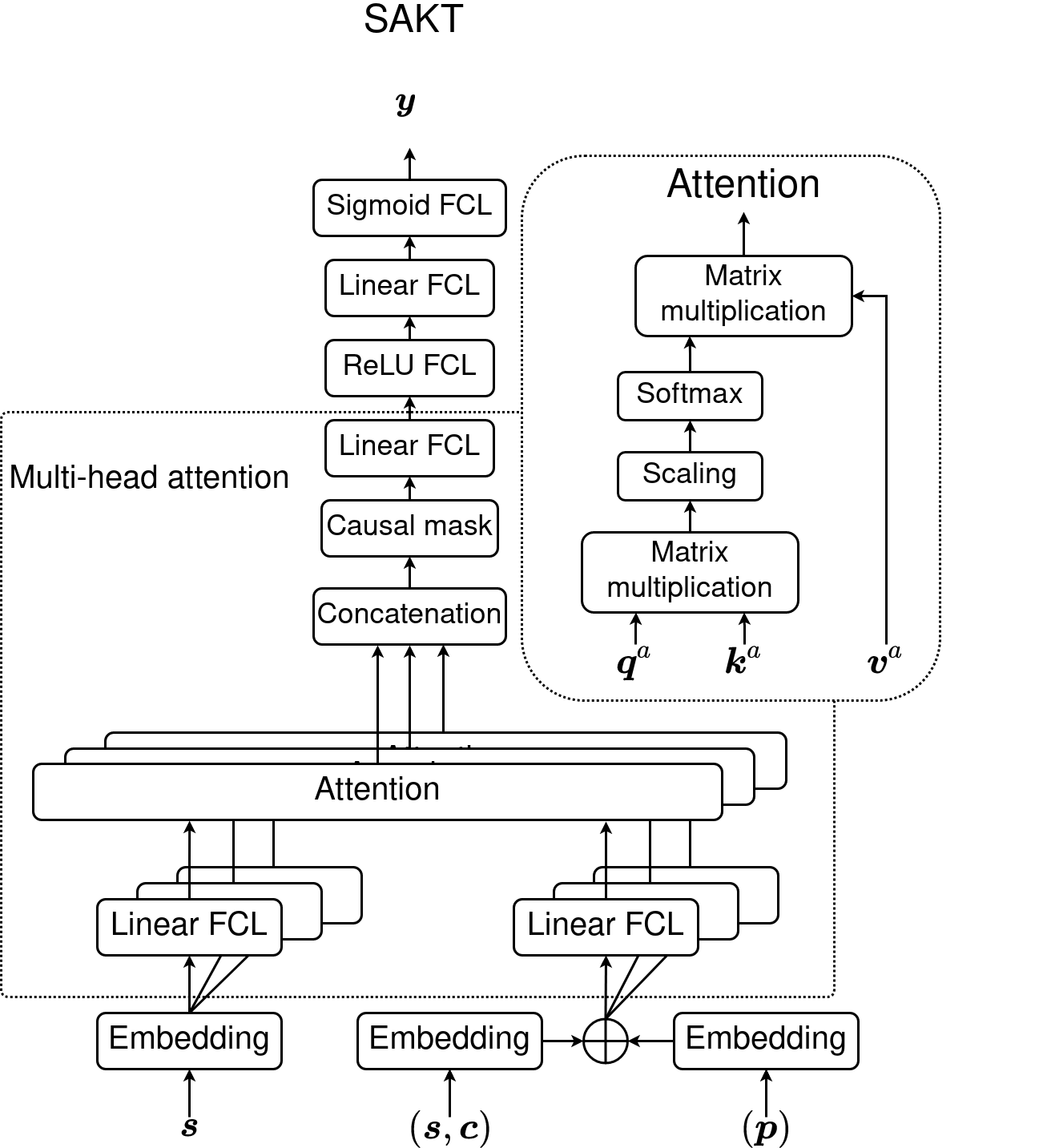}
    \caption{Self-Attentive Knowledge Tracing (SAKT) architecture. In SAKT, value embeddings are used for both value inputs $\bs{v}^a$ and key inputs $\bs{k}^a$ in the attention layer, i.e.\@ $\bs{k}^a=\bs{v}^a$.}
    \label{fig:sakt}
\end{figure}

SAKT processes the outputs of the multi-head attention with  a feed-forward network that consists of three fully connected layers, where the first layer is ReLU activated and the second is linear.
The final layer is a one neuron sigmoid activated output layer for producing exercise correctness probabilities.
The model architecture and the attention mechanism are illustrated in Figure \ref{fig:sakt}.

\subsection{Model variations}

\begin{figure}
    \centering
    \includegraphics[scale=.17]
    {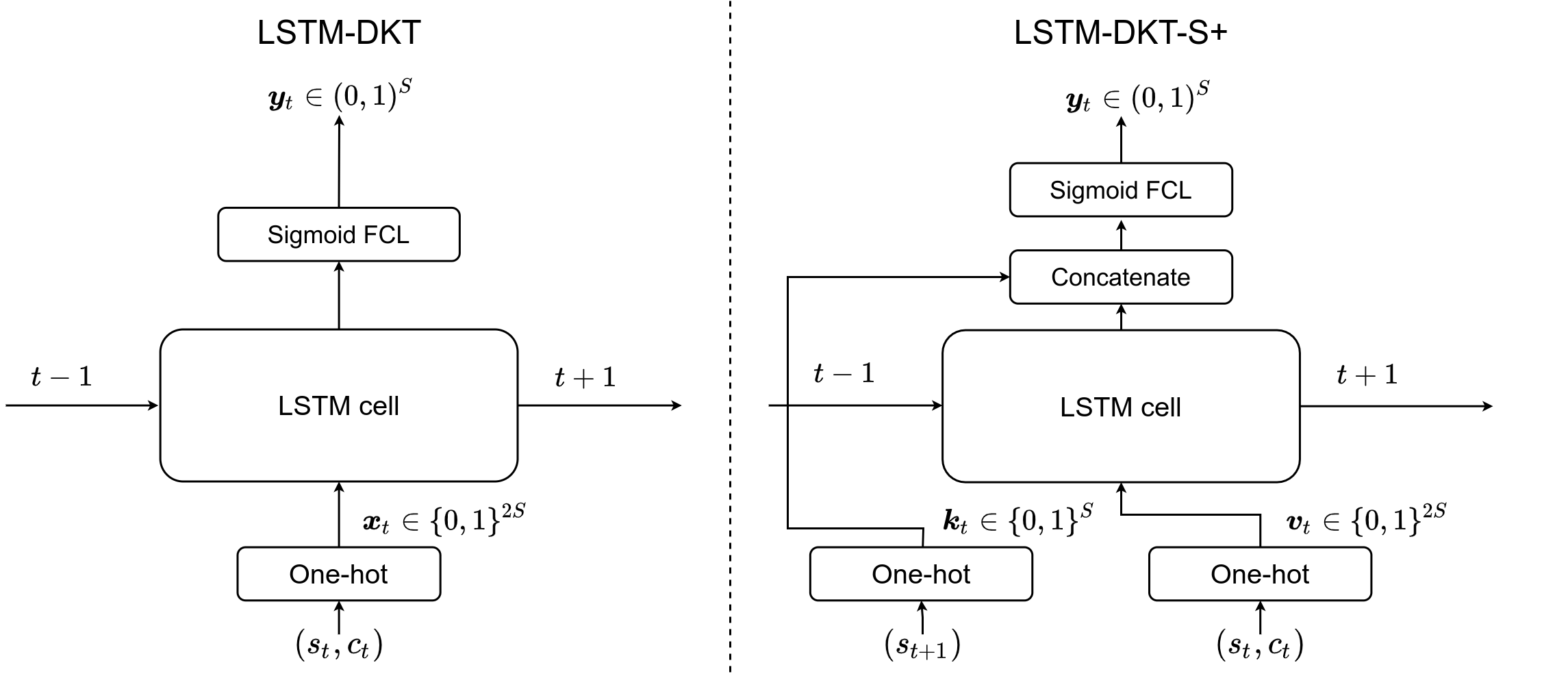}
    \caption{Architectures of LSTM-DKT, described in~\ref{subsec:vanilla-dkt}, and LSTM-DKT-S+ that includes information about the future skill in the input.}
    \label{fig:lstm-dkts-var}
\end{figure}

Next, we describe and summarize the model variations that we include to the present study. First, the RNN-DKT contains two kernel versions, a vanilla kernel and an LSTM kernel. The kernel is the main component of RNN-DKT, and therefore we consider the two RNN-DKT versions, Vanilla-DKT and LSTM-DKT, as separate models. Second, the DKVMN model has a model variant that modifies its recurrent behaviour. The inclusion of the two variations are due to the observed differences between the original DKVMN article and the authors' MXNet implementation as discussed at the end of Section~\ref{subsec:dkvmn}. Third, we also introduce our own variation to LSTM-DKT that is based on the differences of DKVMN and SAKT when compared to the original LSTM-DKT. Aside from the major architectural differences of the models, we seek to quantify the effect of giving the LSTM-DKT model information about the next attempt skill input, as is done in DKVMN and SAKT. The LSTM-DKT-S+ that includes key vector inputs at time step $t+1$ in the same fashion as in DKVMN and SAKT is shown next to the original LSTM-DKT model in Figure \ref{fig:lstm-dkts-var}.

In addition, we examine the impact of input and output processing variations for each model and model variation, i.e.\@ one-hot embedding versus using an embedding layer, and the \red{skills-to-scalar} output in SAKT and DKVMN versus the output-per-skill layer in the original RNN-DKT models. 
\red{Whereas one-hot embedding produces a vector with a size relative to the number of skills in a dataset which risks the vectors becoming impractically large, embedding layer allows one to keep the embedding size smaller by learning how to represent categorical inputs as arbitrary sized real-valued vectors. Piech et al.~\cite{Piech2015} acknowledged this problem in one-hot embedding already in their introduction of the RNN-DKT models. They proposed using random low-dimensional Gaussian vector encodings to tackle the problem, which corresponds to using embedding layer minus the learning.
For the output layer variations there are two distinct differences: 
The most obvious difference is that skills-to-scalar introduces an additional layer with the activation function tanh to the recurrent outputs.
The second, more subtle difference, is in how the learned skills are represented and learned in the output layer(s). When training the output-per-skill variant the skill weights are trained separately, only the output weights that correspond to the next attempt skill is trained out of all of the skill weights.
Conversely, for the skills-to-scalar variant, all of the skill weights (input weights of the skill summary layer) are trained for each output regardless of skill. The final scalar output corresponds the probability of the student getting the next attempt correct regardless of the skill unless the model is provided with next skill information as input.}

\begin{figure}
    \centering
    \includegraphics[scale=.17]
    {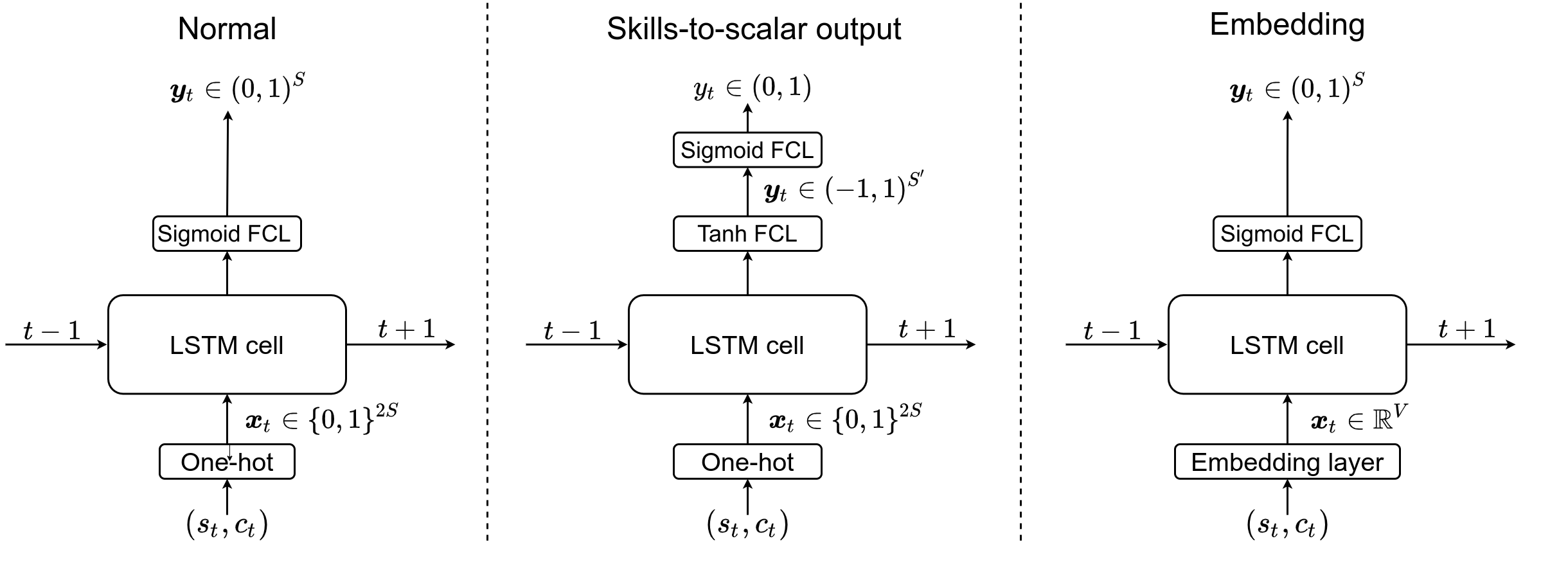}

    \caption{Output layer and input encoding variations shown for the LSTM-DKT model. In the present study, the same variations have been applied to the other models as well.}
    \label{fig:dkt-variations}
\end{figure}

These variations are \red{illustrated} for the LSTM-DKT model in Figure \ref{fig:dkt-variations}. In general, they allow us to gain insight into how much of the models' performance differences can be attributed to minor model modifications that are interchangeable between the models as they affect only pre- and post-sequential processing.

The model variations in input and output processing are minor architectural changes and they can be easily implemented as hyperparameters. Thus, we consider the input and output model variations as hyperparameters among other tunable model variables that may or may not affect model performance. The hyperparameter optimizations are outlined in more detail in the next Section.

\section{Methodology}
\label{sec:methodology}


\subsection{Datasets}

For the purposes of our study, we use seven datasets (summarized in Table~\ref{tbl:datasummary}), which are as follows: 
1) ASSISTments 2009 updated\footnote{Retrieved from \surl{https://github.com/jennyzhang0215/DKVMN/tree/master/data/assist2009_updated}, accessed 2020-01-15}, 
2) ASSISTments 2015\footnote{Retrieved from \surl{https://sites.google.com/site/assistmentsdata/home/2015-assistments-skill-builder-data}, accessed 2020-01-15}, 
3) ASSISTments 2017 Data Mining Competition\footnote{Retrieved from \surl{https://sites.google.com/view/assistmentsdatamining/dataset}, accessed 2020-01-15},
4) Statics 2011\footnote{Retrieved from \surl{https://github.com/jennyzhang0215/DKVMN/tree/master/data/STATICS}, accessed 2020-01-15}, 
5) Synthetic-5 (k=2) and 6) Synthetic-5 (k=5) \footnote{Retrieved from \surl{https://github.com/chrispiech/DeepKnowledgeTracing/tree/master/data/synthetic}, accessed 2020-01-15}, and
7) a new dataset called IntroProg\footnote{Available with the source code and results at \surl{https://tinyurl.com/dlkt-anon-sources-and-res-zip}}.

The first six of the datasets have been previously used in knowledge tracing evaluations, while the seventh one is a new dataset created for the purposes of this study. The ASSISTments 2009 updated, 2015, and 2017 come from the ASSISTments system~\cite{heffernan2006assistment,heffernan2014assistments,feng2009addressing}. The 2009 updated is an updated version of the original 2009 dataset, accounting for DKT-specific evaluation related issues in the original dataset~\cite{xiong2016going}. The Statics 2011 dataset comes from an engineering statics course~\cite{steifoli}, and the Synthetic-5 datasets are simulated datasets which were originally used to test the LSTM-DKT model \cite{Piech2015}. The synthetic datasets have been built using IRT to generate student responses for $k\in 1..5$ hidden exercise concepts for a sequence of 50 exercises per student -- for the present study, we include data for $k=2$ and $k=5$ as these are pre-generated and publicly available. The $k=5$ version has also been used in later works

The seventh dataset, IntroProg, contains information from a total of 3273 students. The data comes from a 2 ECTS\footnote{ECTS stands for European Credit Transfer System. One ECTS is approximately 25 to 30 hours of work, although this naturally varies between students.} introductory programming course organized by a Nordic research-oriented university. In the course, students learn principles of procedural programming (i.e.\@ input/output, variables, conditionals, loops) and learn to work with basic data structures (lists, maps). In total, the course has 61 programming exercises. 
The course is offered as an online textbook with an integrated programming environment and an automated assessment system. Whenever a student submits an exercise to the automated assessment system, the system records the submission as well as information on the correctness of the attempt. 
Each entry in the data includes a student id, an exercise id, and correctness of the attempt.
An attempt is classified as correct when all instructor-written tests pass for the submitted exercise.
There is no limit for the number of submissions to exercises. Only the best submission is considered when grading the course, which means that the student can continue working on an exercise even after a failed submission. This is in contrast to e.g.\@ ASSISTments datasets, where students are given a new exercise once they submit it, regardless of the correctness.

For data preprocessing, we follow the methodology of \cite{zhang2017dynamic}. If the data does not include skill identifiers, exercise identifiers are used instead as skill identifiers. As an exception, for the Statics dataset, we use exercise identifiers instead of skill identifiers. Data rows that do not contain a skill identifier, a user identifier, or correctness are discarded from the analyses. That is, every attempt must contain a skill identifier (or an exercise identifier for datasets where such is used as skill identifier), a user identifier and a correctness value. Similarly, for datasets where the correctness of an attempt was not a binary variable (ASSISTments 2015), we only used the attempts that were correct or incorrect. Finally, if a student has at most one attempt, data from the student is excluded as such data is impractical for training or evaluating the models.

\begin{table}[htb]
    \small
    \centering
    \caption{Summary of the datasets used in our evaluation. ``Max Att.'' stands for maximum attempts in the data for a single student, and ``E. ids'' stands for the total number of exercise identifiers in the data. \label{tbl:datasummary}}
    \begin{tabular}{lrrllrl}
        \toprule
         Dataset & Max Att. &  Students & Attempts & Correct (\%) &  E. ids & Skill ids \\
         \midrule
         ASSISTments 2009 (u) & 1261  &  4151  &  326k  &  214k (66\%) &  110
         \\
         ASSISTments 2015 & 632 &     19917 &    683k &          514k (72\%) &            100 &          - \\
         ASSISTments 2017 & 3057 &      1709 &    943k &         351k (37\%) &           3162 &         102 \\
        Statics 2011 & 1181 &      333 &    189k &         145k (77\%) &           1223 &         98 \\
        Synthetic-5 K=2 & 50 &      4000 &    200k &          137k (69\%) &             50 &          - \\
        Synthetic-5 K=5 &  50 &      4000 &    200k &          122k (61\%) &             50 &          - \\

        IntroProg        & 1857  &    3273 &    172k &          84k (49\%) &             64 &          - \\
        \bottomrule
    \end{tabular}
\end{table}

\subsection{Approach}

\subsubsection{Metrics\label{subsec:methodology-metrics}}

Results are reported using seven metrics: Accuracy (ACC), Area Under the Curve (AUC), Precision, Recall, F1 score, Matthews Correlation Coefficient (MCC), and Root-Mean-Square Error (RMSE).
Multiple metrics are included, since e.g.\@ \cite{caruana2004data} and \cite{national2005thinking} suggest that models can perform differently depending on the metric and that using a single metric can lead to misguided judgement of model performance.
Accuracy is included as an intuitive metric that tells the overall correct prediction percentage.
The AUC, which is more precisely the Area Under the Curve of Receiving Operator Characteristic (AUROC, or ROC-AUC), is a commonly used metric in DLKT models. It is ``equivalent to the probability that the classifier will rank a randomly chosen positive instance higher than a randomly chosen negative instance.''~\cite{fawcett2006introduction}.
Precision gives insight on how many of the predicted positive values were actually positive, whereas Recall tells how many of all positive instances were correctly identified as positive.
F1 score is the harmonic mean of Precision and Recall and is an often used metric although susceptible to imbalances in data \cite{chicco2020advantages}.
Matthews correlation coefficient
(MCC) provides high scores only if predictions are accurate regardless of the rate of negative and positive elements in a dataset. RMSE is a commonly used metric, which unlike the other metrics, is computed from raw prediction values. These metrics are summarized in Table \ref{tbl:metrics}, where confusion matrix terms true positives (TP), false positives (FP), true negatives (TN) and false negative (FN), are used for defining some of the metrics.

\begin{table}[H]
    \centering
    \caption{Definitions for the used metrics. True Positive (TP), False Positive (FP), True Negative (TN), and False Negative (FN) come from confusion matrix terminology.}
    \label{tbl:metrics}
    \small
    \setlength\extrarowheight{10pt}
    \begin{tabular}{m{1.5cm} m{11cm}}
         Accuracy & $\frac{TP + TN}{TP + FP + TN + FN}$\\
         Precision & $\frac{TP}{TP + FP}$\\
         Recall & $\frac{TP}{TP + FN}$\\
         F1 & $ 2 * \frac{\text{Precision} * \text{Recall}}{\text{Precision} + \text{Recall}}$\\[10pt]
         AUC & $\vcenter{\hbox{\strut Area under a plot line with recall on y-axis and false positive rate}\hbox{\strut $(\frac{FP}{FP + TN}$) on x-axis for decision thresholds from 0 to 1}}$\\
         MCC & $\frac{TP \times TN - FP \times FN}{\sqrt{(TP + FP)(TP + FN)(TN + FP)(TN + FN)}}$ \\
         RMSE & $\sqrt{\frac{\sum_{i=1}^{N}(c_i - y_i)^2}{N}}$
    \end{tabular}
\end{table}

For confusion matrix based metrics, we consider the attempt correctness $c_t$ value 1 as a positive label and 0 as a negative label.
To determine the polarity of predicted values, i.e.\@ whether a value is positive or negative (0 or 1), we use 0.5 as the decision threshold.
This means that a prediction $y_t$ is considered positive if it is equal or greater to 0.5.
RMSE is an exception in the used metrics as it does not use confusion matrix categories but is computed directly from model prediction values.
AUC is also distinct from most other included metrics as it is independent of a decision boundary due to it being computed from a plot over all decision thresholds (from 0 to 1).
Out of the seven metrics, AUC, and MCC take class imbalance into account, which can be beneficial on interpreting results from skewed datasets, although it can mask poor performance by weighing underrepresented labels too heavily~\cite{lobo2008auc,jeni2013facing}.

Our simplest baseline model, Mean (majority vote for binary metrics) is by definition unable to produce meaningful AUC, MCC, Recall, Precision (when majority of labels are negative), and by extension F1 scores.
For AUC, this happens for the Mean since for each decision boundary, every predicted value is on the same side of the decision boundary, wherefore Recall equals false positive rate and thus AUC is a constant 0.5.
MCC cannot be computed due to division by zero, which also applies for Precision and Recall when the majority of labels are negative. When the majority of labels are positive, Precision is meaningful but Recall will always be one.

\subsubsection{Naive and non-DLKT model training}

The naive models included in the study are outlined in Section~\ref{subsec:naive-baselines}. The naive models do not require training apart from using training set to determine the mean correctness for the \textit{Mean} model. In addition, we use two non-DLKT baseline models BKT and GLR that are statistical classification models and contain parameters that need to be trained, i.e.\@ learned. These are presented in Sections~\ref{subsec:bkt} and~\ref{subsec:best-lr}.

For BKT, we use the publicly available \texttt{hmm-scalable}\footnote{\surl{https://github.com/IEDMS/standard-bkt},\surl{https://github.com/myudelson/hmm-scalable}, accessed 2020-03-01} implementation by Yudelson, which includes individualized BKT~\cite{yudelson2013individualized}. 
We evaluate the available solvers for the BKT implementation and pick the best results per dataset according to the RMSE metric.
We choose the best solver based on RMSE instead of AUC since by inspecting the results we find that there is minimal difference in AUC scores compared to choosing the best model by AUC. However, especially for the ASSISTments 2015, the best model by AUC is significantly worse on most other metrics. Also, previous studies have shown that AUC is a less viable metric for BKT than RMSE ~\cite{dhanani2014comparison,pelanek2015metrics}.
For other hyperparameters we use the default settings.
For GLR, we use the publicly available implementation\footnote{\surl{https://github.com/theophilee/learner-performance-prediction}, accessed 2021-05-27} by Gervet et al. discussed in~\cite{gervet2020deep}. When training the model, we use the default L-BFGS solver provided in the implementation.

We use \red{$5$-fold} cross validation to evaluate the models, i.e.\@ the data is divided into five parts, and the models are then trained and evaluated five times using one part as the test set and the other parts as the training set. Each part is used once in test, and four times in training. The reported results are averaged over the five training and testing iterations. Dividing the data into the parts is performed by student, meaning that data from each student is present only in one part at a time and not divided over the parts. 

In order to have the GLR model implementation comply with our methodology, we modified the data preprocessing and added \red{$5$-fold} cross validation.

\subsubsection{DLKT Model training and hyperparameter optimization}
\label{subsubsec:hyper-summary-model-training}


All DLKT models are optimized using binary cross-entropy (log loss): \@ $-(c_i\log(y_i) + (1 - c_i)\log(1 - y_i))$, which is an often-used loss function for training neural networks.
The machine learning models are evaluated using $5$-fold cross validation.
The models are given one hundred training iterations with early stopping, where early stopping is performed if ten consecutive iterations do not improve the performance of the model, measured with binary cross-entropy\footnote{\red{Typically, a ``main'' evaluation metric is selected for early stopping. Here, we deem the training metric more suitable as we aim to measure the effect of evaluation metric choice.}}.
A validation set which consists of 10\% of the training data is used to determine the early stopping.
Thus, the training data for each fold comprises 70\% of the data while the remaining 20\% is used for evaluation.

Grid search is used to find optimal hyperparameters for each model, summarized in Table~\ref{tbl:hypersummary}. The best hyperparameters are selected based on \red{averaged} AUC \red{score in the $5$-fold cross-validation}.
The hyperparameters regarding \red{model layers} include recurrent layer size (memory size for DKVMN), key embedding size, value embedding size and skill summary layer size. Additionally, the number of attention heads is \red{tuned} for SAKT as it uses multi-head attention.
The use of layer sizes depends on the model variation as some models and/or their variations either do not have a corresponding layer or the size is a set constant.
For one-hot input model variations, the embedding sizes are not applicable as one-hot size is a constant determined by the count of distinct skill identifiers in a given dataset.
Likewise, output-per-skill output layer size is determined by the same distinct skill count to provide an output for each skill as the name suggests. For model variations with such an output layer, a skill summary layer is not used.
The DKT models apart from LSTM-DKT-S+ do not have separate key inputs, and thus key embedding size does not affect it.

\begin{table}
    \small
    \centering
    \caption{Overview of hyperparameters used for training DLKT models. Although there are a maximum of two options per hyperparameter, the variation count is not a power of two since some options cancel each other out. Also, not all hyperparameters affect every model, e.g.\@ SAKT is the only model that uses attention heads.} 
    \begin{tabular}{lc}
    \toprule
        Hyperparameter & Options \\
    \midrule
        Recurrent layer size	& \{50, 100\} \\
        Key embedding size	& \{20, 50\} \\
        Value embedding size	& \{20, 50\} \\
        Skill summary layer size &\{50, 100\} \\
        Input variation & \{One-hot, Embedding\} \\
        Output variation & \{Output-per-skill, Skill summary + Scalar\} \\
        Learning rate & \{0.01, 0.001\} \\
        Dropout	& \{0.2\} \\
        Attention heads	& \{1, 5\} \\
        Batch size	& \{32\} \\
        Random Seed  & \{13, 42\} \\
        Optimizer & \{Nadam\} \\
        \midrule
        Variations & 72-240 (depending on model) \\
        \bottomrule
    \end{tabular}
    \label{tbl:hypersummary}
\end{table}

Besides layer-specific hyperparameters, we tune the models for learning rate and max attempt count. Although the latter is not necessary, it is used both in DKVMN and SAKT due to implementation restrictions and possibly for faster training times.
While the DKVMN article does not mention max attempt count, the MXNet implementation by the authors uses 200 as the max attempt count.
SAKT authors mention using different max attempt counts for different datasets, where the values for max attempt count range from 50 to 500.
Both DKVMN and SAKT implement max attempt count by splitting exceeding attempts as a new attempt sequence recursively until no attempt sequence exceeds the max attempt count, effectively creating new students into the dataset.
We note that our implementations published as a part of this article do not include this restriction and can be trained with arbitrarily large attempt sequences, although the training time and space requirement increases significantly without max attempt counts.
For the grid search, we use the max attempt count of 200 both for reproducibility purposes and to reduce model training time. 

We also include random number generator seed as a hyperparameter to provide a baseline for evaluating the effect of hyperparameter tuning.
The differences in the results of models whose hyperparameters differ only regarding the random seed can be safely assumed to be caused by randomness.
This gives insight into what extent randomness may affect the \red{$5$-fold} cross validation results.

All DLKT models were trained using input batches of size of 32, 20\% dropout and the Nadam gradient descent optimizer~\cite{dozat2016incorporating}. 
While all of these may affect the results, we decided to not tune them systematically for each model to keep the hyperparameter option space reasonably sized.
In addition to comparing the performance of the models where the best hyperparameters are selected based on AUC, we explore selecting best hyperparameters with other metrics. Target metric used for model training, for instance, can influence selected features and model performance on other metrics~\cite{sanyal2020feature}. 
We analyze the effect of choosing a metric in grid search by comparing the perceived performance loss in other metrics. By loss, we mean the difference between the performance of the model with ``best'' hyperparameters selected based on a given metric, and the best result achieved when selecting models with some other metric. As an example, we compare the performance of a model that was selected based on AUC with models selected by other metrics to determine how the performance over all metrics differ between these models.

We acknowledge that we only included two options for each tuned hyperparameter. This, along with the untuned options, leaves room for speculation whether more broad tuning would provide significantly different results. All in all, considering all datasets and model variations, the number of trained and evaluated DLKT models totals 5,880 cross-validated models.

\subsubsection{Maximum attempt cut}

As briefly mentioned,  DKVMN and SAKT, which leverage maximum attempt counts due to the restrictions in their original implementations, split input sequences according to the maximum attempt count hyperparameter.
This means that if the maximum attempt count is 100 and a user has 950 attempts, the attempts would be split into 9 sets of 100 attempts and one with 50 attempts.
Effectively, this multiplies the number of data points that the models are trained on, since the models consider the different sets from the same student as completely new students.
Instead of having fewer students with longer attempt sequences, the models have more students with some having unnatural starting points for their attempt sequences.
As it is poorly explained why this would be the preferred case, we test to what extent cutting off attempts that exceed the maximum attempt count affect the results. We use the same 200 as maximum attempt count splitting for maximum attempt count cutting. Further, we test the models for no maximum attempt count.
The effect of splitting or cutting is evaluated only for the best hyperparameter options determined by the grid search using the hyperparameter options in Table~\ref{tbl:hypersummary}.

\subsubsection{Hardware and Software}

The model training in this work was performed using computer resources within the Aalto University School of Science ``Science-IT'' project.
All DLKT model training was conducted using CPUs unless otherwise noted.
The models were implemented in Python using TensorFlow version 2.1.0.

Further, to evaluate the effect of hardware, the models were re-trained and evaluated using GPUs. GPU evaluation was only conducted for the hyperparameter values deemed best by grid search using CPUs.
Due to implementation restrictions, as well as to compare machine learning framework versions, the TensorFlow version was updated to
2.6.2 for GPU computation. Thus, we also retrained the models with CPU on the new
TensorFlow 2.6.2 version for the CPU versus GPU comparison and present CPU results for both TensorFlow 2.1.0 and 2.6.2, while the GPU results are presented only for TensorFlow 2.6.2.

\section{Results}
\label{sec:results}



In the following result tables, the best metric values for a given model have been selected by picking the model with the highest AUC score. 
Thus, the actual highest scores for other metrics may be higher than in the presented tables because the best model as determined by AUC might not yield the best results for all metrics. Section~\ref{subsec:metric-and-determining-the-best-model} presents this aspect in greater detail. \red{Full evaluation results are available in the online repository alongside the source code\footnote{\surl{https://tinyurl.com/dlkt-anon-sources-and-res-zip}}.}

\subsection{Baseline models and DLKT models \label{subsec:model-performance-differences}}

Here, we mostly focus on the performance of the models using AUC as the metric. Other metrics are considered in more detail in subsequent sections. These results are outlined in Appendix~\ref{appendix:model-comparison-results}.

The DLKT models consistently produce better predictions than the simple baselines and BKT. The GLR model reaches performance that is above that of the other baselines, including BKT, consistently. For the IntroProg dataset (Table~\ref{tbl:results-introprog} \red{in Appendix~\ref{appendix:model-comparison-results}}), GLR is the best performing model, and its performance is also on par with the performance of most DLKT models on some metrics on the ASSISTments 2015 dataset (Table~\ref{tbl:results-assistments-2015} \red{in Appendix~\ref{appendix:model-comparison-results}}) and the Statics dataset (Table~\ref{tbl:results-statics} \red{in Appendix~\ref{appendix:model-comparison-results}}), while falling behind DLKT models in other datasets.

None of the baselines apart from GLR reach AUC scores that are able to compete with the scores of any of the evaluated DLKT models.
With BKT, which is the second-best performing baseline after GLR, the difference in AUC is smallest in the ASSISTments 2015, IntroProg and Statics (Tables~\ref{tbl:results-assistments-2015}, \ref{tbl:results-introprog} and \ref{tbl:results-statics} \red{in Appendix~\ref{appendix:model-comparison-results}}) where BKT falls behind the bottom DLKT model AUC score by 2\% points, 3.2\% points, and 3.4\% points respectively.
In these three datasets, the difference between BKT and the bottom DLKT model is greater than the difference between the top DLKT model\footnote{LSTM-DKT for ASSISTments 2015, DKVMN and LSTM-DKT for IntroProg, and DKVMN for Statics} and the bottom DLKT model\footnote{SAKT for ASSISTments 2015, Vanilla-DKT for IntroProg, and SAKT for Statics}; 1.1\% points, 0.6\% points and 1.7\% points for ASSISTments 2015, IntroProg and Statics respectively. Notably however, the best performing model for IntroProg is not a DLKT model but GLR: the difference in AUC score between GLR and the top-performing DLKT models LSTM-DKT and DKVMN is 1.6\% points.

When considering BKT, it is in some cases unable to beat the naive baselines, in addition to its poorer performance compared to the DLKT models or GLR. This is evidenced in the Synthetic-K2 dataset (Table~\ref{tbl:results-synthetic-k2} \red{in Appendix~\ref{appendix:model-comparison-results}}), where BKT achieves a lower AUC score (0.635) than the NaP3M (0.654) or the NaP9M (0.709).

Overall, the DLKT models surpass GLR and other baselines as DLKT models outperform the best-performing baseline GLR on 3 out of 5 real datasets and on both synthetic datasets. A clear difference between model performances of the two best DLKT models (DKVMN and LSTM-DKT) versus baselines can be seen for each metric in Figure \ref{fig:baseline-comparison}, which shows scores averaged over datasets.

\subsection{Model performance on different metrics \label{subsec:model-performance-and-metrics}}

In the previous Section~\ref{subsec:model-performance-differences}, we focused on the performance of the models when using AUC as the metric. Next, we look deeper into the differences in model performance when considering other metrics. These metrics are accuracy, precision, recall, F1, MCC, RMSE, which are presented in Section~\ref{subsec:methodology-metrics}.

\begin{figure}[ht!]
    \centering
    \includegraphics[scale=.42]
    {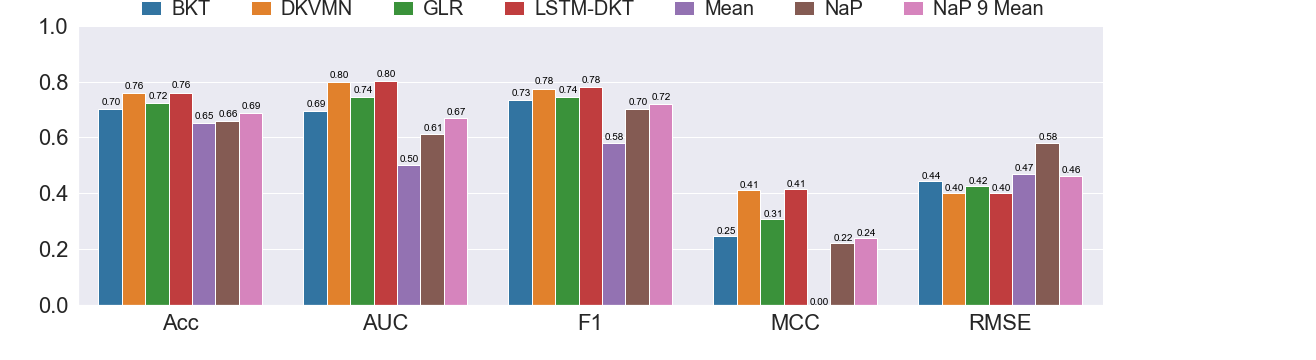}
    \caption{Result comparison over the two best performing deep learning models over baselines. Scores are averaged over all datasets}
    \label{fig:baseline-comparison}
\end{figure}

As noted previously, the DLKT models overall outperform GLR as well as other baselines when using AUC. The same observation can be made when studying the average performance of the models over all datasets, presented in Figure~\ref{fig:baseline-comparison} (the figure includes the two best performing DLKT models and the baselines). When contrasted with AUC results, we however observe more variation between the performances of the DLKT models and the baselines. 
When looking at the results in Tables~\ref{tbl:results-introprog} and \ref{tbl:results-statics} \red{in Appendix~\ref{appendix:model-comparison-results}} for different metrics, GLR ranks among the best on IntroProg and Statics datasets. In IntroProg, GLR receives the best scores for Accuracy, AUC and MCC (0.4\% points, 1.6\% points, and 3.1\% points respectively). In Statics, GLR receives scores similar to most DLKT models. In the other datasets, DLKT models perform better on all metrics, excluding precision and recall. 

When ranking the models based on their performance in the different metrics, we observe a degree of variation between the metrics and the datasets. For example, for the ASSISTments 2015 dataset (Table \ref{tbl:results-assistments-2015} \red{in Appendix~\ref{appendix:model-comparison-results}}), the simple Mean baseline has F1-score (0.849) which is equal to the F1-Score of GLR, topping all DLKT models. Similarly, for the same dataset, the Mean baseline's accuracy (0.738) is close to the accuracy of the complex models, falling short 1.3\% points from the best performing model (LSTM-DKT). When studying the average performance of the GLR and the best two DLKT models over all datasets, shown in Figure~\ref{fig:baseline-comparison}, we see that the differences when measured with RMSE are somewhat small, for Accuracy and F1 moderate, and for AUC and MCC the differences are rather clear.

Considering only the DLKT models, the inter-model performance differs depending on the observed metric, and the differences partially depend on the dataset. For example, for the ASSISTments 2009 Updated dataset (Table \ref{tbl:results-assistments-2009-updated} \red{in Appendix~\ref{appendix:model-comparison-results}}), the accuracy, F1-Score and RMSE between the DLKT models, excluding SAKT, is at most 1\% point, while AUC shows 1.6\% point difference between the best and worst DLKT models.

There are some cases where the ranking of the models depends on the chosen metric, although the differences are often small. As an example, for the IntroProg dataset (Table \ref{tbl:results-introprog} \red{in Appendix~\ref{appendix:model-comparison-results}}) when comparing SAKT and GLR, SAKT slightly outperforms in F1-Score (0.759 vs 0.755), while GLR outperforms in e.g.\@ AUC (0.843 vs 0.825). Similar observations can be made also for the ASSISTments 2015 dataset (Table~\ref{tbl:results-assistments-2015} \red{in Appendix~\ref{appendix:model-comparison-results}}), where SAKT outperforms GLR in AUC (0.714 vs 0.702) and MCC (0.230 vs 0.204), while GLR slightly outperforms SAKT in F1-Score (0.849 vs 0.846) and Accuracy (0.750 vs 0.748).

\begin{figure}[ht!]
    \centering
    \includegraphics[scale=.32]
    {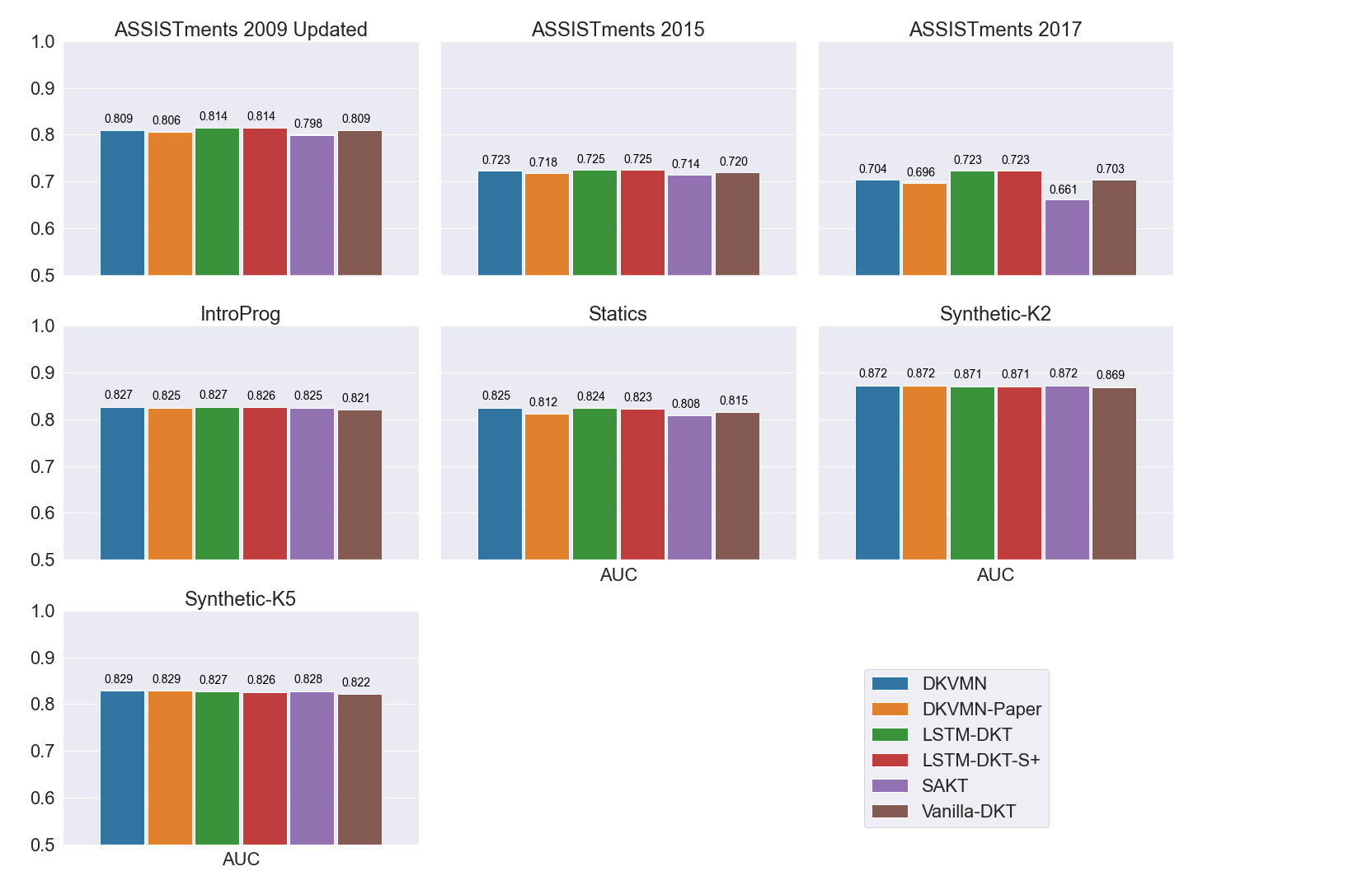}
    \caption{Best AUC scores of each DLKT model per dataset.}
    \label{fig:auc-data-over-modeller}
\end{figure}

As for the top performing model all in all, there is no absolute victor, since no model dominates all datasets and every metric. Although the differences are not major, we find that LSTM-DKT (both original and S+ versions) and DKVMN (MXNet version, i.e.\@ not DKVMN-Paper) perform consistently as the best or near the best over the evaluated models across the varying metrics. For AUC, this can be seen in Figure~\ref{fig:auc-data-over-modeller}. 
Also, the top-performing DLKT models receive on average better scores for all metrics compared to the baseline GLR or other baselines as can be seen in Figure \ref{fig:baseline-comparison}.

On closer inspection into the DLKT models, we observe that LSTM-DKT and LSTM-DKT-S+ hold the top AUC score on ASSISTments 2009, ASSISTments 2015 and ASSISTments 2017, while DKVMN holds the top AUC score on Statics, Synthetic-K2 (shared with SAKT) and Synthetic-K5. Even though LSTM-DKT models are ranked first on more datasets, the only dataset where the performance difference can be considered other than marginal is ASSISTments 2017 where the difference in AUC score is 2.1\% points. For all other datasets, the difference between the LSTM-DKT models and DKVMN is at most 0.5\% points.

Additionally, when considering the LSTM-DKT and LSTM-DKT-S+, we observe that in most cases their performance is almost the same if not the same across all datasets and metrics. The largest differences can be observed in ASSISTments 2017 in F1-Score, where the difference is 0.4\% points in favor of LSTM-DKT-S+, in IntroProg in MCC, where the difference is 0.4\% points in favor of LSTM-DKT, and in Synthetic-K2 in MCC, where the difference is 0.4\% points in favor of LSTM-DKT-S+.

All of the above model differences are derived from comparing models that have been hyperparameter tuned for the AUC score.

\subsection{Metrics and Determining the Best Model Hyperparameters}
\label{subsec:metric-and-determining-the-best-model}

\begin{figure}[ht!]
    \centering
    \label{fig:selection-metric-diffs}
    \includegraphics[scale=.31]{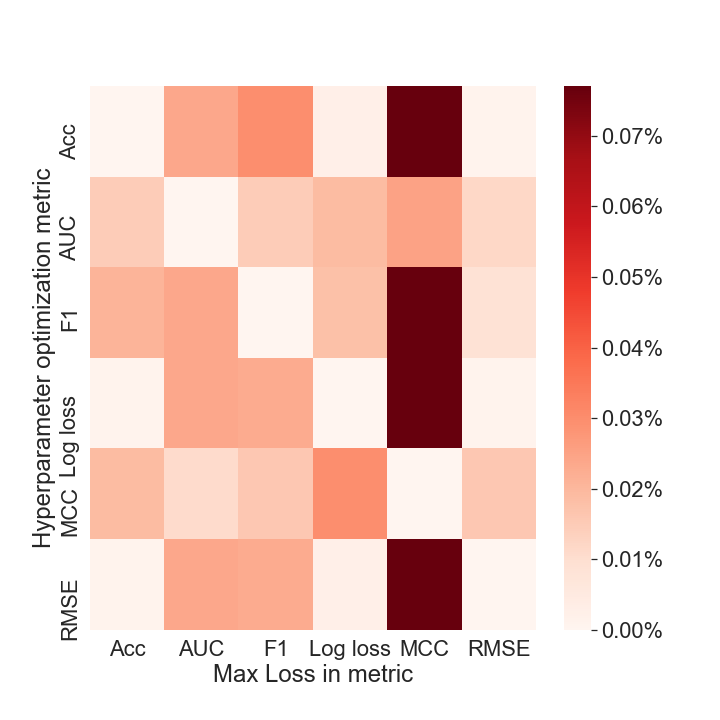}
    \includegraphics[scale=.31]{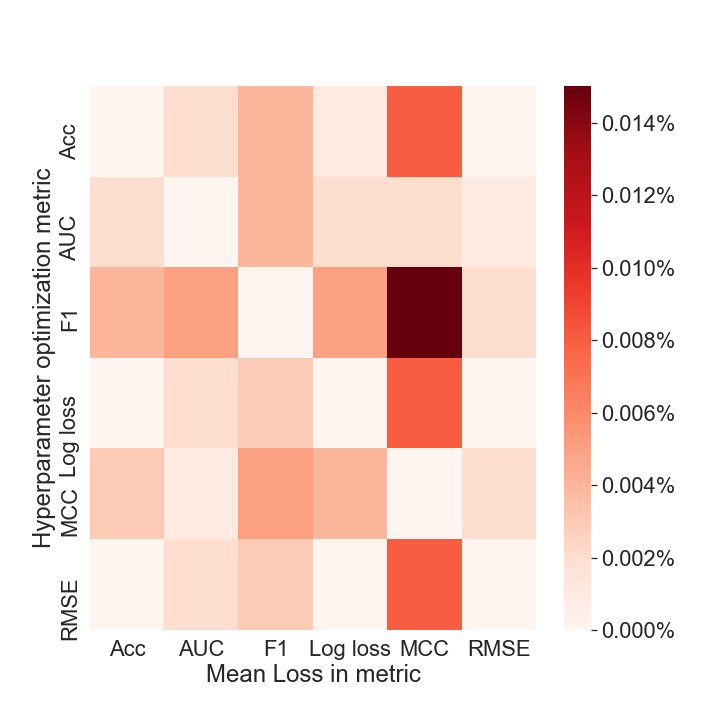}
    \caption{\red{Mean and maximum differences in metric scores compared to optimal over datasets and DLKT models for using different metrics to select the ``best'' hyperparameters.
    The loss indicates how much lower scores for other metrics can be expected when choosing a metric for hyperparameter optimization.}}
\end{figure}

In the previous parts, the comparisons were based on models that were tuned for the AUC score. To consider the impact of tuning models for other metrics, we analyzed the model evaluation results for all the hyperparameter variations for each model and dataset.
This was conducted to gain an overview of to what extent the model that is ranked the best depends on the metric that is used to pick the best model out of the trained models. These results are summarized in \red{Figure \ref{fig:selection-metric-diffs}, which shows the mean and max losses over our evaluated datasets and models} for a given metric when some other metric is used to pick the best trained model among models with different hyperparameter options.  
For this metric comparison we included the results of maximum attempt count analysis, which is discussed in the next section. In the metric comparison, in addition to the evaluation metrics, we also include log loss that was used in model training.

Overall, picking a model based on the performance measured using a specific metric most likely does not mean that the model is the best when considering the other metrics. On average, the mean losses are small but noticeable, although in some cases the mean losses can be measured in multiple percentage points. For example, when picking a model based on F1-score and then looking at the MCC score, the mean loss is 1.5\% points. For the other options, the mean losses are under 1\% point, and mostly under 0.5\% points. When considering the max loss, i.e.\@ highest loss when optimizing a specific metric and considering another metric within all the models and the datasets, the differences can be large. For example, when picking a model based on Accuracy, F1-Score, RMSE, or Log loss, and then looking at the MCC, the maximum loss is 7.7\% points. When picking a model based on Accuracy and then looking at the F1-Score, the maximum loss is 3.0\% points (similar to picking MCC and looking at Log loss). In most cases, the maximum loss is under 2.5\% points.

When focusing on AUC, which is often used as the main comparison metric in knowledge tracing studies, also consequently in our replication study, we observe that if AUC is used for hyperparameter optimization, we sacrifice up to 2.5\% points in MCC and up to 1.5\% points in Accuracy and F1-Score. In other words, using another hyperparameter set for the same model and dataset would achieve 2.5\% points higher MCC than the hyperparameter set used to obtain the best AUC score.

To summarize, according to our results, there does not appear to be a metric that would be optimal for optimizing all metrics, as all metrics when used for optimization risk losses for at least one other metric. The optimization metric comparison does however suggest that F1-score conveys the most risk, as evidenced by the mean losses which are the highest out of the studied metrics. On the other hand, looking at the max losses, all of the metrics convey a risk.

\subsection{Optimal Hyperparameters and Model Variations \label{subsec:hyperparameter-and-model-variations}}

Here we inspect the impact of hyperparameter tuning on model performance, including analyzing the effect of the input and output model variations.  The evaluated hyperparameter combinations are outlined in Table~\ref{tbl:hypersummary} in Section~\ref{subsubsec:hyper-summary-model-training}. 

The hyperparameter tuning results are presented as follows. First, we briefly outline the optimal hyperparameters in general, which is followed by input and output model variations (output-per-skill vs \red{skills-to-scalar} output, one-hot input vs embedding layer, max attempt cut vs max attempt split vs no max attempt count). To discern the magnitude of the effect of the chosen hyperparameters, we also outline results related to random seed selection and used hardware (CPU vs GPU).

\subsubsection{Optimal Hyperparameters}

The hyperparameters that yielded the best performance as measured by the AUC score are outlined in tables in Appendix~\ref{appendix:best-hyperparameters}, one table per dataset. In general, the optimal hyperparameters are model- and dataset-specific. When considering layer sizes, 
the recurrent layer size is more often smaller (50) than larger (100) for all models apart from Vanilla-DKT, key and value layer sizes do not show a pattern, and summary layer size appears more often smaller (50) for DKVMN and DKVMN-Paper and larger (100) for other models when it is used.

As for the learning rates, the models most often perform better with 0.001 as the learning rate when compared to the other option 0.01. The only exception is DKVMN-Paper, which always performs better with 0.01 except for the two synthetic datasets. Also, in the ASSISTments 2017 dataset all models but SAKT and Vanilla-DKT perform better with 0.01 learning rate. The two models, SAKT and Vanilla-DKT are also the only ones that perform best with learning rate 0.001 on all the evaluated datasets.
The number of attention-heads for SAKT also differs across datasets with 1 as the optimal for the datasets IntroProg and Statics and 5 for other datasets.

\subsubsection{Input and Output Model Variations}

\red{In order to triangulate the differences in the evaluated model architectures, we compared input and output layer variations found between the model architectures that were compatible for all of the models. We compared the output layer variations output-per-skill against skill summary layer and a scalar output layer (skills-to-scalar output layers), and the input layer variations one-hot input against embedding layer.}

When comparing output-per-skill and \red{skills-to-scalar} output, for LSTM-DKT, Vanilla-DKT and LSTM-DKT-S+, using an output-per-skill layer provided either as good or better results as opposed to using \red{skills-to-scalar} output layers. For SAKT, the differences are minimal for the best results, but with output-per-skill layer the SAKT model appears more robust to other hyperparameters as the worst scores are much higher than with \red{skills-to-scalar} output.
On the other hand, DKVMN shows an opposite effect compared to LSTM-DKT, Vanilla-DKT and LSTM-DKT-S+ with sometimes clearly better performance for skills-to-scalar output layers.
We highlight a few of these findings in Tables \ref{tbl:assistments-2017-model-output-per-skill} and \ref{tbl:assistments-2015-model-output-per-skill} included in Appendix~\ref{appendix:best-hyperparameters}.
Notably, in the ASSISTments 2017 dataset (Table \ref{tbl:assistments-2017-model-output-per-skill} \red{in Appendix~\ref{appendix:best-hyperparameters}}), DKVMN performs clearly worse when using output-per-skill than when using \red{skills-to-scalar} output layers (difference of 1.9\% points), while the difference for DKVMN-Paper is 3.8\% points. The DKT models on the other hand perform better with output-per-skill layer (LSTM-DKT 1.2\% points, LSTM-DKT-S+ 0.5\% points and Vanilla-DKT 2.0\% points)
For ASSISTments 2015 (Table \ref{tbl:assistments-2015-model-output-per-skill} \red{in Appendix~\ref{appendix:best-hyperparameters}}) DKVMN differences are much smaller and DKVMN performs slightly better with output-per-skill layer unlike in ASSISTments 2017. The results for LSTM-DKT, Vanilla-DKT and LSTM-DKT-S+ show a similar pattern as in ASSISTments 2017.
When considering the performance of the evaluated models overall, \red{skills-to-scalar} output results have more variance than the output-per-skill results, as depicted by the standard deviation of the results in Tables \ref{tbl:assistments-2017-model-output-per-skill} and \ref{tbl:assistments-2015-model-output-per-skill} \red{in Appendix~\ref{appendix:best-hyperparameters}}. This holds especially for SAKT.

When considering one-hot input and embedding layer, the differences in performance again depend on the model and the dataset, although the differences are often slight. There are also a few exceptions, however, as shown in Tables \ref{tbl:introprog-model-one-hot-input} and \ref{tbl:assistments-2009-updated-model-one-hot-input} \red{in Appendix~\ref{appendix:best-hyperparameters}}. In the Statics dataset (Table \ref{tbl:introprog-model-one-hot-input} \red{in Appendix~\ref{appendix:best-hyperparameters}}), where the differences are subtle, only SAKT performs worse when using one-hot embedding (0.8\% point difference), while DKVMN-Paper performs marginally better when using one-hot embedding (0.7\% points). On the other hand, in the ASSISTments 2009 Updated dataset (Table \ref{tbl:assistments-2009-updated-model-one-hot-input} \red{in Appendix~\ref{appendix:best-hyperparameters}}), DKVMN-Paper, and SAKT perform significantly worse when using one-hot embedding (3.3\% points and 4.6\% points, respectively), and Vanilla-DKT performs slightly worse (0.3\% points). For the other models, the differences are non-existent. This illustrates that for some models such as SAKT, the effect of the input encoding is influenced by the data. 

However, with one-hot inputs, the models often achieve higher minimum scores compared to using embedding layers. In the ASSISTments 2009 Updated dataset, we see that LSTM-DKT and LSTM-DKT-S+ achieve much higher minimum scores with one-hot inputs. And similarly, Vanilla-DKT and SAKT have higher minimum scores for IntroProg dataset. This suggests that while using embedding layers appears to be the go-to choice when seeking maximal performance, using one-hot inputs can be a safer choice as they seem to be more robust regarding bad choice of hyperparameters.

\subsubsection{Maximum attempt count}

We considered the maximum attempt count used in original DKVMN and SAKT studies in three ways.
The first approach was to split the input data into multiple students if a student's attempt count exceeded the maximum attempt count, the second approach was to discard the excessive attempts, and the third approach was to not use a maximum attempt count. We test the first two approaches with maximum attempt count 200 and 500. This analysis was conducted only for the best hyperparameters selected using grid-search as outlined in~\ref{subsubsec:hyper-summary-model-training}. The comparison results are shown in Table \ref{tbl:split-vs-cut} in Appendix \ref{appendix:max-attempt-count}.

When averaging over all the datasets, shown in Table~\ref{tbl:split-vs-cut-avg} \red{in Appendix~\ref{appendix:max-attempt-count}}, the results suggest that -- on average -- most models benefit from using no maximum attempt count. SAKT and DKVMN-Paper are the exceptions that are appear unhindered by the use of maximum attempt count. The average differences between using and not using a maximum attempt count are subtle for all the DLKT models, however.

Upon inspection of the results for individual datasets, we find different patterns. In the ASSISTments 2015 dataset and in the Synthetic datasets (both K2 and K5), the differences are negligible to non-existent. In the case of the Synthetic datasets, this is explainable with the data having fewer attempts than the used maximum attempt counts. In other datasets, some noticeable differences can be observed.
As an example, the largest difference in the ASSISTments 2009 dataset for DKVMN-Paper is 1.8\% points in AUC in favor of using maximum attempt count (both cut 500 and split 200 perform similarly).
For the ASSISTments 2017 dataset, the largest difference can be observed for LSTM-DKT, where there is a 2.7\% point difference in AUC in favor of not using a maximum attempt count (when compared to cut 200). However, the drop in AUC from no maximum attempt count to the best split result (200) and best cut result (500) is 1.0\% and 1.1\% points respectively.
Similarly, for Vanilla-DKT in the Statics dataset, there is a 3.1\% point difference in AUC in favor of not using a maximum attempt count (when compared to cut 500). For no maximum attempt count versus best split (200), the difference is much smaller (0.7\% points) but quite large (2.4\% points) versus the best cut (200).

Notable differences are present also in other metrics, for instance 3.9\% point difference in MCC for both DKVMN and LSMT-DKT in IntroProg (no maximum attempt count vs split 200). Furthermore, all the metrics do not show a certain approach consistently as the best. As an example, when considering the Statics dataset, the best performing option for DKVMN in terms of AUC or MCC is not using a maximum attempt count, while the best performing option for DKVMN in terms of Accuracy, F1-Score or RMSE is using cut 200.
Averaged over all datasets, the metrics seem to tell a similar story as AUC. Although, there are slight differences, e.g.\@ DKVMN is best without maximum attempt count according to all the metrics apart from F1-Score (precision and recall excluded).

\subsubsection{Random seed, Hardware and Software Version}

As the random seed and the used hardware can also influence the performance of the models, we included two random seeds and two hardware types into the evaluation. As per software version matching challenges for GPU computation within the computing resources at our disposal (and also to examine potential effect of changing the machine learning framework version), we resorted to using TensorFlow 2.6.2 for the GPU calculations (instead of TensorFlow 2.1.0 that we used for CPU computation for the other analyses). To take this into account, when comparing the GPU and CPU, we evaluated the models on CPU with both TensorFlow 2.1.0 and 2.6.2. This evaluation is conducted as a control to which we can compare the results from tuning the other hyperparameters that are more directly related to the models themselves.  
 
Regarding random seed, the effects from changing the random seed are negligible (up to 0.1\% points change) when looking at the best models. However, more considerable differences between the random seeds can be observed for the results with suboptimal hyperparameters. For instance, for SAKT with the ASSISTments 2015 dataset (Table \ref{tbl:assistments-2015-model-seed} \red{in Appendix~\ref{appendix:best-hyperparameters}}), the difference for the worst models is 1.8\% points. In effect, this indicates that the effect of the random seed is larger for non-optimal hyperparameters, while the better models are more robust to the effect of the random seed. Moreover, when considering the standard deviation of the models' performance, there are noticeable differences between the models. As an example, the standard deviation of the models' performance when using different random seeds is noticeably higher for SAKT than for other models. 

To consider the effect of hardware, we retrained the models on GPU with their optimal hyperparameters according to the grid search results for the AUC metric obtained on CPU trained models. The GPU versus CPU results are shown in Table \ref{tbl:cpu-vs-gpu} which can be found in the Appendix \ref{appendix:cpu-vs-gpu}. As noted previously, the GPU results were obtained using TensorFlow 2.6.2 as opposed to 2.1.0, which was used for the grid search. Due to this, in this evaluation, we also retrained the models with the best hyperparameters on CPU using TensorFlow 2.6.2. 

Mostly, the differences between the two computation unit types are again subtle, ranging from 0.0 to 0.6\% points (e.g.\@ Statics dataset LSTM-DKT-S+ or SAKT in MCC score), similar to the random seed comparison. The differences are more often close to zero than above 0.2\% points, especially for accuracy and RMSE.
In general, we find slightly more differences between the TensorFlow versions on CPU than between CPU and GPU.
The largest overall difference we find is 1.7\% points (CPU-tf2.1.0 vs CPU-tf2.6.2) in F1 score in the ASSISTments 2017 dataset for Vanilla DKT. With the same TensorFlow versions the difference is only 0.1\% points, when comparing CPU and GPU.
For the same dataset and model, the AUC and MCC differences, 0.6\% points and 0.9\% points, are also notable.

\subsection{Model Performance and Previously Published Results \label{subsec:comparison-with-previously-published-results}}

Here, we outline our results in light of previously published results. Tables \ref{tbl:paper-comp-ass09} (ASSISTments 2009 / ASSISTments 2009 updated) and \ref{tbl:paper-comp-stat11} (Statics 2011)  summarize observed model performances in this and prior work:~\cite{Piech2015} (DKT),~\cite{zhang2017dynamic} (DKVMN), ~\cite{p2019selfattentive} (SAKT),~\cite{gervet2020deep} (GLR). The tables include AUC scores and use two datasets that are common between the articles. In addition, results from models evaluated in this study are included from~\cite{yeung2018addressing} (DKT+) into the tables, although our present evaluation does not include LSTM-DKT+.

The results presented in this article agree with the previously reported results to some extent. When considering the standard deviation of the model-specific results between the articles, we note that there are considerable differences. For example, for SAKT, the standard deviation is 4.61\% points in the ASSISTments 2009 Updated dataset, and 2.47\% points in the Statics 2011 dataset. Similarly, for LSTM-DKT, the standard deviation is 2.69\% points in the ASSISTments 2009 Updated dataset, and 1.03\% points in the Statics 2011 dataset. For DKVMN, the standard deviations are 0.4\% points and 0.75\% points for the ASSISTments 2009 Updated dataset and the Statics 2011 dataset, respectively. We also observe a minor difference (0.4\% points) in Statics 2011 dataset for GLR but a large difference (4.3\% points) in ASSISTments 2009 Updated dataset.

\begin{table}[ht!]
\begin{threeparttable}
\footnotesize
\centering
\caption{AUC score matrix for models trained on the ASSISTments2009 (Updated) dataset reported on previous articles and this article, labeled as \emph{this}.
}
\begin{tabular}{l|rlllll}
\toprule
  Article  & LSTM-DKT &    LSTM-DKT+ &  DKVMN &   SAKT &    BKT & GLR \\
    \midrule
            DKT* &  86 &       - &      - &      - &   67 & - \\
           DKT+ &  82.212 &  82.227 &      - &      - &      - & - \\
          DKVMN &  80.53 &       - &  81.57 &      - &      - & -\\
           SAKT &  82.0 &   82.2 &   81.6 &   84.8 &      - & - \\
           GLR  &  75.7 &     - &     - & 75.6 &      - & 77.2 \\
          \emph{this} & 81.4 & - &  80.9 & 79.8 &  71.0  & 72.9 \\
    \midrule
         avg (sd)  & 81.3 (3.33) & 82.2 (0.02) & 81.4 (0.40) & 80.1 (4.61) & 69.0 (2.83) & 75.1 (3.04)\\
\bottomrule
\end{tabular}
\label{tbl:paper-comp-ass09}
\begin{tablenotes}
\item A row contains an article identifier (specified in Table \ref{tbl:models}) and the best AUC scores for models reported in that article. Note that the high variance is partly explained by differences in the data used in training the models in the different studies. Most notably, the scores in the DKT article, noted with an asterisk, stems from using an older version of the data. Note also that our result for DKVMN in this table differs from our best results, as the results for original output layer is used here for replication purposes.
\end{tablenotes}
\end{threeparttable}
\end{table}

\begin{table}[ht!]
\begin{threeparttable}
\footnotesize
\centering
\caption{AUC score matrix models trained on the Statics 2011 dataset reported on deep learning model papers and this article, labeled as \emph{this}.}
\begin{tabular}{l|rrrrrr}
\toprule
Article & LSTM-DKT & LSTM-DKT+ & DKVMN & SAKT & BKT & GLR \\
\midrule
          DKT+ &     81.59 &      83.49 &    - &   - &   - & -\\
         DKVMN &     80.20 &        - &  82.84 &   - &   - & - \\
          SAKT &     81.5 &      83.5 &  81.4 &  85.3 &   - & - \\
          GLR &     81.5 &      - &      - &  81.3 &   - & 81.9 \\
         \emph{this} &  82.4 & - & 82.5 & 80.8 &  77.4 & 81.5 \\
    \midrule
          avg (sd)      & 81.7 (1.03) & 83.5 (0.01) & 82.25 (0.75) & 82.5 (2.47) & 77.4 (-) & 81.7 (0.28) \\
  \bottomrule
\end{tabular}
\label{tbl:paper-comp-stat11}
\begin{tablenotes}
\item A row contains an article name and best AUC scores for models reported in that article.
\end{tablenotes}
\end{threeparttable}
\end{table}

In addition to the results shown in  Tables \ref{tbl:paper-comp-ass09} (ASSISTments 2009 / ASSISTments 2009 updated) and \ref{tbl:paper-comp-stat11}, we briefly discuss the results from other datasets (all results are available in the online repository).  

When considering DKVMN results, we observe that the original article shows that DKVMN outperforms LSTM-DKT, mostly by a few AUC percentage points and on all four datasets that the models were evaluated on. In contrast, in our study when including the hyperparameter optimizations and model variations (shown in Appendix~\ref{appendix:best-hyperparameters} in Tables~\ref{tbl:best-hyperparams-assistments-2009-updated},~\ref{tbl:best-hyperparams-assistments-2015},~\ref{tbl:best-hyperparams-assistments-2017},~\ref{tbl:best-hyperparams-introprog},~\ref{tbl:best-hyperparams-statics},~\ref{tbl:best-hyperparams-synthetic-k2}, and~\ref{tbl:best-hyperparams-synthetic-k5}), DKVMN falls behind LSTM-DKT on the ASSISTments 2009 Updated, ASSISTments 2015, and ASSISTments 2017 datasets.
Conversely, on Synthetic-5 DKVMN is the slightly better performing model while in the original study DKVMN falls short of LSTM-DKT by 2.4\% points. 

For SAKT, we observe worse performance than reported in the original article, similar to the Gervet et al~\cite{gervet2020deep} study. While the SAKT article reports considerable improvements on both LSTM-DKT and DKVMN on Statics, Assistments 2009, 2015 and 2017 datasets, in our results, SAKT has worse performance than either DKVMN or LSTM-DKT on all but the IntroProg (not present in original) and the Synthetic datasets (only K-5 in original), where the differences are small to negligible.

We acknowledge that the tested hyperparameter combinations, although at times not reported in the articles, might differ between the studies, which can explain some of the observed larger differences. Some differences can also be attributed to variations in used data. It seems that the data preprocessing differs between the studies as the final used dataset sizes differ at times, as seen in Table \ref{tbl:paper-comp-dataset-sizes}. For example, while both we and Gervet et al.~\cite{gervet2020deep} used the ASSISTments 2009 Updated dataset, the data in the repository for the GLR article was considerably smaller (14\% fewer lines) than the dataset at our disposal. In addition, we also observe that the DKT+ article has a typo in the Statics dataset size (189,927), where the correct dataset size is 189,297. The same typo appears also in the SAKT article.

\begin{table}[ht!]
\footnotesize
\centering
\caption{Number of interactions used in different datasets in our article result comparisons as reported in the articles.}
\begin{tabular}{l|rrr}
\toprule
	& ASSISTments 2009 Updated &	ASSISTments 2015 &	Statics \\
\midrule

DKT+ & 328,291        & 708,631  & 189,927 \\
DKVMN & 325,637	      & 683,801  & 189,297 \\
SAKT & 328k           & 708,631  & 189,927 \\
GLR & 278,868       & 658,887  & 189,297 \\
\emph{this} & 325,515 & 683,331  & 189,297\\
\bottomrule
\end{tabular}
\label{tbl:paper-comp-dataset-sizes}
\end{table}

The differences between interaction counts 708,631 and 683,801 in ASSISTments 2015 in DKT+ and SAKT when compared to DKVMN is due to the preprocessing step explained in DKVMN article~\cite{zhang2017dynamic}, where correctness values other than 0 or 1 are excluded. The data in our work has slightly fewer interactions than that of DKVMN due to our exclusion of students with less than two attempts.

Note that when downloading the dataset from the ASSISTments site\footnote{\url{https://sites.google.com/site/assistmentsdata/home/assistment-2009-2010-data/skill-builder-data-2009-2010}, accessed 2021-04-01}, the only dataset that is openly accessible is the corrected and collapsed skill builder dataset. In collapsed dataset, the number of interactions is 346,860, which does not directly match any of the datasets used in the studies outlined in Table~\ref{tbl:paper-comp-dataset-sizes}. This could be due to small differences in preprocessing between the studies. Furthermore, we acknowledge that some of the publicly available datasets may have changed over time -- as an example, the ASSISTments site and the available datasets have been updated even during the present work\footnote{e.g.\@ comparing the present -- 2021/12/15 -- version of the ASSISTments site with \url{https://sites.google.com/site/assistmentsdata/home/assistment-2009-2010-data/skill-builder-data-2009-2010}, accessed 2021-04-01}.

\section{Discussion}
\label{sec:discussion}



\subsection{Evaluation results - no silver bullet}
\label{subsec:evaluation}

Overall, when comparing the model performance, no single model consistently outperformed all the other models in all the metrics. The best performance is observed for the DLKT models, especially LSTM-DKT and DKVMN. The GLR also performed on par with the DLKT models in some datasets, while falling behind in others. The performance of BKT is typically better than the naive baseline models (mean and variants of next as previous), but worse than DLKT models and GLR. We acknowledge, however, that our BKT implementation is not heavily tuned. In addition, we point out that the naive baselines included in this study performed relatively well in one of the datasets, which showcases the benefit of including simple baselines when reporting results of more complex models. In ASSISTments 2015, the F1-score of Mean baseline model is the same as for the best performing model which is GLR. This indicates that in such cases, the usefulness of the best models can be questionable even if they outperform other complex models when they can not significantly outperform extremely simple statistical models. \red{The observed high performance of the naive baselines in ASSISTments 2015 is possibly due to high skew in combination with relatively low number of attempts per student.}


When considering the evolution of the knowledge tracing field, our evaluation and recent other evaluations of DLKT models~\cite{gervet2020deep,mandalapu2021we,pandey2021empirical} show that the introduction of DLKT to the field has clearly advanced the performance of knowledge tracing.
The major contributing factor seems to be the use of deep learning methodologies in general, and recent work that has focused on further exploration of deep learning for knowledge tracing.
Thus, the move from simpler models to deep learning models has shown robust and verified improvements in knowledge tracing performance.
Multiple more recent approaches appear promising~\cite{Nakagawa:2019:GKT:3350546.3352513,liu2019ekt,ghosh2020context,choi2020towards,cheng2020domain,pandey2020rkt,oya2021lstm,shin2021saint+,song2021jkt}, and many claim significant performance improvements, but their results still require verification via replication studies.

Many of these newer models include slightly different inputs, such as skill and previous correctness as separate inputs~\cite{choi2020towards}, additional time related inputs~\cite{shin2021saint+} or leveraging both exercise and skill labels~\cite{song2021jkt}, giving rise to the question of whether and how much the older architectures would also benefit from such input additions.
We agree with this direction of exploring new inputs since DLKT models are powerful models that are likely to benefit from such additional information. 
Also, similar input modification can be seen in the move from DKT~\cite{Piech2015} to models that incorporate next skill id as additional input as in DKVMN~\cite{zhang2017dynamic} and SAKT~\cite{p2019selfattentive}, although this addition does not appear to provide performance boost as evidenced by our experiments.

\red{As a minor note, in our experiments, SAKT did not live up to the expectations laid out in the original article where the model was introduced. This is in line with prior results by Gervet et al.~\cite{gervet2020deep} where SAKT also underperformed compared to the original article~\cite{p2019selfattentive}. However, newer studies that continued on the path started by SAKT of using self-attentive models have shown promising results: for example, RKT~\cite{pandey2020rkt}, AKT~\cite{ghosh2020context} and SAINT+~\cite{shin2021saint+}.}


Creating new models by introducing new types of inputs is a direction that has been witnessed also earlier in the KT field.
As an example, Performance Factors Analysis~\cite{PavlikJr2009} is an improvement over Learning Factors Analysis~\cite{Cen2006} that includes student ability as a separate input, achieving better performance.
Similarly, there is a wide variety ways how BKT models can be improved by adding new inputs~\cite{khajah2016deep}, e.g., adding item difficulty as an input~\cite{pardos2011kt}. Effectively, as suggested by Khajah et al~\cite{khajah2016deep}, deep learning models can leverage regularities in the data that prior models cannot without adding such capability through explicitly added input features.
As the deep learning models themselves are capable of performing feature engineering~\cite{lecun2015deep}, adding new inputs can provide them an easier starting point for this process. 

On the other hand, one of the challenges of the introduction of DLKT models is the decrease in interpretability. As deep learning models are complex layered structures -- effectively partially black boxes -- the intricacies of the models are not easy to understand~\cite{molnar2020interpretable}. During training, connections between neurons in the layers are re-weighted and re-evaluated to optimize the output. This effectively means that there is a vast amount of trainable parameters, i.e., weights. For example, most of the models evaluated in this study have tens of thousands of weights that are updated during training. 
This leads to difficulty in for instance interpreting the effect of individual input features on the model outputs. Possible remedies for this include feature visualization, concept detection and finding influential instances~\cite{molnar2020interpretable}, although these often require forming hypotheses of what might work and what might not work and they remain heuristics for feature importance.

In contrast, when considering BKT or GLR, interpreting the results is more straightforward. For example, manual fine tuning of BKT through introduction of parameters can lead to performance comparable with DLKT models~\cite{khajah2016deep}, in addition to the state of being able to understand how the parameters are used. Similarly, when using GLR, ranking the importance of the features is straightforward. This has implications as understanding the models helps us to understand why some models perform better than others, as well as to understand -- for example -- contextual factors that contribute to the performance.

Due to this, DLKT models can lead to a disconnect between the use of learning theories and knowledge tracing. One could even see a link between using deep learning for knowledge tracing and using machine learning for natural language processing, where one of the famous quotes is ``Every time I fire a linguist, the performance of the speech recognizer goes up'' (Frederick Jelinek in the 1980s).

Key takeaways:

\begin{itemize}
    \item No single model consistently outperforms all other models in all metrics. In our evaluation results, DKVMN and LSTM-DKT are most often ranked at the top and have the best performance on average. 
    \item DLKT models in general come with an increase in model performance, at the cost of model interpretability.
    New models seem to emphasize leveraging inputs differently or adding additional inputs. 
    \item Naive baseline models that are easy to implement and interpret -- that is, models with little to no predictive power -- can help understand the relative performance of more complex models.
    \item A prominent area of improving the performance of DLKT models is introducing different approaches to processing input and providing new input features. 
\end{itemize}

\subsection{Metrics in reporting and training - not just AUC}
\label{subsec:metrics}

A multitude of metrics for model evaluation exist and relying on a single metric can easily lead to misguided judgement of performance~\cite{national2005thinking}. The usefulness of one metric over another depends on the task at hand,
and as Gunawardana et al.~\cite{gunawardana2009survey} state, ``The decision on the proper evaluation metric is often critical, as each metric may favor a different algorithm''. For instance, in identifying at-risk students, recall can be considered preferable over precision, since it can be argued that finding struggling students is preferable over finding non-struggling students. Misidentifying a well-performing student as a struggling student is not as costly as vice-versa, as the downside is that a well-performing student might be offered additional support that they do not need. Thus, in this case, tuning precision and recall in favor of recall might be more favorable than tuning  precision and recall in favor of precision or with equal weights.

In knowledge tracing, we are not predicting dropping a course but rather single attempts at exercises, which makes the weighing of false positives and false negatives less clear. Although, without delving deeper into the matter, when considering the case of early intervention to help struggling students, similarly to retention prediction, it might be better to intervene more often than not. But the interventions still need to happen accurately enough to keep them valuable in the minds of learners.

DLKT studies often compare the performance of models using the AUC (ROC-AUC) metric which is also popular in other domains (e.g.\@ medicine~\cite{kim2017development,huang2019patient} and natural language processing~\cite{pahikkala2009matrix}). How AUC compares to accuracy, another popular metric, has drawn a lot of attention. Some formal and quantitative studies have shown AUC to be consistent with and more discriminative than accuracy~\cite{ling2003auc,huang2005using,halimu2019empirical}, and thus it has been claimed as superior. This, however, does not show that accuracy is worse than AUC, merely that it is \red{less} likely than AUC to show differences between the models. There is no guarantee that better AUC translates to better model~\cite{jeni2013facing,ozenne2015precision,dhanani2014comparison}.

An often heard critique of accuracy (and also F1 metric), as opposed to e.g.\@ AUC or MCC, is that high accuracy and F1 score can be a product of skewed data where positive labels outnumber negative labels. In such a case, the model may only predict positive labels well~\cite{chicco2020advantages} and receive seemingly good metric scores, while still performing poorly with the minority class. AUC solves this issue by effectively accounting for skew in data. However, AUC with skewed data has also been argued to be a poor combination, since, in AUC, the minority class has the same impact on the metric score as the majority class~\cite{ferri2009experimental}. Indeed, there is evidence that AUC may mask poor model performance~\cite{jeni2013facing,ozenne2015precision} and that AUC is less suitable than RMSE, for instance for BKT model evaluation~\cite{dhanani2014comparison}.

\red{One remedy to problems in AUC could be the visualization of the whole curve instead of reporting merely the area under it. However, drawing conclusions from the visualized curve also becomes more difficult at the same time, since comparing curves is not as straightforward as numbers.}

We argue that if we provide model accuracy and F1 scores alongside mean prediction with equally high scores, it easily breaks the illusion that a model with high accuracy is good. Accuracy and F1 scores can be good metrics with little risk of misinterpretation even on highly skewed datasets when mean or majority vote baseline is presented as comparison.

This leads to a dilemma in choosing the metrics to report and to determine model rankings. As an example of the metric choice dilemma, in our results for the ASSISTments 2015 dataset, DLKT models hold the best AUC and MCC scores by a significant margin. On the other hand, both the GLR and BKT models achieve almost the same performance on most other metrics. Also, the simple mean baseline is not far from the DLKT models in terms of accuracy, and the mean baseline holds the best F1-score tied with GLR. The DLKT models outperform GLR and BKT only on MCC and AUC which are the metrics often presented as alternatives to the ``misleading'' accuracy and other metrics that can be influenced by data skew.

\red{In a related study, Effenberger et al.~\cite{effenberger2020impact} evaluated the metrics MAE (mean absolute error) and RMSE (root mean squared error) in detail for student modeling. Similarly to our case, they reported cases where the choice of metric affected model ranking and also drew attention to the possibility of picking a ``suitable'' metric for a newly proposed model to make it appear better in comparison to previous models. They also showed that besides metric choice, the computation methodology (RMSE over whole data vs average over RMSEs per student data) of the metric may also affect model ranking. }

Choosing a metric is not solely a problem in reporting, but also in hyperparameter tuning. When considering which metric to use when selecting the best hyperparameters for models via grid search, we observed that different hyperparameters may be chosen as the best when the metric used to choose them is changed. Consequently, the model with best hyperparameters according to one metric may not
be optimal when considering other metrics. This problem is also noted by Sanyal et al.~\cite{sanyal2020feature}, who inspected how optimizing feature selection on different metrics influenced model performance and observed that metric selection can lead to large performance differences between datasets and selected features. 

This further highlights the importance of choosing and understanding metrics for model comparison.

With scant rigorously studied information on evaluation metrics for knowledge tracing, and especially for DLKT, based on previous studies and our results, we suggest providing multiple metrics for evaluation. At least one unaffected by skew (e.g.\@ AUC, MCC) and one affected by skew (e.g.\@ Accuracy, F1-score) as proposed by \cite{jeni2013facing}. In addition a generic error metric, such as RMSE, should be included for reliability, which is also suggested in \cite{liu2011measuring}. Providing Area Under Precision-Recall Curve (AUC-PC), which unlike AUC, is affected by skew but similarly to AUC is not affected by a decision threshold, could be a potential main metric to replace AUC since it has been shown to mask poor performance less than AUC~\cite{ozenne2015precision,saito2015precision}.

Key takeaways:

\begin{itemize}
    \item Even though AUC is one of the most widely used metrics for evaluating model performance, with skewed data, it can mask poor model performance.
    \item Relying on a single metric in model evaluations can lead to misinformed decisions on model quality. Metrics that account and do not account for data imbalance should be used in model evaluations.
    \item The model that receives the best scores on a certain metric does not necessarily achieve the best results for other metrics. Thus, the metric that is used to determine the best model or best model hyperparameters is important to choose well, as well as to report for transparency.
\end{itemize}

\subsection{Hyperparameters and architecture variations - mind your randomness}
\label{subsec:hyperparameters}

Overall, hyperparameter tuning had a significant impact on model performance. This is to be expected as hyperparameter tuning heavily affects both overall model complexity (e.g.\@ layer sizes affect the number of trainable parameters in a model) and model training itself (e.g.\@ learning rate). Even though we explored a relatively small hyperparameter space (2 options per hyperparameter, as shown in Table~\ref{tbl:hypersummary}), the differences between the worst and best hyperparameter combinations can be over 20\% points (see e.g.\@ SAKT in Table~\ref{tbl:introprog-model-one-hot-input}). The relative impact of hyperparameter tuning on the model performance depended on the model, as some models were less susceptible to their hyperparameters (see e.g.\@ Tables~\ref{tbl:assistments-2017-model-output-per-skill}-\ref{tbl:introprog-model-one-hot-input}). For instance, DKVMN is relatively stable over the hyperparameter combinations, while e.g., Vanilla-DKT is much less so. 

In our exploration of the effect of different input (one-hotting vs. use of embedding layer) and output (skill summary layer and \red{skill-to-scalar} output vs. output per skill) variations for DLKT models, we identify two main findings.
First, while for many datasets and models there are mostly no differences in performance when using one-hot embedding when compared to using embedding layers.
There are some exceptions, where using embedding layers considerably outperforms one-hotting up to 4.6\% point difference (for SAKT in the ASSISTments 2009 Updated dataset in Table~\ref{tbl:assistments-2009-updated-model-one-hot-input}). This leaves little reason to one-hot inputs as opposed to using embedding layers, especially as one-hotting can cause memory problems when training models if the one-hot embeddings become large due to high number of skills in data. On the other hand, while embedding layers appear to be the go-to choice when seeking maximal performance, using one-hot inputs may be a safer choice as they seem to be more robust regarding bad choice of hyperparameters in some datasets. Second, when considering output variations, using the more recent output version (skill layer and \red{skills-to-scalar} output layers) showed more variance in model performance compared to the output per skill layer used in the first DLKT model DKT. We did not find indications that one would consistently be better than the other.

We also considered the impact of a maximum attempt count that has been incorporated in both DKVMN and SAKT (although the DKVMN article does not mention this). Both DKVMN and SAKT split student attempt sequence by a maximum attempt count, but neither article discussed this to an extent or analyzed the effect of the approach. To better understand how using a maximum attempt count affects KT model performance, we analyzed the effect of using maximum attempt count to split the student assignment sequences into smaller chunks thus artificially increasing the number of total students. This analysis was conducted by re-training our models using the best hyperparameters according to our grid search tuning with an additional hyperparameter: maximum attempt count filter method (none vs cut vs split). None indicates that no maximum attempt count was applied, cut indicates that the attempts after the maximum attempt count were discarded, and split effectively divides sequences longer than the maximum attempt count into multiple sequences that have at most maximum attempt count as their length. When testing the effect of the maximum attempt count, we used 200 and 500 as the choices for maximum attempt count.

Overall, as shown in Appendix~\ref{appendix:max-attempt-count}, we found significant differences (e.g.\@ up to 3.1\% points AUC) in model performances depending on the filter method and some differences between the maximum attempt count 200 and 500, although not in all models in all datasets. For instance in the Statics dataset, no model is unaffected by the additional hyperparameter tuning. The Synthetic datasets were not influenced by the filter as the attempt sequences are shorter than 200.
These results showcase that maximum attempt count should not be overlooked when tuning hyperparameters and comparing knowledge tracing models. This is further emphasized by SAKT and Vanilla-DKT appearing to benefit from applying the maximum attempt count (split or cut) in comparison to other models. The other models perform better with no maximum attempt count while SAKT and Vanilla-DKT are on average much less affected. In other words, tuning maximum attempt count can bias evaluation in favor of certain models and this is something that should be accounted for and discussed. We do not however, suggest that using maximum attempt count should be completely discouraged as it does speed up model training and reduce model space requirements, although there appears to be little other benefit according to our evaluation.

To quantify the effect of variation in performance due to properties not related to models, we also explored the effect of random seed and hardware (CPU vs GPU).
First, we found that the values used as random seeds have very slight effects on model performance (see Table~\ref{tbl:assistments-2015-model-seed}), although the effects are considerable on poor choice of hyperparameters. This indicates that the danger of randomness affecting model performance is present but mitigated by a large hyperparameter space for tuning.
Second, we found mostly little to no differences in model performance depending on the hardware that was used to run the model (CPU or GPU). Some differences are larger, however. The biggest is 0.6\% points difference for MCC and F1-Score, while the biggest difference in AUC is merely 0.3\% points.
On the other hand, we observed a tad more and larger differences between TensorFlow versions, with the largest being 1.7\% points in F1-Score and the largest AUC score difference is 0.6\% points. Thus, although unlikely based on our results, it is possible to have large performance difference when tuning hyperparameters on one machine learning framework version and then using the hyperparameters on another version another.
Consequently, one should not take granted that the tuned hyperparameters work the same in another setting.

Even though the performance differences between the best models are usually negligible (e.g.\@ 0.1\% points AUC) when considering one of these non-model properties, the combination of the effects of hardware, seed, and hyperparameter tuning could easily change the ranking of the best performing models.
This suggests that minor improvements in performance could be due to random chance, and thus to regard some model as the new ``state-of-the-art'', the improvement in e.g.\@ AUC scores should be considerable and consistent across multiple datasets.
Based on our results, we would be careful to consider even a 1\% point increase in performance as a true improvement in terms of model architecture unless such an improvement was repeated in multiple studies and contexts.

Key takeaways:

\begin{itemize}
    \item Hyperparameter tuning has a significant impact on model performance. Our results indicate that optimal hyperparameters are model- and dataset-specific.
    \item Input and output variations, which we used as hyperparameters but which could be pre- and post-processing steps, also impact model performance.
    \item Maximum attempt count filtering strategy influences model performance, and its effect depends on used model and dataset.
    \item Randomness (e.g.\@ random seed, hardware) and machine learning framework version can affect model performance, although the observed impact in our evaluations is mostly very slight. 
\end{itemize}

\subsection{Data - no one model to rule them all}
\label{subsec:data}

Overall, some models appear to be more capable of benefiting from big data than others, and the dataset and its underlying distributions has a clear impact on model performance. Data size is certainly a factor in DLKT model generalizability across datasets due to deep learning models' symbiotic relation with big data that follows from the models' tendency to overfit. In our results, we noticed no clear pattern with data size affecting model performance, and no model outperformed all the other models in all datasets. GLR and BKT fare relatively well compared to DLKT models in the ASSISTments 2015 dataset with the most students (19917) and the second most attempts (683k) but poorly in the ASSISTments 2017 dataset with the most attempts (943k). Also the SAKT model performs on par with the other DLKT models only in the IntroProg dataset which has the least attempts (172k) our datasets.

Other studies have come to different conclusions, which highlights that the impact of the data is still an open question. For example, Mandalapu et al. \cite{mandalapu2021we} note that SAKT performs better than LSTM-DKT in larger datasets. In the same study though, a non-deep-learning logistic regression model beat both SAKT and LSTM-DKT even when supposedly deep learning excels on large data. Apart from data quantity, the underlying distribution of the data has a great influence on the performance of the models. Gervet et al. \cite{gervet2020deep} suggest that the number of learners per learning item or knowledge component (KC) is more important than the total number of interactions, and that some models benefit more from large amount of training data than others.

Similar to other studies, we also used synthetic (Synthetic-K5 and Synthetic-K2) datasets in our study. One interesting phenomenon we noticed (see Figure~\ref{fig:auc-data-over-modeller}) is that the AUCs for the synthetic datasets seem to be more stable between different deep knowledge tracing models, i.e.\@ the differences in AUC and other metrics are smaller, but still large compared to logistic regression and other baselines.

Our present work followed the methodology most commonly used in knowledge tracing studies where data from all students is used for training the models (accounting for train/test validation splits etc.). Some prior work has however suggested that one way to improve model performance would be to use a subset of the available data~\cite{faraway2018small} -- for example only a part of the students -- to train the models, since some students seem to produce higher quality data than others~\cite{alexandron2019mooc,yudelson2014better}, which can lead to model performance improvement over all students~\cite{yudelson2014better}. This topic should also be further explored.

On a more general level, evaluations of knowledge tracing models typically rely on individual datasets. There are options in the machine learning domain that could potentially be used to better benefit from existing data; as an example, one could benefit from the use of domain adaptation and transfer learning techniques that have been successfully used in, for example, natural language processing~\cite{zhuang2020comprehensive}. Some work already exists in adapting these methodologies for knowledge tracing~\cite{cheng2020domain}. However, we envision pre-trained knowledge tracing models similar to GPT-3~\cite{brown2020language} that could then be fine-tuned with context-specific data.

Key takeaways:

\begin{itemize}
    \item No single model outperformed all other models on all datasets.
    \item Even the best DLKT models did not always yield superior performance compared to other models in all datasets. The best model and also model type (DLKT vs non-DLKT) is dependent on data.  
    \item When considering the adoption of knowledge tracing, to ensure the best fit for a specific context, evaluate multiple models in that context instead of choosing the most recent state-of-the-art model.
\end{itemize}

\subsection{Replication process and findings - the devil is in the details}
\label{subsec:replication}

One of the key aspects of science is providing sufficient details of the used methodology that allows tracing the steps of the researchers to conduct similar research and to improve on it. Replication studies can be conducted in different ways~\cite{Ihantola2015,patil2016statistical}, where one is seeking to reproduce earlier findings with the same data and the same methodology. In such a case, the objective is to examine the methodology to determine whether the steps are explained clearly enough and to explore whether there are considerations that were omitted from the original study. Earlier studies have suggested, however, that keeping the same data and methodology, but changing the researchers, can already lead to challenges with replication~\cite{Ihantola2015}. Another approach to replication would be seeking to replicate the effect found in the original study with new data and possibly new or improved methodologies. In this case, the objective could be to study whether the data has an impact on the effect, and whether the effect generalizes beyond the original context.


To summarize our findings from the replication process, we found multiple issues, which are as follows. First, we observed differences between a model description in a published article and an associated code repository. Second, we observed methodological differences between a published article and an associated code repository. Third, we identified differences in data set sizes between articles, even though the data sets have been labeled the same and thus also likely understood to be the same. Finally, fourth, in some cases, we were unable to reach similar performance as reported in the original articles.

There are naturally a multitude of explanations for our findings.
As an example, for the first and the second case, it is possible that the code that authors have added to their repository is an earlier version of their work, and does not represent the version reported in the article.
This situation can be problematic however, as others may directly rely on the available implementations, instead of reimplementing the work based on the article.
For the third case, it is possible that there are differences in data preprocessing steps that are not sometimes fully explained in the articles.
We did, however, also observe a case where it is likely that data sizes were originally mistyped in an article and then copy-pasted to another article by different authors.
For the fourth case, it is possible that some methodological steps have been omitted in the article, which could lead to better results.
In this case, however, others have also struggled to replicate the earlier performance.

These findings follow the trend visible also in other fields, where researchers have identified problems with replicating results from prior published studies~\cite{Baker2016,open2015,Ioannidis2005,Moonesinghe2007}. 

There are a lot of positive signals as well. As an example, authors often had placed their research code and used data in repositories for inspection, and there are multiple open datasets that can be used to evaluate model performance. Some authors also reported the hyperparameter variations and preprocessing steps they had used when training the models, and the articles often included evaluations of proposed models against other recently proposed models over evaluating only against simpler baselines that are easy to outperform. One commendable work in this line of work is centralized algorithm and data repositories such as DataShop \cite{koedinger2010data}, which help reproduction, although considering replication, using the same data and methodology can disallow further examination of implementation details, which may lead to overlooking issues in implementation.

Our replication results show that old models can easily surpass newer ones and vice versa given the right hyperparameters. When introducing a new model that outperforms others, it is important to report the used hyperparameter options for both the new and the old models, and to verify that the better results are not due to more rigorous hyperparameter tuning for the introduced model. Similarly, data preprocessing decisions such as splitting student data based on maximum attempt count or omitting data by minimum attempt count can impact model performance, and even non-model specific properties such as hardware has an influence over the results.

Examples of the difficulty in replicating prior work can be easily found by looking at the reported performance of models. Some papers agree on the results of some models on certain datasets and disagree on the results of others. The degree of variance in observed performances is also relatively high, although the biggest differences could be explainable by the use of different versions of data.

In order to replicate machine learning work well, as much information as possible about the original work is beneficial, which is why we would like to emphasize the importance of source code accessibility. When the source code for a work is easy to find and analyze, many aspects of a machine learning model can be viewed that are often missing from a scientific paper. Whether missing pieces of implementation details are due to lack of rigor, estimated importance or a limitation of the publishing platform (e.g.\@ paper length), they might include key components that explain why one model is better than the other. 

This raises an interesting point about replicating machine learning work in general. Different types of expertise are needed to effectively work in the ML domain. On one hand, one needs to have sufficient mathematical skills to understand mathematical notations which are the most common way to communicate new ML models in academic publications. On the other hand, also good skills are required in programming and the frameworks being used in order to understand the source code of the models when those models are available openly. In most works, the mathematical notation of the model present in the publication hopefully matches the source code of the model. When this is not the case, it is not evident whether the researchers replicating the work should follow the mathematical notation (which likely includes the novel aspects of the model) or the source code (which likely was used to compute the results presented in the article). It is a good question whether conference organizers and journal editors should require reviewers to also review the source code and evaluate whether it matches the mathematical notation in the publication. We acknowledge that this can be a highly time-consuming process, however.

Key takeaways:

\begin{itemize}
    \item In our study, we found multiple discrepancies in algorithm and data descriptions, which have the potential to influence study outcomes.
    \item Publishing source code and data is a good practice that should be continued. Preferably, links to code and data should be included in articles for ease of access and transparency. Further, we recommend linking to a specific version of code and data to verify that the linked code is indeed the definitive version used in the study and to allow for further improvements while keeping the connection to the original article.
    \item When publishing studies, include methodological details that allow replication, including data preprocessing steps, evaluated hyperparameter options, and other specifics of model training.
    \item  When presenting a new state-of-the-art model, other evaluated models should be evaluated with the same rigor (e.g.\@ data preprocessing, hyperparameter tuning) as the proposed model to clarify that improvements stem from the proposed model algorithm and not from other factors.
\end{itemize}

\subsection{\red{Limitations of work}}
\label{subsec:limitations}

Here, we summarize the key limitations of this work.
First, we acknowledge that there are a wide variety of knowledge tracing models, including newer ones, that were not included in the empirical evaluation. 
\red{Multiple of these are briefly discussed in section \ref{subsec:related-dlkt-models}.}
Limiting the number of evaluated models was a deliberate choice, as we meticulously reimplemented the algorithms as well as compared and contrasted the available implementations with the details outlined in the respective articles.

Second, while we implemented most of the models compared in this study, for the baseline Bayesian Knowledge Tracing we used Yudelson's implementation that is available on GitHub\footnote{\url{https://github.com/myudelson/hmm-scalable}, accessed 2020-03-01}. Similarly, for the logistic regression, we used the Best-LR (GLR) model with slight modifications by Gervet et al.~\cite{gervet2020deep}. Their model, too, is available on GitHub\footnote{\url{https://github.com/theophilee/learner-performance-prediction}, accessed 2021-05-27}. Thus, our results related to BKT and GLR are not full replications (where the model would be reimplemented from scratch). We decided not to reimplement the models as our focus in this work is on scrutinizing DLKT models, furthermore BKT and GLR were only used as baselines to which we can compare the DLKT models.

Third, we have used seven datasets in this study, six publicly available ones and one novel dataset (IntroProg) which is published alongside this work. Thus, our results are only applicable to these datasets, and do not necessarily extend to other datasets. As an example, many recent studies have included the EdNet dataset~\cite{choi2020ednet} in their evaluations, including the studies by Mandalapu et al.~\cite{mandalapu2021we} and Pandey et al.~\cite{pandey2021empirical}. When compared to the datasets used in this study, the EdNet dataset is larger and thus may allow the models to utilize information in a way that is not possible in the present datasets.

Fourth, we acknowledge that our explored hyperparameter space is not very extensive, and it is possible that further tuning could have realized better gains. All in all, in our study, there are between 72 and 240 hyperparameter variations per model, and in total, we evaluated 5,880 cross-validated DLKT models. With the resources available to us, significantly extending the explored hyperparameter space would have led to far longer model training time.

\red{Fifth, although we explored a range of hyperparameters, we did not look into item-aware input and output, which also could influence the results. For an exploration of this aspect of knowledge tracing models, see e.g. studies by Gervet et al.~\cite{gervet2020deep} and Vie and Kashima~\cite{vie2019knowledge}.}


Finally, we acknowledge that when selecting best hyperparameter values for models, we selected them based on AUC, similarly as is done in e.g.\@ \cite{zhang2017dynamic}. As discussed in Section~\ref{subsec:metric-and-determining-the-best-model}, using another metric to select the best models would have led to slightly different results. We decided against extensive exploration of selection metric when reporting results, as it would have led to more complex reporting (effectively multiplying the reported results by the number of metrics), and as AUC is commonly used as a principal metric in the knowledge tracing community.

\section{Conclusion}
\label{sec:conclusions}

In this article, we reviewed models for knowledge tracing, evaluating the performance of eleven models.
We evaluated deep learning knowledge tracing (DLKT) models (Vanilla-DKT, LSTM-DKT, LSTM-DKT-S+, DKVMN, DKVMN-Paper, SAKT) and baseline models (Mean, Next as Previous, Next as Previous N's Mean, BKT, GLR).
Out of these, LSTM-DKT-S+ is our own variant of LSTM-DKT, which takes next skills into account as inputs. For the DKVMN, the two versions are presented as the version in the article and the version in the repository differed from each other. See Table~\ref{tbl:models} for details of the models.
All models were evaluated against seven datasets using seven metrics.
For deep learning knowledge tracing models, we evaluated the impact of input and output variations (one-hot embedding versus embedding layer; output-per-skill layer versus skill layer and \red{skills-to-scalar} output layers), and also maximum attempt count handling on model performance. The effect of non-model properties such as hardware and random seed were examined as a baseline to which model improvements could be compared to.
The deep learning knowledge tracing models were reimplemented for this study.

The motivation of our study was five-fold.
First, to re-evaluate the proposed models of earlier studies by reimplementing them and comparing them to each other and to simple baselines.
Second, to highlight how the choice of metric used in reporting affect perceived model performance and hyperparameter tuning.
Third, to show how hyperparameters and variations in model input and output architecture can have an effect on models' performance.
Fourth, to explore model performance across different contexts (datasets).
Fifth, to emphasize the importance of replication studies and give pointers on how to make such studies easier, as well as to give pointers to help make results of future KT studies more robust. We also publish our implementations, datasets and evaluation code for use in future research.

To summarize, our research questions and their answers are as follows:

\textbf{RQ1} How do DLKT models compare to naive baselines and non deep-learning KT models? \textbf{Answer:} The evaluated DLKT models generally outperform the baseline models. DLKT models in general outperform the non-DLKT baselines BKT and the logistic regression model GLR. GLR does, however, achieve performance on par with some of the DLKT models on two datasets and is the best performing model on one dataset. BKT is on par with the DLKT models and GLR on one dataset, in all metrics but AUC and MCC, but BKT fares worse on other datasets. Naive baselines mostly perform poorly when compared to the more complex models, but on one dataset, the accuracy of the best DLKT models is not much better than the Mean model, and the Mean model also achieves better F1-Score than the DLKT models. This highlights both the performance of the DLKT models as well as the importance of including naive baseline models into KT model comparisons to verify usefulness of models.

\textbf{RQ2} How do DLKT models perform on the same and different datasets as originally evaluated with? \textbf{Answer:} In our evaluations, LSTM-DKT, LSTM-DKT-S+ and DKVMN had the best performance on average out of the evaluated models, but the differences in performance of any of the DLKT models were not great.
We found that the relative performance of the models depended on the context, i.e.\@ the dataset.
We did not find differences in performance between LSTM-DKT and our variant with additional next skill input LSTM-DKT-S+, but we found that the DKVMN version implemented based on the DKVMN authors' repository performed on average slightly better than the DKVMN-Paper version, which is the version introduced in the article.
We found considerable variance between previously reported results for our evaluated models, especially for SAKT, which is in line with other prior comparison studies.
When comparing the results over different metrics, we observed that the ranking of the models can differ depending on the inspected metric, which highlights the importance of reporting multiple metrics for model evaluation. We also analyzed the effect of using different metrics for hyperparameter tuning and found the same pattern there; the model which receives the best score on one metric does not always receive the best score on another metric.

\textbf{RQ3} What is the impact of variations in architecture and hyperparameters on DLKT models' performance? \textbf{Answer: } Overall, the impact of variations in hyperparameters contributes significantly to model performance. The extent to which model performance depends on hyperparameters could be seen as a quality factor of the models, where more robust models are less dependent on extensive hyperparameter tuning. Explored variations in model architecture, one hot vs embedding layer for input, and output per skill layer vs skill layer and \red{skills-to-scalar} output layers, showed some variation in performance, up to 4.6\% point AUC. Non-model properties, i.e.\@ hardware, machine learning framework version, and random seed, had mostly negligible impact on model performance. Although, we found some cases where machine learning framework version and the hardware led to over 0.5\% point difference in results in various metrics. Maximum attempt count filtering (no filtering, cut, split -- split has been used implicitly and explicitly in some previous work) had also a noticeable impact in model performance, and should be taken into account and reported if used when training models.

As a part of the work, we reimplemented the deep learning knowledge tracing algorithms following the details presented in the respective articles. During the implementation, we observed inconsistencies between algorithm descriptions and implementations, as well as inconsistencies between the reported dataset sizes. As an example, the architecture of the DKVMN differed between the article and the implementation, and the ASSISTments 2015 dataset had nearly 10\% difference in size between the articles likely due to varying preprocessing steps.

Furthermore, we highlight the need to identify the sources for empirical gains, which has been pointed out to be a concern within the machine learning discipline~\cite{lipton2018troubling}: in our evaluations, we observed that some of the previous findings reported in the literature may have more to do with hyperparameter tuning than proposed neural network structures. 

We call out others to also perform similar studies where KT models are evaluated through replication and reimplementation of the models, where the preprocessing of the data and any possible hyperparameter tuning is explicitly stated and performed with the same level of rigor for all compared models.

\section*{Acknowledgements}


We acknowledge the computational resources provided by the Aalto Science-IT project. We are grateful for the grant by the Media Industry Research Foundation of Finland which partially funded this work. We thank the reviewers for their valuable comments that helped improved this manuscript.

\bibliographystyle{jedm-template/acmtrans}

\bibliography{999-references} 

\begin{thebibliography}{}

\bibitem[\protect\citeauthoryear{Abdelrahman and Wang}{Abdelrahman and
  Wang}{2019}]{Abdelrahman2019}
{\sc Abdelrahman, G.} {\sc and} {\sc Wang, Q.} 2019.
\newblock Knowledge tracing with sequential key-value memory networks.
\newblock In {\em Proceedings of the 42nd International ACM SIGIR Conference on
  Research and Development in Information Retrieval - SIGIR'19}. ACM Press, New
  York, New York, USA, 175--184.

\bibitem[\protect\citeauthoryear{Ahadi, Hellas, Ihantola, Korhonen, and
  Petersen}{Ahadi et~al\mbox{.}}{2016}]{Ahadi:2016}
{\sc Ahadi, A.}, {\sc Hellas, A.}, {\sc Ihantola, P.}, {\sc Korhonen, A.}, {\sc
  and} {\sc Petersen, A.} 2016.
\newblock Replication in computing education research: Researcher attitudes and
  experiences.
\newblock In {\em Proceedings of the 16th Koli Calling International Conference
  on Computing Education Research}. Koli Calling ’16. Association for
  Computing Machinery, New York, NY, USA, 2–11.

\bibitem[\protect\citeauthoryear{Alexandron, Yoo, Ruip{\'e}rez-Valiente, Lee,
  and Pritchard}{Alexandron et~al\mbox{.}}{2019}]{alexandron2019mooc}
{\sc Alexandron, G.}, {\sc Yoo, L.~Y.}, {\sc Ruip{\'e}rez-Valiente, J.~A.},
  {\sc Lee, S.}, {\sc and} {\sc Pritchard, D.~E.} 2019.
\newblock Are {MOOC} learning analytics results trustworthy? with fake
  learners, they might not be!
\newblock {\em International journal of artificial intelligence in
  education\/}~{\em 29,\/}~4, 484--506.

\bibitem[\protect\citeauthoryear{Anderson, Bahn{\'\i}k, Barnett-Cowan, Bosco,
  Chandler, Chartier, Cheung, Christopherson, Cordes, Cremata,
  et~al\mbox{.}}{Anderson et~al\mbox{.}}{2016}]{anderson2016}
{\sc Anderson, C.~J.}, {\sc Bahn{\'\i}k, {\v{S}}.}, {\sc Barnett-Cowan, M.},
  {\sc Bosco, F.~A.}, {\sc Chandler, J.}, {\sc Chartier, C.~R.}, {\sc Cheung,
  F.}, {\sc Christopherson, C.~D.}, {\sc Cordes, A.}, {\sc Cremata, E.~J.},
  {\sc et~al\mbox{.}} 2016.
\newblock Response to comment on ``estimating the reproducibility of
  psychological science''.
\newblock {\em Science\/}~{\em 351,\/}~6277, 1037--1037.

\bibitem[\protect\citeauthoryear{Anderson, Boyle, and Reiser}{Anderson
  et~al\mbox{.}}{1985}]{anderson1985intelligent}
{\sc Anderson, J.~R.}, {\sc Boyle, C.~F.}, {\sc and} {\sc Reiser, B.~J.} 1985.
\newblock Intelligent tutoring systems.
\newblock {\em Science\/}~{\em 228,\/}~4698, 456--462.

\bibitem[\protect\citeauthoryear{Asendorpf, Conner, De~Fruyt, De~Houwer,
  Denissen, Fiedler, Fiedler, Funder, Kliegl, Nosek, et~al\mbox{.}}{Asendorpf
  et~al\mbox{.}}{2013}]{asendorpf2013}
{\sc Asendorpf, J.~B.}, {\sc Conner, M.}, {\sc De~Fruyt, F.}, {\sc De~Houwer,
  J.}, {\sc Denissen, J.~J.}, {\sc Fiedler, K.}, {\sc Fiedler, S.}, {\sc
  Funder, D.~C.}, {\sc Kliegl, R.}, {\sc Nosek, B.~A.}, {\sc et~al\mbox{.}}
  2013.
\newblock Recommendations for increasing replicability in psychology.
\newblock {\em European Journal of Personality\/}~{\em 27,\/}~2, 108--119.

\bibitem[\protect\citeauthoryear{Baker}{Baker}{2016}]{Baker2016}
{\sc Baker, M.} 2016.
\newblock Is there a reproducibility crisis?
\newblock {\em Nature\/}~{\em 533}, 452--454.

\bibitem[\protect\citeauthoryear{Baker, Corbett, and Aleven}{Baker
  et~al\mbox{.}}{2008}]{d2008more}
{\sc Baker, R.~S.}, {\sc Corbett, A.~T.}, {\sc and} {\sc Aleven, V.} 2008.
\newblock More accurate student modeling through contextual estimation of slip
  and guess probabilities in bayesian knowledge tracing.
\newblock In {\em International conference on intelligent tutoring systems}.
  Springer, 406--415.

\bibitem[\protect\citeauthoryear{Begley and Ellis}{Begley and
  Ellis}{2012}]{begley2012}
{\sc Begley, C.~G.} {\sc and} {\sc Ellis, L.~M.} 2012.
\newblock Drug development: Raise standards for preclinical cancer research.
\newblock {\em Nature\/}~{\em 483,\/}~7391, 531--533.

\bibitem[\protect\citeauthoryear{Bhandari~Neupane, Neupane, Luo, Yoshida, Sun,
  and Williams}{Bhandari~Neupane
  et~al\mbox{.}}{2019}]{bhandari2019characterization}
{\sc Bhandari~Neupane, J.}, {\sc Neupane, R.~P.}, {\sc Luo, Y.}, {\sc Yoshida,
  W.~Y.}, {\sc Sun, R.}, {\sc and} {\sc Williams, P.~G.} 2019.
\newblock Characterization of leptazolines a--d, polar oxazolines from the
  cyanobacterium leptolyngbya sp., reveals a glitch with the
  “willoughby--hoye” scripts for calculating nmr chemical shifts.
\newblock {\em Organic letters\/}~{\em 21,\/}~20, 8449--8453.

\bibitem[\protect\citeauthoryear{{Bianchini} and {Scarselli}}{{Bianchini} and
  {Scarselli}}{2014}]{bianchi2014shallow}
{\sc {Bianchini}, M.} {\sc and} {\sc {Scarselli}, F.} 2014.
\newblock On the complexity of neural network classifiers: A comparison between
  shallow and deep architectures.
\newblock {\em IEEE Transactions on Neural Networks and Learning
  Systems\/}~{\em 25,\/}~8, 1553--1565.

\bibitem[\protect\citeauthoryear{Bouthillier, Delaunay, Bronzi, Trofimov,
  Nichyporuk, Szeto, Mohammadi~Sepahvand, Raff, Madan, Voleti,
  et~al\mbox{.}}{Bouthillier et~al\mbox{.}}{2021}]{bouthillier2021accounting}
{\sc Bouthillier, X.}, {\sc Delaunay, P.}, {\sc Bronzi, M.}, {\sc Trofimov,
  A.}, {\sc Nichyporuk, B.}, {\sc Szeto, J.}, {\sc Mohammadi~Sepahvand, N.},
  {\sc Raff, E.}, {\sc Madan, K.}, {\sc Voleti, V.}, {\sc et~al\mbox{.}} 2021.
\newblock Accounting for variance in machine learning benchmarks.
\newblock {\em Proceedings of Machine Learning and Systems\/}~{\em 3}.

\bibitem[\protect\citeauthoryear{Brown, Mann, Ryder, Subbiah, Kaplan, Dhariwal,
  Neelakantan, Shyam, Sastry, Askell, et~al\mbox{.}}{Brown
  et~al\mbox{.}}{2020}]{brown2020language}
{\sc Brown, T.~B.}, {\sc Mann, B.}, {\sc Ryder, N.}, {\sc Subbiah, M.}, {\sc
  Kaplan, J.}, {\sc Dhariwal, P.}, {\sc Neelakantan, A.}, {\sc Shyam, P.}, {\sc
  Sastry, G.}, {\sc Askell, A.}, {\sc et~al\mbox{.}} 2020.
\newblock Language models are few-shot learners.
\newblock {\em arXiv preprint arXiv:2005.14165\/}.

\bibitem[\protect\citeauthoryear{Caruana and Niculescu-Mizil}{Caruana and
  Niculescu-Mizil}{2004}]{caruana2004data}
{\sc Caruana, R.} {\sc and} {\sc Niculescu-Mizil, A.} 2004.
\newblock Data mining in metric space: an empirical analysis of supervised
  learning performance criteria.
\newblock In {\em Proceedings of the tenth ACM SIGKDD international conference
  on Knowledge discovery and data mining}. 69--78.

\bibitem[\protect\citeauthoryear{Cen, Koedinger, and Junker}{Cen
  et~al\mbox{.}}{2006}]{Cen2006}
{\sc Cen, H.}, {\sc Koedinger, K.}, {\sc and} {\sc Junker, B.} 2006.
\newblock {Learning factors analysis–a general method for cognitive model
  evaluation and improvement}.
\newblock In {\em International Conference on Intelligent Tutoring Systems}.
  164--175.

\bibitem[\protect\citeauthoryear{Chang, Beck, Mostow, and Corbett}{Chang
  et~al\mbox{.}}{2006}]{chang2006does}
{\sc Chang, K.-m.}, {\sc Beck, J.~E.}, {\sc Mostow, J.}, {\sc and} {\sc
  Corbett, A.} 2006.
\newblock Does help help? a {Bayes} net approach to modeling tutor
  interventions.
\newblock In {\em AAAI2006 Workshop on Educational Data Mining}.

\bibitem[\protect\citeauthoryear{Cheng, Liu, and Chen}{Cheng
  et~al\mbox{.}}{2020}]{cheng2020domain}
{\sc Cheng, S.}, {\sc Liu, Q.}, {\sc and} {\sc Chen, E.} 2020.
\newblock Domain adaption for knowledge tracing.
\newblock {\em arXiv preprint arXiv:2001.04841\/}.

\bibitem[\protect\citeauthoryear{Chicco and Jurman}{Chicco and
  Jurman}{2020}]{chicco2020advantages}
{\sc Chicco, D.} {\sc and} {\sc Jurman, G.} 2020.
\newblock The advantages of the {Matthews} correlation coefficient ({MCC}) over
  {F1} score and accuracy in binary classification evaluation.
\newblock {\em BMC genomics\/}~{\em 21,\/}~1, 1--13.

\bibitem[\protect\citeauthoryear{Choffin, Popineau, Bourda, and Vie}{Choffin
  et~al\mbox{.}}{2019}]{choffin2019das3h}
{\sc Choffin, B.}, {\sc Popineau, F.}, {\sc Bourda, Y.}, {\sc and} {\sc Vie,
  J.-J.} 2019.
\newblock Das3h: modeling student learning and forgetting for optimally
  scheduling distributed practice of skills.
\newblock {\em arXiv preprint arXiv:1905.06873\/}.

\bibitem[\protect\citeauthoryear{Choi, Lee, Cho, Baek, Kim, Cha, Shin, Bae, and
  Heo}{Choi et~al\mbox{.}}{2020}]{choi2020towards}
{\sc Choi, Y.}, {\sc Lee, Y.}, {\sc Cho, J.}, {\sc Baek, J.}, {\sc Kim, B.},
  {\sc Cha, Y.}, {\sc Shin, D.}, {\sc Bae, C.}, {\sc and} {\sc Heo, J.} 2020.
\newblock Towards an appropriate query, key, and value computation for
  knowledge tracing.
\newblock In {\em Proceedings of the Seventh ACM Conference on Learning@
  Scale}. 341--344.

\bibitem[\protect\citeauthoryear{Choi, Lee, Shin, Cho, Park, Lee, Baek, Bae,
  Kim, and Heo}{Choi et~al\mbox{.}}{2020}]{choi2020ednet}
{\sc Choi, Y.}, {\sc Lee, Y.}, {\sc Shin, D.}, {\sc Cho, J.}, {\sc Park, S.},
  {\sc Lee, S.}, {\sc Baek, J.}, {\sc Bae, C.}, {\sc Kim, B.}, {\sc and} {\sc
  Heo, J.} 2020.
\newblock {EdNet}: A large-scale hierarchical dataset in education.
\newblock In {\em International Conference on Artificial Intelligence in
  Education}. Springer, 69--73.

\bibitem[\protect\citeauthoryear{Chrysafiadi and Virvou}{Chrysafiadi and
  Virvou}{2013}]{chrysafiadi2013student}
{\sc Chrysafiadi, K.} {\sc and} {\sc Virvou, M.} 2013.
\newblock Student modeling approaches: A literature review for the last decade.
\newblock {\em Expert Systems with Applications\/}~{\em 40,\/}~11, 4715--4729.

\bibitem[\protect\citeauthoryear{Collaboration et~al\mbox{.}}{Collaboration
  et~al\mbox{.}}{2015}]{open2015}
{\sc Collaboration, O.~S.} {\sc et~al\mbox{.}} 2015.
\newblock Estimating the reproducibility of psychological science.
\newblock {\em Science\/}~{\em 349,\/}~6251, aac4716.

\bibitem[\protect\citeauthoryear{Corbett and Anderson}{Corbett and
  Anderson}{1994}]{corbett1994knowledge}
{\sc Corbett, A.~T.} {\sc and} {\sc Anderson, J.~R.} 1994.
\newblock Knowledge tracing: Modeling the acquisition of procedural knowledge.
\newblock {\em User modeling and user-adapted interaction\/}~{\em 4,\/}~4,
  253--278.

\bibitem[\protect\citeauthoryear{Corbett, Koedinger, and Anderson}{Corbett
  et~al\mbox{.}}{1997}]{corbett1997intelligent}
{\sc Corbett, A.~T.}, {\sc Koedinger, K.~R.}, {\sc and} {\sc Anderson, J.~R.}
  1997.
\newblock Intelligent tutoring systems.
\newblock In {\em Handbook of human-computer interaction}. Elsevier, 849--874.

\bibitem[\protect\citeauthoryear{Council, Committee, et~al\mbox{.}}{Council
  et~al\mbox{.}}{2005}]{national2005thinking}
{\sc Council, N.~R.}, {\sc Committee, C.~R.}, {\sc et~al\mbox{.}} 2005.
\newblock Chapter 3: Principles for developing metrics.
\newblock In {\em Thinking strategically: the appropriate use of metrics for
  the climate change science program}. National Academies Press.

\bibitem[\protect\citeauthoryear{Devasena, Sumathi, Gomathi, and
  Hemalatha}{Devasena et~al\mbox{.}}{2011}]{devasena2011effectiveness}
{\sc Devasena, C.~L.}, {\sc Sumathi, T.}, {\sc Gomathi, V.}, {\sc and} {\sc
  Hemalatha, M.} 2011.
\newblock Effectiveness evaluation of rule based classifiers for the
  classification of iris data set.
\newblock {\em Bonfring International Journal of Man Machine Interface\/}~{\em
  1,\/}~Special Issue Inaugural Special Issue, 05--09.

\bibitem[\protect\citeauthoryear{Dhanani, Lee, Phothilimthana, and
  Pardos}{Dhanani et~al\mbox{.}}{2014}]{dhanani2014comparison}
{\sc Dhanani, A.}, {\sc Lee, S.~Y.}, {\sc Phothilimthana, P.~M.}, {\sc and}
  {\sc Pardos, Z.} 2014.
\newblock A comparison of error metrics for learning model parameters in
  bayesian knowledge tracing.
\newblock In {\em Workshop Approaching Twenty Years of Knowledge Tracing
  (BKT20y). Citeseer}. 8--9.

\bibitem[\protect\citeauthoryear{Ding and Larson}{Ding and
  Larson}{2019}]{ding2019deep}
{\sc Ding, X.} {\sc and} {\sc Larson, E.~C.} 2019.
\newblock Why deep knowledge tracing has less depth than anticipated.

\bibitem[\protect\citeauthoryear{Dozat}{Dozat}{2016}]{dozat2016incorporating}
{\sc Dozat, T.} 2016.
\newblock Incorporating nesterov momentum into adam.

\bibitem[\protect\citeauthoryear{Effenberger and Pel{\'a}nek}{Effenberger and
  Pel{\'a}nek}{2020}]{effenberger2020impact}
{\sc Effenberger, T.} {\sc and} {\sc Pel{\'a}nek, R.} 2020.
\newblock Impact of methodological choices on the evaluation of student models.
\newblock In {\em International Conference on Artificial Intelligence in
  Education}. Springer, 153--164.

\bibitem[\protect\citeauthoryear{Fanelli}{Fanelli}{2011}]{fanelli2011}
{\sc Fanelli, D.} 2011.
\newblock Negative results are disappearing from most disciplines and
  countries.
\newblock {\em Scientometrics\/}~{\em 90,\/}~3, 891--904.

\bibitem[\protect\citeauthoryear{Faraway and Augustin}{Faraway and
  Augustin}{2018}]{faraway2018small}
{\sc Faraway, J.~J.} {\sc and} {\sc Augustin, N.~H.} 2018.
\newblock When small data beats big data.
\newblock {\em Statistics \& Probability Letters\/}~{\em 136}, 142--145.

\bibitem[\protect\citeauthoryear{Fawcett}{Fawcett}{2006}]{fawcett2006introduction}
{\sc Fawcett, T.} 2006.
\newblock An introduction to {ROC} analysis.
\newblock {\em Pattern recognition letters\/}~{\em 27,\/}~8, 861--874.

\bibitem[\protect\citeauthoryear{Feng, Heffernan, and Koedinger}{Feng
  et~al\mbox{.}}{2009}]{feng2009addressing}
{\sc Feng, M.}, {\sc Heffernan, N.}, {\sc and} {\sc Koedinger, K.} 2009.
\newblock Addressing the assessment challenge with an online system that tutors
  as it assesses.
\newblock {\em User Modeling and User-Adapted Interaction\/}~{\em 19,\/}~3,
  243--266.

\bibitem[\protect\citeauthoryear{Ferri, Hern{\'a}ndez-Orallo, and
  Modroiu}{Ferri et~al\mbox{.}}{2009}]{ferri2009experimental}
{\sc Ferri, C.}, {\sc Hern{\'a}ndez-Orallo, J.}, {\sc and} {\sc Modroiu, R.}
  2009.
\newblock An experimental comparison of performance measures for
  classification.
\newblock {\em Pattern Recognition Letters\/}~{\em 30,\/}~1, 27--38.

\bibitem[\protect\citeauthoryear{Gal and Ghahramani}{Gal and
  Ghahramani}{2016}]{gal2016theoretically}
{\sc Gal, Y.} {\sc and} {\sc Ghahramani, Z.} 2016.
\newblock A theoretically grounded application of dropout in recurrent neural
  networks.
\newblock In {\em Advances in neural information processing systems}.
  1019--1027.

\bibitem[\protect\citeauthoryear{Gardner, Yang, Baker, and Brooks}{Gardner
  et~al\mbox{.}}{2019}]{gardner2019modeling}
{\sc Gardner, J.}, {\sc Yang, Y.}, {\sc Baker, R.~S.}, {\sc and} {\sc Brooks,
  C.} 2019.
\newblock Modeling and experimental design for mooc dropout prediction: A
  replication perspective.
\newblock In {\em Proceedings of The 12th International Conference on
  Educational Data Mining (EDM 2019)}. ERIC.

\bibitem[\protect\citeauthoryear{Gers and Schmidhuber}{Gers and
  Schmidhuber}{2001}]{gers2001lstm}
{\sc Gers, F.~A.} {\sc and} {\sc Schmidhuber, E.} 2001.
\newblock Lstm recurrent networks learn simple context-free and
  context-sensitive languages.
\newblock {\em IEEE Transactions on Neural Networks\/}~{\em 12,\/}~6,
  1333--1340.

\bibitem[\protect\citeauthoryear{Gervet, Koedinger, Schneider, Mitchell,
  et~al\mbox{.}}{Gervet et~al\mbox{.}}{2020}]{gervet2020deep}
{\sc Gervet, T.}, {\sc Koedinger, K.}, {\sc Schneider, J.}, {\sc Mitchell, T.},
  {\sc et~al\mbox{.}} 2020.
\newblock When is deep learning the best approach to knowledge tracing?
\newblock {\em JEDM| Journal of Educational Data Mining\/}~{\em 12,\/}~3,
  31--54.

\bibitem[\protect\citeauthoryear{Ghahramani}{Ghahramani}{1997}]{ghahramani1997learning}
{\sc Ghahramani, Z.} 1997.
\newblock Learning dynamic bayesian networks.
\newblock In {\em International School on Neural Networks, Initiated by IIASS
  and EMFCSC}. Springer, 168--197.

\bibitem[\protect\citeauthoryear{Ghosh, Heffernan, and Lan}{Ghosh
  et~al\mbox{.}}{2020}]{ghosh2020context}
{\sc Ghosh, A.}, {\sc Heffernan, N.}, {\sc and} {\sc Lan, A.~S.} 2020.
\newblock Context-aware attentive knowledge tracing.
\newblock In {\em Proceedings of the 26th ACM SIGKDD International Conference
  on Knowledge Discovery \& Data Mining}. 2330--2339.

\bibitem[\protect\citeauthoryear{Gilbert, King, Pettigrew, and Wilson}{Gilbert
  et~al\mbox{.}}{2016}]{gilbert2016}
{\sc Gilbert, D.~T.}, {\sc King, G.}, {\sc Pettigrew, S.}, {\sc and} {\sc
  Wilson, T.} 2016.
\newblock Comment on ``estimating the reproducibility of psychological
  science''.
\newblock {\em Science\/}~{\em 351,\/}~6277, 1037.

\bibitem[\protect\citeauthoryear{Golden}{Golden}{1995}]{golden1995}
{\sc Golden, M.~A.} 1995.
\newblock Replication and non-quantitative research.
\newblock {\em PS: Political Science \& Politics\/}~{\em 28,\/}~03, 481--483.

\bibitem[\protect\citeauthoryear{Gong, Beck, and Heffernan}{Gong
  et~al\mbox{.}}{2010}]{gong2010comparing}
{\sc Gong, Y.}, {\sc Beck, J.~E.}, {\sc and} {\sc Heffernan, N.~T.} 2010.
\newblock Comparing knowledge tracing and performance factor analysis by using
  multiple model fitting procedures.
\newblock In {\em International conference on intelligent tutoring systems}.
  Springer, 35--44.

\bibitem[\protect\citeauthoryear{Gonz{\'a}lez-Brenes, Huang, and
  Brusilovsky}{Gonz{\'a}lez-Brenes et~al\mbox{.}}{2014}]{gonzalez2014general}
{\sc Gonz{\'a}lez-Brenes, J.}, {\sc Huang, Y.}, {\sc and} {\sc Brusilovsky, P.}
  2014.
\newblock General features in knowledge tracing to model multiple subskills,
  temporal item response theory, and expert knowledge.
\newblock In {\em The 7th International Conference on Educational Data Mining}.
  University of Pittsburgh, 84--91.

\bibitem[\protect\citeauthoryear{Graves}{Graves}{2013}]{graves2013generating}
{\sc Graves, A.} 2013.
\newblock Generating sequences with recurrent neural networks.
\newblock {\em arXiv preprint arXiv:1308.0850\/}.

\bibitem[\protect\citeauthoryear{Graves, Wayne, and Danihelka}{Graves
  et~al\mbox{.}}{2014}]{graves2014neural}
{\sc Graves, A.}, {\sc Wayne, G.}, {\sc and} {\sc Danihelka, I.} 2014.
\newblock Neural turing machines.
\newblock {\em arXiv preprint arXiv:1410.5401\/}.

\bibitem[\protect\citeauthoryear{Gunawardana and Shani}{Gunawardana and
  Shani}{2009}]{gunawardana2009survey}
{\sc Gunawardana, A.} {\sc and} {\sc Shani, G.} 2009.
\newblock A survey of accuracy evaluation metrics of recommendation tasks.
\newblock {\em Journal of Machine Learning Research\/}~{\em 10,\/}~12.

\bibitem[\protect\citeauthoryear{Halimu, Kasem, and Newaz}{Halimu
  et~al\mbox{.}}{2019}]{halimu2019empirical}
{\sc Halimu, C.}, {\sc Kasem, A.}, {\sc and} {\sc Newaz, S.~S.} 2019.
\newblock Empirical comparison of area under {ROC} curve ({AUC}) and mathew
  correlation coefficient (mcc) for evaluating machine learning algorithms on
  imbalanced datasets for binary classification.
\newblock In {\em Proceedings of the 3rd international conference on machine
  learning and soft computing}. 1--6.

\bibitem[\protect\citeauthoryear{Hambleton and Swaminathan}{Hambleton and
  Swaminathan}{1985}]{hambleton1985item}
{\sc Hambleton, R.~K.} {\sc and} {\sc Swaminathan, H.} 1985.
\newblock {\em Item response theory: Principles and applications}.
\newblock Springer.

\bibitem[\protect\citeauthoryear{Heffernan and Heffernan}{Heffernan and
  Heffernan}{2014}]{heffernan2014assistments}
{\sc Heffernan, N.~T.} {\sc and} {\sc Heffernan, C.~L.} 2014.
\newblock The assistments ecosystem: Building a platform that brings scientists
  and teachers together for minimally invasive research on human learning and
  teaching.
\newblock {\em International Journal of Artificial Intelligence in
  Education\/}~{\em 24,\/}~4, 470--497.

\bibitem[\protect\citeauthoryear{Heffernan, Turner, Lourenco, Macasek,
  Nuzzo-Jones, and Koedinger}{Heffernan
  et~al\mbox{.}}{2006}]{heffernan2006assistment}
{\sc Heffernan, N.~T.}, {\sc Turner, T.~E.}, {\sc Lourenco, A.~L.}, {\sc
  Macasek, M.~A.}, {\sc Nuzzo-Jones, G.}, {\sc and} {\sc Koedinger, K.~R.}
  2006.
\newblock The assistment builder: Towards an analysis of cost effectiveness of
  its creation.
\newblock In {\em Flairs Conference}. 515--520.

\bibitem[\protect\citeauthoryear{Hochreiter and Schmidhuber}{Hochreiter and
  Schmidhuber}{1997a}]{hochreiter1997long}
{\sc Hochreiter, S.} {\sc and} {\sc Schmidhuber, J.} 1997a.
\newblock Long short-term memory.
\newblock {\em Neural computation\/}~{\em 9,\/}~8, 1735--1780.

\bibitem[\protect\citeauthoryear{Hochreiter and Schmidhuber}{Hochreiter and
  Schmidhuber}{1997b}]{hochreiter1997lstm}
{\sc Hochreiter, S.} {\sc and} {\sc Schmidhuber, J.} 1997b.
\newblock {LSTM} can solve hard long time lag problems.
\newblock In {\em Advances in neural information processing systems}. 473--479.

\bibitem[\protect\citeauthoryear{Huang and Ling}{Huang and
  Ling}{2005}]{huang2005using}
{\sc Huang, J.} {\sc and} {\sc Ling, C.~X.} 2005.
\newblock Using {AUC} and accuracy in evaluating learning algorithms.
\newblock {\em IEEE Transactions on knowledge and Data Engineering\/}~{\em
  17,\/}~3, 299--310.

\bibitem[\protect\citeauthoryear{Huang, Shea, Qian, Masurkar, Deng, and
  Liu}{Huang et~al\mbox{.}}{2019}]{huang2019patient}
{\sc Huang, L.}, {\sc Shea, A.~L.}, {\sc Qian, H.}, {\sc Masurkar, A.}, {\sc
  Deng, H.}, {\sc and} {\sc Liu, D.} 2019.
\newblock Patient clustering improves efficiency of federated machine learning
  to predict mortality and hospital stay time using distributed electronic
  medical records.
\newblock {\em Journal of biomedical informatics\/}~{\em 99}, 103291.

\bibitem[\protect\citeauthoryear{Ihantola, Vihavainen, Ahadi, Butler,
  B\"{o}rstler, Edwards, Isohanni, Korhonen, Petersen, Rivers, Rubio, Sheard,
  Skupas, Spacco, Szabo, and Toll}{Ihantola et~al\mbox{.}}{2015}]{Ihantola2015}
{\sc Ihantola, P.}, {\sc Vihavainen, A.}, {\sc Ahadi, A.}, {\sc Butler, M.},
  {\sc B\"{o}rstler, J.}, {\sc Edwards, S.~H.}, {\sc Isohanni, E.}, {\sc
  Korhonen, A.}, {\sc Petersen, A.}, {\sc Rivers, K.}, {\sc Rubio, M.~A.}, {\sc
  Sheard, J.}, {\sc Skupas, B.}, {\sc Spacco, J.}, {\sc Szabo, C.}, {\sc and}
  {\sc Toll, D.} 2015.
\newblock Educational data mining and learning analytics in programming:
  Literature review and case studies.
\newblock In {\em Proceedings of the 2015 ITiCSE on Working Group Reports}.
  ITICSE-WGR '15. ACM, New York, NY, USA, 41--63.

\bibitem[\protect\citeauthoryear{Ioannidis}{Ioannidis}{2005a}]{Ioannidis2005-2}
{\sc Ioannidis, J.~P.} 2005a.
\newblock Contradicted and initially stronger effects in highly cited clinical
  research.
\newblock {\em Jama\/}~{\em 294,\/}~2, 218--228.

\bibitem[\protect\citeauthoryear{Ioannidis}{Ioannidis}{2005b}]{Ioannidis2005}
{\sc Ioannidis, J.~P.} 2005b.
\newblock Why most published research findings are false.
\newblock {\em PLoS Med\/}~{\em 2,\/}~8, e124.

\bibitem[\protect\citeauthoryear{Ioannidis, Munafo, Fusar-Poli, Nosek, and
  David}{Ioannidis et~al\mbox{.}}{2014}]{ioannidis2014}
{\sc Ioannidis, J.~P.}, {\sc Munafo, M.~R.}, {\sc Fusar-Poli, P.}, {\sc Nosek,
  B.~A.}, {\sc and} {\sc David, S.~P.} 2014.
\newblock Publication and other reporting biases in cognitive sciences:
  detection, prevalence, and prevention.
\newblock {\em Trends in Cognitive Sciences\/}~{\em 18,\/}~5, 235--241.

\bibitem[\protect\citeauthoryear{Jeni, Cohn, and De~La~Torre}{Jeni
  et~al\mbox{.}}{2013}]{jeni2013facing}
{\sc Jeni, L.~A.}, {\sc Cohn, J.~F.}, {\sc and} {\sc De~La~Torre, F.} 2013.
\newblock Facing imbalanced data--recommendations for the use of performance
  metrics.
\newblock In {\em 2013 Humaine association conference on affective computing
  and intelligent interaction}. IEEE, 245--251.

\bibitem[\protect\citeauthoryear{Johns, Mahadevan, and Woolf}{Johns
  et~al\mbox{.}}{2006}]{johns2006estimating}
{\sc Johns, J.}, {\sc Mahadevan, S.}, {\sc and} {\sc Woolf, B.} 2006.
\newblock Estimating student proficiency using an item response theory model.
\newblock In {\em International conference on intelligent tutoring systems}.
  Springer, 473--480.

\bibitem[\protect\citeauthoryear{{Kamijo} and {Tanigawa}}{{Kamijo} and
  {Tanigawa}}{1990}]{kamijo1990stockrnn}
{\sc {Kamijo}, K.} {\sc and} {\sc {Tanigawa}, T.} 1990.
\newblock Stock price pattern recognition-a recurrent neural network approach.
\newblock In {\em 1990 IJCNN International Joint Conference on Neural
  Networks}. 215--221 vol.1.

\bibitem[\protect\citeauthoryear{K{\"a}ser, Klingler, Schwing, and
  Gross}{K{\"a}ser et~al\mbox{.}}{2014}]{kaser2014beyond}
{\sc K{\"a}ser, T.}, {\sc Klingler, S.}, {\sc Schwing, A.~G.}, {\sc and} {\sc
  Gross, M.} 2014.
\newblock Beyond knowledge tracing: Modeling skill topologies with bayesian
  networks.
\newblock In {\em International conference on intelligent tutoring systems}.
  Springer, 188--198.

\bibitem[\protect\citeauthoryear{Khajah, Lindsey, and Mozer}{Khajah
  et~al\mbox{.}}{2016}]{khajah2016deep}
{\sc Khajah, M.}, {\sc Lindsey, R.~V.}, {\sc and} {\sc Mozer, M.~C.} 2016.
\newblock How deep is knowledge tracing?
\newblock In {\em Proceedings of the 9th International Conference on
  Educational Data Mining (EDM 2016)}. ERIC.

\bibitem[\protect\citeauthoryear{Khajah, Wing, Lindsey, and Mozer}{Khajah
  et~al\mbox{.}}{2014}]{khajah2014integratingA}
{\sc Khajah, M.}, {\sc Wing, R.}, {\sc Lindsey, R.}, {\sc and} {\sc Mozer, M.}
  2014.
\newblock Integrating latent-factor and knowledge-tracing models to predict
  individual differences in learning.
\newblock In {\em Educational Data Mining 2014}. Citeseer.

\bibitem[\protect\citeauthoryear{Khajah, Huang, Gonz{\'a}lez-Brenes, Mozer, and
  Brusilovsky}{Khajah et~al\mbox{.}}{2014}]{khajah2014integrating}
{\sc Khajah, M.~M.}, {\sc Huang, Y.}, {\sc Gonz{\'a}lez-Brenes, J.~P.}, {\sc
  Mozer, M.~C.}, {\sc and} {\sc Brusilovsky, P.} 2014.
\newblock Integrating knowledge tracing and item response theory: A tale of two
  frameworks.
\newblock In {\em CEUR Workshop Proceedings}. Vol. 1181. University of
  Pittsburgh, 7--15.

\bibitem[\protect\citeauthoryear{Kim, Cho, and Oh}{Kim
  et~al\mbox{.}}{2017}]{kim2017development}
{\sc Kim, S.~J.}, {\sc Cho, K.~J.}, {\sc and} {\sc Oh, S.} 2017.
\newblock Development of machine learning models for diagnosis of glaucoma.
\newblock {\em PloS one\/}~{\em 12,\/}~5, e0177726.

\bibitem[\protect\citeauthoryear{Koedinger, Baker, Cunningham, Skogsholm,
  Leber, and Stamper}{Koedinger et~al\mbox{.}}{2010}]{koedinger2010data}
{\sc Koedinger, K.~R.}, {\sc Baker, R.~S.}, {\sc Cunningham, K.}, {\sc
  Skogsholm, A.}, {\sc Leber, B.}, {\sc and} {\sc Stamper, J.} 2010.
\newblock A data repository for the {EDM} community: {The PSLC DataShop}.
\newblock {\em Handbook of educational data mining\/}~{\em 43}, 43--56.

\bibitem[\protect\citeauthoryear{Lalwani and Agrawal}{Lalwani and
  Agrawal}{2017}]{lalwani2017few}
{\sc Lalwani, A.} {\sc and} {\sc Agrawal, S.} 2017.
\newblock Few hundred parameters outperform few hundred thousand.
\newblock In {\em Proceedings of the 10th International Conference on
  Educational Data Mining, EDM}. Vol.~17. ERIC, 448--453.

\bibitem[\protect\citeauthoryear{LeCun, Bengio, and Hinton}{LeCun
  et~al\mbox{.}}{2015}]{lecun2015deep}
{\sc LeCun, Y.}, {\sc Bengio, Y.}, {\sc and} {\sc Hinton, G.} 2015.
\newblock Deep learning.
\newblock {\em nature\/}~{\em 521,\/}~7553, 436--444.

\bibitem[\protect\citeauthoryear{Lin and Chi}{Lin and
  Chi}{2016}]{lin2016intervention}
{\sc Lin, C.} {\sc and} {\sc Chi, M.} 2016.
\newblock Intervention-{BKT}: incorporating instructional interventions into
  bayesian knowledge tracing.
\newblock In {\em International conference on intelligent tutoring systems}.
  Springer, 208--218.

\bibitem[\protect\citeauthoryear{Lin and Chi}{Lin and
  Chi}{2017}]{lin2017comparisons}
{\sc Lin, C.} {\sc and} {\sc Chi, M.} 2017.
\newblock A comparisons of {BKT}, {RNN} and {LSTM} for learning gain
  prediction.
\newblock In {\em International Conference on Artificial Intelligence in
  Education}. Springer, 536--539.

\bibitem[\protect\citeauthoryear{Ling, Huang, Zhang, et~al\mbox{.}}{Ling
  et~al\mbox{.}}{2003}]{ling2003auc}
{\sc Ling, C.~X.}, {\sc Huang, J.}, {\sc Zhang, H.}, {\sc et~al\mbox{.}} 2003.
\newblock {AUC}: a statistically consistent and more discriminating measure
  than accuracy.
\newblock In {\em Ijcai}. Vol.~3. 519--524.

\bibitem[\protect\citeauthoryear{Lipton and Steinhardt}{Lipton and
  Steinhardt}{2018}]{lipton2018troubling}
{\sc Lipton, Z.~C.} {\sc and} {\sc Steinhardt, J.} 2018.
\newblock Troubling trends in machine learning scholarship.
\newblock {\em arXiv preprint arXiv:1807.03341\/}.

\bibitem[\protect\citeauthoryear{Liu, White, and Newell}{Liu
  et~al\mbox{.}}{2011}]{liu2011measuring}
{\sc Liu, C.}, {\sc White, M.}, {\sc and} {\sc Newell, G.} 2011.
\newblock Measuring and comparing the accuracy of species distribution models
  with presence--absence data.
\newblock {\em Ecography\/}~{\em 34,\/}~2, 232--243.

\bibitem[\protect\citeauthoryear{Liu, Huang, Yin, Chen, Xiong, Su, and Hu}{Liu
  et~al\mbox{.}}{2019}]{liu2019ekt}
{\sc Liu, Q.}, {\sc Huang, Z.}, {\sc Yin, Y.}, {\sc Chen, E.}, {\sc Xiong, H.},
  {\sc Su, Y.}, {\sc and} {\sc Hu, G.} 2019.
\newblock {EKT}: Exercise-aware knowledge tracing for student performance
  prediction.
\newblock {\em IEEE Transactions on Knowledge and Data Engineering\/}~{\em
  33,\/}~1, 100--115.

\bibitem[\protect\citeauthoryear{Liu, Tong, Liu, Zhao, Chen, Ma, and Wang}{Liu
  et~al\mbox{.}}{2019}]{liu2019exploiting}
{\sc Liu, Q.}, {\sc Tong, S.}, {\sc Liu, C.}, {\sc Zhao, H.}, {\sc Chen, E.},
  {\sc Ma, H.}, {\sc and} {\sc Wang, S.} 2019.
\newblock Exploiting cognitive structure for adaptive learning.
\newblock In {\em Proceedings of the 25th ACM SIGKDD International Conference
  on Knowledge Discovery \& Data Mining}. 627--635.

\bibitem[\protect\citeauthoryear{Lobo, Jim{\'e}nez-Valverde, and Real}{Lobo
  et~al\mbox{.}}{2008}]{lobo2008auc}
{\sc Lobo, J.~M.}, {\sc Jim{\'e}nez-Valverde, A.}, {\sc and} {\sc Real, R.}
  2008.
\newblock {AUC}: a misleading measure of the performance of predictive
  distribution models.
\newblock {\em Global ecology and Biogeography\/}~{\em 17,\/}~2, 145--151.

\bibitem[\protect\citeauthoryear{Luong, Pham, and Manning}{Luong
  et~al\mbox{.}}{2015}]{luong2015effective}
{\sc Luong, M.-T.}, {\sc Pham, H.}, {\sc and} {\sc Manning, C.~D.} 2015.
\newblock Effective approaches to attention-based neural machine translation.
\newblock {\em arXiv preprint arXiv:1508.04025\/}.

\bibitem[\protect\citeauthoryear{Ma, Adesope, Nesbit, and Liu}{Ma
  et~al\mbox{.}}{2014}]{ma2014intelligent}
{\sc Ma, W.}, {\sc Adesope, O.~O.}, {\sc Nesbit, J.~C.}, {\sc and} {\sc Liu,
  Q.} 2014.
\newblock Intelligent tutoring systems and learning outcomes: A meta-analysis.
\newblock {\em Journal of educational psychology\/}~{\em 106,\/}~4, 901.

\bibitem[\protect\citeauthoryear{Mackey}{Mackey}{2012}]{Mackey2012}
{\sc Mackey, A.} 2012.
\newblock Why (or why not), when and how to replicate research.
\newblock {\em Replication research in applied linguistics\/}~{\em 2146}.

\bibitem[\protect\citeauthoryear{Makel, Plucker, and Hegarty}{Makel
  et~al\mbox{.}}{2012}]{makel2012}
{\sc Makel, M.~C.}, {\sc Plucker, J.~A.}, {\sc and} {\sc Hegarty, B.} 2012.
\newblock Replications in psychology research how often do they really occur?
\newblock {\em Perspectives on Psychological Science\/}~{\em 7,\/}~6, 537--542.

\bibitem[\protect\citeauthoryear{Mandalapu, Gong, and Chen}{Mandalapu
  et~al\mbox{.}}{2021}]{mandalapu2021we}
{\sc Mandalapu, V.}, {\sc Gong, J.}, {\sc and} {\sc Chen, L.} 2021.
\newblock Do we need to go deep? knowledge tracing with big data.
\newblock {\em arXiv preprint arXiv:2101.08349\/}.

\bibitem[\protect\citeauthoryear{Mao, Lin, and Chi}{Mao
  et~al\mbox{.}}{2018}]{mao2018deep}
{\sc Mao, Y.}, {\sc Lin, C.}, {\sc and} {\sc Chi, M.} 2018.
\newblock Deep learning vs. bayesian knowledge tracing: Student models for
  interventions.
\newblock {\em JEDM| Journal of Educational Data Mining\/}~{\em 10,\/}~2,
  28--54.

\bibitem[\protect\citeauthoryear{{Mikolov}, {Kombrink}, {Burget}, {Černocký},
  and {Khudanpur}}{{Mikolov} et~al\mbox{.}}{2011}]{mikolov2011ext_rnn}
{\sc {Mikolov}, T.}, {\sc {Kombrink}, S.}, {\sc {Burget}, L.}, {\sc
  {Černocký}, J.}, {\sc and} {\sc {Khudanpur}, S.} 2011.
\newblock Extensions of recurrent neural network language model.
\newblock In {\em 2011 IEEE International Conference on Acoustics, Speech and
  Signal Processing (ICASSP)}. 5528--5531.

\bibitem[\protect\citeauthoryear{Miller, Fisch, Dodge, Karimi, Bordes, and
  Weston}{Miller et~al\mbox{.}}{2016}]{miller2016key}
{\sc Miller, A.}, {\sc Fisch, A.}, {\sc Dodge, J.}, {\sc Karimi, A.-H.}, {\sc
  Bordes, A.}, {\sc and} {\sc Weston, J.} 2016.
\newblock Key-value memory networks for directly reading documents.
\newblock {\em arXiv preprint arXiv:1606.03126\/}.

\bibitem[\protect\citeauthoryear{Molnar}{Molnar}{2020}]{molnar2020interpretable}
{\sc Molnar, C.} 2020.
\newblock {\em Interpretable machine learning}.
\newblock Lulu. com.

\bibitem[\protect\citeauthoryear{Montero, Arora, Kelly, Milne, and
  Mozer}{Montero et~al\mbox{.}}{2018}]{montero2018does}
{\sc Montero, S.}, {\sc Arora, A.}, {\sc Kelly, S.}, {\sc Milne, B.}, {\sc and}
  {\sc Mozer, M.} 2018.
\newblock Does deep knowledge tracing model interactions among skills?.
\newblock {\em International Educational Data Mining Society\/}.

\bibitem[\protect\citeauthoryear{Moonesinghe, Khoury, and Janssens}{Moonesinghe
  et~al\mbox{.}}{2007}]{Moonesinghe2007}
{\sc Moonesinghe, R.}, {\sc Khoury, M.~J.}, {\sc and} {\sc Janssens, A. C.~J.}
  2007.
\newblock Most published research findings are false -- but a little
  replication goes a long way.
\newblock {\em PLoS Med\/}~{\em 4,\/}~2, e28.

\bibitem[\protect\citeauthoryear{Muma}{Muma}{1993}]{muma1993}
{\sc Muma, J.~R.} 1993.
\newblock The need for replication.
\newblock {\em Journal of Speech, Language, and Hearing Research\/}~{\em
  36,\/}~5, 927--930.

\bibitem[\protect\citeauthoryear{Muschelli}{Muschelli}{2020}]{muschelli2020roc}
{\sc Muschelli, J.} 2020.
\newblock Roc and auc with a binary predictor: a potentially misleading metric.
\newblock {\em Journal of classification\/}~{\em 37,\/}~3, 696--708.

\bibitem[\protect\citeauthoryear{Nakagawa, Iwasawa, and Matsuo}{Nakagawa
  et~al\mbox{.}}{2019}]{Nakagawa:2019:GKT:3350546.3352513}
{\sc Nakagawa, H.}, {\sc Iwasawa, Y.}, {\sc and} {\sc Matsuo, Y.} 2019.
\newblock Graph-based knowledge tracing: Modeling student proficiency using
  graph neural network.
\newblock In {\em IEEE/WIC/ACM International Conference on Web Intelligence}.
  WI '19. ACM, New York, NY, USA, 156--163.

\bibitem[\protect\citeauthoryear{Nwana}{Nwana}{1990}]{nwana1990intelligent}
{\sc Nwana, H.~S.} 1990.
\newblock Intelligent tutoring systems: an overview.
\newblock {\em Artificial Intelligence Review\/}~{\em 4,\/}~4, 251--277.

\bibitem[\protect\citeauthoryear{Oya and Morishima}{Oya and
  Morishima}{2021}]{oya2021lstm}
{\sc Oya, T.} {\sc and} {\sc Morishima, S.} 2021.
\newblock {LSTM-SAKT}: {LSTM}-encoded {SAKT}-like transformer for knowledge
  tracing, 2nd place solution for riiid! answer correctness prediction.
\newblock {\em arXiv preprint arXiv:2102.00845\/}.

\bibitem[\protect\citeauthoryear{Ozenne, Subtil, and Maucort-Boulch}{Ozenne
  et~al\mbox{.}}{2015}]{ozenne2015precision}
{\sc Ozenne, B.}, {\sc Subtil, F.}, {\sc and} {\sc Maucort-Boulch, D.} 2015.
\newblock The precision--recall curve overcame the optimism of the receiver
  operating characteristic curve in rare diseases.
\newblock {\em Journal of clinical epidemiology\/}~{\em 68,\/}~8, 855--859.

\bibitem[\protect\citeauthoryear{Pahikkala, Pyysalo, Boberg, J{\"a}rvinen, and
  Salakoski}{Pahikkala et~al\mbox{.}}{2009}]{pahikkala2009matrix}
{\sc Pahikkala, T.}, {\sc Pyysalo, S.}, {\sc Boberg, J.}, {\sc J{\"a}rvinen,
  J.}, {\sc and} {\sc Salakoski, T.} 2009.
\newblock Matrix representations, linear transformations, and kernels for
  disambiguation in natural language.
\newblock {\em Machine Learning\/}~{\em 74,\/}~2, 133--158.

\bibitem[\protect\citeauthoryear{Pandey and Karypis}{Pandey and
  Karypis}{2019}]{p2019selfattentive}
{\sc Pandey, S.} {\sc and} {\sc Karypis, G.} 2019.
\newblock A self-attentive model for knowledge tracing.

\bibitem[\protect\citeauthoryear{Pandey, Karypis, and Srivastava}{Pandey
  et~al\mbox{.}}{2021}]{pandey2021empirical}
{\sc Pandey, S.}, {\sc Karypis, G.}, {\sc and} {\sc Srivastava, J.} 2021.
\newblock An empirical comparison of deep learning models for knowledge tracing
  on large-scale dataset.
\newblock {\em arXiv preprint arXiv:2101.06373\/}.

\bibitem[\protect\citeauthoryear{Pandey and Srivastava}{Pandey and
  Srivastava}{2020}]{pandey2020rkt}
{\sc Pandey, S.} {\sc and} {\sc Srivastava, J.} 2020.
\newblock {RKT}: Relation-aware self-attention for knowledge tracing.
\newblock In {\em Proceedings of the 29th ACM International Conference on
  Information \& Knowledge Management}. 1205--1214.

\bibitem[\protect\citeauthoryear{Pardos, Heffernan, Ruiz, and Beck}{Pardos
  et~al\mbox{.}}{2008}]{pardos2008composition}
{\sc Pardos, Z.}, {\sc Heffernan, N.}, {\sc Ruiz, C.}, {\sc and} {\sc Beck, J.}
  2008.
\newblock The composition effect: Conjuntive or compensatory? an analysis of
  multi-skill math questions in its.
\newblock In {\em Educational Data Mining 2008}.

\bibitem[\protect\citeauthoryear{Pardos and Heffernan}{Pardos and
  Heffernan}{2010}]{pardos2010modeling}
{\sc Pardos, Z.~A.} {\sc and} {\sc Heffernan, N.~T.} 2010.
\newblock Modeling individualization in a bayesian networks implementation of
  knowledge tracing.
\newblock In {\em International Conference on User Modeling, Adaptation, and
  Personalization}. Springer, 255--266.

\bibitem[\protect\citeauthoryear{Pardos and Heffernan}{Pardos and
  Heffernan}{2011}]{pardos2011kt}
{\sc Pardos, Z.~A.} {\sc and} {\sc Heffernan, N.~T.} 2011.
\newblock {KT-IDEM}: Introducing item difficulty to the knowledge tracing
  model.
\newblock In {\em International conference on user modeling, adaptation, and
  personalization}. Springer, 243--254.

\bibitem[\protect\citeauthoryear{Patil, Peng, and Leek}{Patil
  et~al\mbox{.}}{2016}]{patil2016statistical}
{\sc Patil, P.}, {\sc Peng, R.~D.}, {\sc and} {\sc Leek, J.~T.} 2016.
\newblock A statistical definition for reproducibility and replicability.
\newblock {\em BioRxiv\/}, 066803.

\bibitem[\protect\citeauthoryear{{Pavlik Jr}, Cen, and Koedinger}{{Pavlik Jr}
  et~al\mbox{.}}{2009}]{PavlikJr2009}
{\sc {Pavlik Jr}, P.~I.}, {\sc Cen, H.}, {\sc and} {\sc Koedinger, K.~R.} 2009.
\newblock {Performance Factors Analysis–A New Alternative to Knowledge
  Tracing.}
\newblock {\em Online Submission\/}.

\bibitem[\protect\citeauthoryear{Pearl}{Pearl}{1985}]{pearl1985bayesian}
{\sc Pearl, J.} 1985.
\newblock Bayesian netwcrks: A model cf self-activated memory for evidential
  reasoning.
\newblock In {\em Proceedings of the 7th Conference of the Cognitive Science
  Society, University of California, Irvine, CA, USA}. 15--17.

\bibitem[\protect\citeauthoryear{Pearl}{Pearl}{1988}]{pearl1988probabilistic}
{\sc Pearl, J.} 1988.
\newblock {\em Probabilistic Reasoning in Intelligent Systems: Networks of
  Plausible Inference}.
\newblock Morgan Kaufmann.

\bibitem[\protect\citeauthoryear{Pel{\'a}nek}{Pel{\'a}nek}{2014}]{pelanek2014application}
{\sc Pel{\'a}nek, R.} 2014.
\newblock Application of time decay functions and the elo system in student
  modeling.
\newblock In {\em Educational Data Mining 2014}. Citeseer.

\bibitem[\protect\citeauthoryear{Pel{\'a}nek}{Pel{\'a}nek}{2015}]{pelanek2015metrics}
{\sc Pel{\'a}nek, R.} 2015.
\newblock Metrics for evaluation of student models.
\newblock {\em Journal of Educational Data Mining\/}~{\em 7,\/}~2, 1--19.

\bibitem[\protect\citeauthoryear{Peng}{Peng}{2011}]{peng2011}
{\sc Peng, R.~D.} 2011.
\newblock Reproducible research in computational science.
\newblock {\em Science\/}~{\em 334,\/}~6060, 1226--1227.

\bibitem[\protect\citeauthoryear{Piech, Bassen, Huang, Ganguli, Sahami, Guibas,
  and Sohl-Dickstein}{Piech et~al\mbox{.}}{2015}]{Piech2015}
{\sc Piech, C.}, {\sc Bassen, J.}, {\sc Huang, J.}, {\sc Ganguli, S.}, {\sc
  Sahami, M.}, {\sc Guibas, L.~J.}, {\sc and} {\sc Sohl-Dickstein, J.} 2015.
\newblock {Deep knowledge tracing}.
\newblock In {\em Advances in neural information processing systems}. 505--513.

\bibitem[\protect\citeauthoryear{Pu, Yudelson, Ou, and Huang}{Pu
  et~al\mbox{.}}{2020}]{pu2020deep}
{\sc Pu, S.}, {\sc Yudelson, M.}, {\sc Ou, L.}, {\sc and} {\sc Huang, Y.} 2020.
\newblock Deep knowledge tracing with transformers.
\newblock In {\em International Conference on Artificial Intelligence in
  Education}. Springer, 252--256.

\bibitem[\protect\citeauthoryear{Rafferty, Brunskill, Griffiths, and
  Shafto}{Rafferty et~al\mbox{.}}{2011}]{rafferty2011faster}
{\sc Rafferty, A.~N.}, {\sc Brunskill, E.}, {\sc Griffiths, T.~L.}, {\sc and}
  {\sc Shafto, P.} 2011.
\newblock Faster teaching by pomdp planning.
\newblock In {\em International Conference on Artificial Intelligence in
  Education}. Springer, 280--287.

\bibitem[\protect\citeauthoryear{Rowe and Lester}{Rowe and
  Lester}{2010}]{rowe2010modeling}
{\sc Rowe, J.~P.} {\sc and} {\sc Lester, J.~C.} 2010.
\newblock Modeling user knowledge with dynamic bayesian networks in interactive
  narrative environments.
\newblock In {\em Sixth Artificial Intelligence and Interactive Digital
  Entertainment Conference}.

\bibitem[\protect\citeauthoryear{Saad, Prokhorov, and Wunsch}{Saad
  et~al\mbox{.}}{1998}]{saad1998comparative}
{\sc Saad, E.~W.}, {\sc Prokhorov, D.~V.}, {\sc and} {\sc Wunsch, D.~C.} 1998.
\newblock Comparative study of stock trend prediction using time delay,
  recurrent and probabilistic neural networks.
\newblock {\em IEEE Transactions on neural networks\/}~{\em 9,\/}~6,
  1456--1470.

\bibitem[\protect\citeauthoryear{Saha and Raghava}{Saha and
  Raghava}{2006}]{saha2006prediction}
{\sc Saha, S.} {\sc and} {\sc Raghava, G. P.~S.} 2006.
\newblock Prediction of continuous b-cell epitopes in an antigen using
  recurrent neural network.
\newblock {\em Proteins: Structure, Function, and Bioinformatics\/}~{\em
  65,\/}~1, 40--48.

\bibitem[\protect\citeauthoryear{Saito and Rehmsmeier}{Saito and
  Rehmsmeier}{2015}]{saito2015precision}
{\sc Saito, T.} {\sc and} {\sc Rehmsmeier, M.} 2015.
\newblock The precision-recall plot is more informative than the {ROC} plot
  when evaluating binary classifiers on imbalanced datasets.
\newblock {\em PloS one\/}~{\em 10,\/}~3, e0118432.

\bibitem[\protect\citeauthoryear{Sanyal, Bosch, and Paquette}{Sanyal
  et~al\mbox{.}}{2020}]{sanyal2020feature}
{\sc Sanyal, D.}, {\sc Bosch, N.}, {\sc and} {\sc Paquette, L.} 2020.
\newblock Feature selection metrics: Similarities, differences, and
  characteristics of the selected models.
\newblock {\em International Educational Data Mining Society\/}.

\bibitem[\protect\citeauthoryear{Shin, Shim, Yu, Lee, Kim, and Choi}{Shin
  et~al\mbox{.}}{2021}]{shin2021saint+}
{\sc Shin, D.}, {\sc Shim, Y.}, {\sc Yu, H.}, {\sc Lee, S.}, {\sc Kim, B.},
  {\sc and} {\sc Choi, Y.} 2021.
\newblock {SAINT+}: Integrating temporal features for ednet correctness
  prediction.
\newblock In {\em LAK21: 11th International Learning Analytics and Knowledge
  Conference}. 490--496.

\bibitem[\protect\citeauthoryear{Song, Li, Tang, Zhao, Chen, and Guan}{Song
  et~al\mbox{.}}{2021}]{song2021jkt}
{\sc Song, X.}, {\sc Li, J.}, {\sc Tang, Y.}, {\sc Zhao, T.}, {\sc Chen, Y.},
  {\sc and} {\sc Guan, Z.} 2021.
\newblock {JKT}: A joint graph convolutional network based deep knowledge
  tracing.
\newblock {\em Information Sciences\/}~{\em 580}, 510--523.

\bibitem[\protect\citeauthoryear{Spellman}{Spellman}{2012}]{spellman2012}
{\sc Spellman, B.~A.} 2012.
\newblock Introduction to the special section data, data, everywhere...
  especially in my file drawer.
\newblock {\em Perspectives on Psychological Science\/}~{\em 7,\/}~1, 58--59.

\bibitem[\protect\citeauthoryear{Steif and Bier}{Steif and
  Bier}{2014}]{steifoli}
{\sc Steif, P.} {\sc and} {\sc Bier, N.} 2014.
\newblock {OLI} engineering statics-fall 2011, {F}ebruary 2014.

\bibitem[\protect\citeauthoryear{Stevens}{Stevens}{2017}]{stevens2017replicability}
{\sc Stevens, J.~R.} 2017.
\newblock Replicability and reproducibility in comparative psychology.
\newblock {\em Frontiers in psychology\/}~{\em 8}, 862.

\bibitem[\protect\citeauthoryear{Su, Liu, Liu, Huang, Yin, Chen, Ding, Wei, and
  Hu}{Su et~al\mbox{.}}{2018}]{su2018exercise}
{\sc Su, Y.}, {\sc Liu, Q.}, {\sc Liu, Q.}, {\sc Huang, Z.}, {\sc Yin, Y.},
  {\sc Chen, E.}, {\sc Ding, C.}, {\sc Wei, S.}, {\sc and} {\sc Hu, G.} 2018.
\newblock Exercise-enhanced sequential modeling for student performance
  prediction.
\newblock In {\em Proceedings of the AAAI Conference on Artificial
  Intelligence}. Vol.~32.

\bibitem[\protect\citeauthoryear{Sukhbaatar, Weston, Fergus,
  et~al\mbox{.}}{Sukhbaatar et~al\mbox{.}}{2015}]{sukhbaatar2015end}
{\sc Sukhbaatar, S.}, {\sc Weston, J.}, {\sc Fergus, R.}, {\sc et~al\mbox{.}}
  2015.
\newblock End-to-end memory networks.
\newblock In {\em Advances in neural information processing systems}.
  2440--2448.

\bibitem[\protect\citeauthoryear{Trifa, Hedhili, and Chaari}{Trifa
  et~al\mbox{.}}{2019}]{trifa2019knowledge}
{\sc Trifa, A.}, {\sc Hedhili, A.}, {\sc and} {\sc Chaari, W.~L.} 2019.
\newblock Knowledge tracing with an intelligent agent, in an e-learning
  platform.
\newblock {\em Education and Information Technologies\/}~{\em 24,\/}~1,
  711--741.

\bibitem[\protect\citeauthoryear{VanLehn}{VanLehn}{2011}]{vanlehn2011relative}
{\sc VanLehn, K.} 2011.
\newblock The relative effectiveness of human tutoring, intelligent tutoring
  systems, and other tutoring systems.
\newblock {\em Educational Psychologist\/}~{\em 46,\/}~4, 197--221.

\bibitem[\protect\citeauthoryear{Vaswani, Shazeer, Parmar, Uszkoreit, Jones,
  Gomez, Kaiser, and Polosukhin}{Vaswani
  et~al\mbox{.}}{2017}]{vaswani2017attention}
{\sc Vaswani, A.}, {\sc Shazeer, N.}, {\sc Parmar, N.}, {\sc Uszkoreit, J.},
  {\sc Jones, L.}, {\sc Gomez, A.~N.}, {\sc Kaiser, {\L}.}, {\sc and} {\sc
  Polosukhin, I.} 2017.
\newblock Attention is all you need.
\newblock In {\em Advances in neural information processing systems}.
  5998--6008.

\bibitem[\protect\citeauthoryear{Vie and Kashima}{Vie and
  Kashima}{2019}]{vie2019knowledge}
{\sc Vie, J.-J.} {\sc and} {\sc Kashima, H.} 2019.
\newblock Knowledge tracing machines: Factorization machines for knowledge
  tracing.
\newblock In {\em Proceedings of the AAAI Conference on Artificial
  Intelligence}. Vol.~33. 750--757.

\bibitem[\protect\citeauthoryear{Weston, Chopra, and Bordes}{Weston
  et~al\mbox{.}}{2014}]{weston2014memory}
{\sc Weston, J.}, {\sc Chopra, S.}, {\sc and} {\sc Bordes, A.} 2014.
\newblock Memory networks.
\newblock {\em arXiv preprint arXiv:1410.3916\/}.

\bibitem[\protect\citeauthoryear{Wilson, Xiong, Khajah, Lindsey, Zhao, Karklin,
  Van~Inwegen, Han, Ekanadham, Beck, et~al\mbox{.}}{Wilson
  et~al\mbox{.}}{2016}]{wilson2016estimating}
{\sc Wilson, K.~H.}, {\sc Xiong, X.}, {\sc Khajah, M.}, {\sc Lindsey, R.~V.},
  {\sc Zhao, S.}, {\sc Karklin, Y.}, {\sc Van~Inwegen, E.~G.}, {\sc Han, B.},
  {\sc Ekanadham, C.}, {\sc Beck, J.~E.}, {\sc et~al\mbox{.}} 2016.
\newblock Estimating student proficiency: Deep learning is not the panacea.
\newblock In {\em In Neural Information Processing Systems, Workshop on Machine
  Learning for Education}. 3.

\bibitem[\protect\citeauthoryear{Xiong, Zhao, Van~Inwegen, and Beck}{Xiong
  et~al\mbox{.}}{2016}]{xiong2016going}
{\sc Xiong, X.}, {\sc Zhao, S.}, {\sc Van~Inwegen, E.~G.}, {\sc and} {\sc Beck,
  J.~E.} 2016.
\newblock Going deeper with deep knowledge tracing.
\newblock {\em International Educational Data Mining Society\/}.

\bibitem[\protect\citeauthoryear{Yeung and Yeung}{Yeung and
  Yeung}{2018}]{yeung2018addressing}
{\sc Yeung, C.-K.} {\sc and} {\sc Yeung, D.-Y.} 2018.
\newblock Addressing two problems in deep knowledge tracing via
  prediction-consistent regularization.
\newblock In {\em Proceedings of the Fifth Annual ACM Conference on Learning at
  Scale}. 1--10.

\bibitem[\protect\citeauthoryear{Yudelson, Fancsali, Ritter, Berman, Nixon, and
  Joshi}{Yudelson et~al\mbox{.}}{2014}]{yudelson2014better}
{\sc Yudelson, M.}, {\sc Fancsali, S.}, {\sc Ritter, S.}, {\sc Berman, S.},
  {\sc Nixon, T.}, {\sc and} {\sc Joshi, A.} 2014.
\newblock Better data beats big data.
\newblock In {\em Educational Data Mining 2014}. Citeseer.

\bibitem[\protect\citeauthoryear{Yudelson, Koedinger, and Gordon}{Yudelson
  et~al\mbox{.}}{2013}]{yudelson2013individualized}
{\sc Yudelson, M.~V.}, {\sc Koedinger, K.~R.}, {\sc and} {\sc Gordon, G.~J.}
  2013.
\newblock Individualized bayesian knowledge tracing models.
\newblock In {\em International conference on artificial intelligence in
  education}. Springer, 171--180.

\bibitem[\protect\citeauthoryear{Zhang, Shi, King, and Yeung}{Zhang
  et~al\mbox{.}}{2017}]{zhang2017dynamic}
{\sc Zhang, J.}, {\sc Shi, X.}, {\sc King, I.}, {\sc and} {\sc Yeung, D.-Y.}
  2017.
\newblock Dynamic key-value memory networks for knowledge tracing.
\newblock In {\em Proceedings of the 26th international conference on World
  Wide Web}. 765--774.

\bibitem[\protect\citeauthoryear{Zhuang, Qi, Duan, Xi, Zhu, Zhu, Xiong, and
  He}{Zhuang et~al\mbox{.}}{2020}]{zhuang2020comprehensive}
{\sc Zhuang, F.}, {\sc Qi, Z.}, {\sc Duan, K.}, {\sc Xi, D.}, {\sc Zhu, Y.},
  {\sc Zhu, H.}, {\sc Xiong, H.}, {\sc and} {\sc He, Q.} 2020.
\newblock A comprehensive survey on transfer learning.
\newblock {\em Proceedings of the IEEE\/}~{\em 109,\/}~1, 43--76.

\end{thebibliography}

\newpage
\appendix


\section{Model Comparison Results}
\label{appendix:model-comparison-results}

\begin{table}[H]
\footnotesize\centering
\caption{Results for ASSISTments 2009 Updated dataset.}
\label{tbl:results-assistments-2009-updated}
\begin{tabular}{llllllll}
\toprule
      Model &            Acc &            AUC &      Precision &          Recall &             F1 &            MCC &           RMSE \\
\midrule
Vanilla-DKT &      .757±.009 &      .809±.010 &      .766±.021 &       .892±.011 &      .824±.014 &      .448±.020 &      .403±.006 \\
   LSTM-DKT & \bf{.761±.009} & \bf{.814±.011} &      .767±.017 &       .898±.018 & \bf{.827±.016} & \bf{.455±.021} & \bf{.400±.006} \\
LSTM-DKT-S+ & \bf{.761±.011} & \bf{.814±.010} &      .765±.018 &       .901±.018 & \bf{.827±.016} & \bf{.455±.021} & \bf{.400±.006} \\
      DKVMN &      .758±.011 &      .809±.010 &      .765±.020 &       .895±.014 &      .825±.016 &      .450±.018 &      .403±.006 \\
DKVMN-Paper &      .755±.011 &      .806±.010 &      .760±.023 &       .902±.014 &      .825±.015 &      .443±.019 &      .405±.006 \\
       SAKT &      .752±.015 &      .798±.008 &      .759±.017 &       .895±.031 &      .821±.021 &      .434±.015 &      .408±.008 \\
        GLR &      .711±.029 &      .729±.032 &      .715±.033 &       .894±.045 &      .794±.038 &      .324±.074 &      .438±.014 \\
        BKT &      .699±.010 &      .710±.029 &      .716±.033 &       .857±.015 &      .780±.024 &      .313±.039 &      .448±.004 \\
       Mean &      .633±.058 &      .500±.000 &      .633±.058 & \bf{1.000±.000} &      .774±.043 &      .000±.000 &      .484±.019 \\
        NaP &      .713±.025 &      .686±.025 & \bf{.768±.038} &       .776±.035 &      .772±.036 &      .373±.051 &      .535±.023 \\
 NaP 3 Mean &      .681±.022 &      .695±.034 &      .733±.033 &       .771±.039 &      .751±.036 &      .288±.061 &      .485±.016 \\
 NaP 5 Mean &      .697±.026 &      .698±.034 &      .736±.033 &       .803±.050 &      .768±.041 &      .315±.057 &      .470±.015 \\
 NaP 9 Mean &      .694±.028 &      .694±.035 &      .727±.032 &       .816±.058 &      .769±.044 &      .301±.061 &      .463±.015 \\
\bottomrule
\end{tabular}
\end{table}

\begin{table}[H]
\footnotesize\centering
\caption{Results for ASSISTments 2015 dataset.}
\label{tbl:results-assistments-2015}
\begin{tabular}{llllllll}
\toprule
      Model &            Acc &            AUC &      Precision &          Recall &             F1 &            MCC &           RMSE \\
\midrule
Vanilla-DKT &      .749±.028 &      .720±.021 &      .767±.027 &       .943±.011 &      .846±.020 &      .230±.015 &      .414±.019 \\
   LSTM-DKT & \bf{.751±.027} & \bf{.725±.020} &      .769±.025 &       .943±.013 &      .847±.021 & \bf{.239±.013} &      .413±.019 \\
LSTM-DKT-S+ & \bf{.751±.027} & \bf{.725±.020} &      .769±.026 &       .943±.013 &      .847±.020 &      .238±.015 & \bf{.412±.019} \\
      DKVMN &      .750±.028 &      .723±.020 &      .769±.026 &       .941±.013 &      .846±.021 & \bf{.239±.015} &      .413±.019 \\
DKVMN-Paper &      .750±.027 &      .718±.022 &      .769±.028 &       .943±.008 &      .847±.020 &      .237±.017 &      .414±.019 \\
       SAKT &      .748±.028 &      .714±.023 &      .767±.029 &       .942±.007 &      .846±.020 &      .230±.021 &      .415±.019 \\
        GLR &      .750±.031 &      .702±.027 &      .763±.031 &       .959±.010 & \bf{.849±.022} &      .204±.017 &      .416±.021 \\
        BKT &      .747±.026 &      .694±.020 &      .761±.026 &       .955±.009 &      .847±.018 &      .209±.012 &      .421±.015 \\
       Mean &      .738±.033 &      .500±.000 &      .738±.033 & \bf{1.000±.000} & \bf{.849±.022} &      .000±.000 &      .440±.017 \\
        NaP &      .690±.036 &      .594±.013 & \bf{.786±.032} &       .792±.034 &      .789±.033 &      .190±.026 &      .556±.032 \\
 NaP 3 Mean &      .701±.036 &      .602±.024 &      .778±.033 &       .830±.029 &      .803±.031 &      .176±.027 &      .484±.026 \\
 NaP 5 Mean &      .704±.035 &      .624±.024 &      .771±.033 &       .850±.025 &      .808±.029 &      .158±.025 &      .463±.023 \\
 NaP 9 Mean &      .712±.033 &      .624±.023 &      .769±.032 &       .868±.022 &      .816±.027 &      .162±.022 &      .454±.021 \\
\bottomrule
\end{tabular}
\end{table}

\begin{table}[H]
\footnotesize\centering
\caption{Results for ASSISTments 2017 dataset.}
\label{tbl:results-assistments-2017}
\begin{tabular}{llllllll}
\toprule
      Model &            Acc &            AUC &      Precision &         Recall &             F1 &            MCC &           RMSE \\
\midrule
Vanilla-DKT &      .681±.015 &      .703±.009 &      .615±.028 &      .374±.049 &      .464±.044 &      .271±.020 &      .453±.008 \\
   LSTM-DKT & \bf{.692±.017} & \bf{.723±.010} & \bf{.630±.019} &      .412±.039 &      .498±.035 & \bf{.302±.016} & \bf{.446±.009} \\
LSTM-DKT-S+ & \bf{.692±.016} & \bf{.723±.010} &      .626±.019 &      .420±.040 & \bf{.502±.035} & \bf{.302±.014} & \bf{.446±.009} \\
      DKVMN &      .680±.011 &      .704±.009 &      .611±.037 &      .377±.044 &      .466±.045 &      .270±.026 &      .452±.006 \\
DKVMN-Paper &      .678±.012 &      .696±.009 &      .617±.031 &      .342±.071 &      .438±.068 &      .257±.032 &      .454±.006 \\
       SAKT &      .672±.014 &      .661±.022 &      .617±.026 &      .298±.082 &      .397±.084 &      .235±.036 &      .462±.004 \\
        GLR &      .659±.004 &      .648±.005 &      .602±.008 &      .254±.007 &      .357±.009 &      .205±.007 &      .467±.002 \\
        BKT &      .645±.016 &      .623±.002 &      .565±.020 &      .279±.018 &      .373±.013 &      .183±.009 &      .475±.006 \\
       Mean &      .627±.005 &      .500±.000 &      .000±.000 &      .000±.000 &      .000±.000 &      .000±.000 &      .484±.001 \\
        NaP &      .591±.001 &      .562±.002 &      .451±.007 & \bf{.450±.007} &      .450±.007 &      .124±.004 &      .640±.001 \\
 NaP 3 Mean &      .630±.005 &      .573±.002 &      .505±.004 &      .414±.007 &      .455±.005 &      .182±.005 &      .534±.002 \\
 NaP 5 Mean &      .637±.004 &      .600±.002 &      .518±.005 &      .387±.008 &      .443±.006 &      .187±.005 &      .503±.002 \\
 NaP 9 Mean &      .641±.004 &      .608±.003 &      .528±.005 &      .356±.008 &      .425±.007 &      .186±.006 &      .488±.001 \\
\bottomrule
\end{tabular}
\end{table}
\begin{table}[H]
\footnotesize\centering
\caption{Results for IntroProg dataset.}
\label{tbl:results-introprog}
\begin{tabular}{llllllll}
\toprule
      Model &            Acc &            AUC &      Precision &         Recall &             F1 &            MCC &           RMSE \\
\midrule
Vanilla-DKT &      .752±.028 &      .821±.012 &      .756±.036 &      .746±.048 &      .751±.042 &      .483±.029 &      .410±.016 \\
   LSTM-DKT &      .757±.027 &      .827±.013 &      .754±.032 &      .762±.056 &      .758±.044 &      .491±.030 &      .406±.016 \\
LSTM-DKT-S+ &      .755±.027 &      .826±.014 &      .752±.030 &      .761±.060 &      .756±.045 &      .487±.029 &      .406±.017 \\
      DKVMN &      .756±.028 &      .827±.015 & \bf{.757±.030} &      .755±.055 &      .756±.043 &      .490±.031 &      .406±.018 \\
DKVMN-Paper &      .754±.027 &      .825±.013 &      .752±.033 &      .759±.058 &      .755±.045 &      .486±.029 &      .407±.016 \\
       SAKT &      .754±.029 &      .825±.015 &      .743±.023 & \bf{.776±.068} & \bf{.759±.045} &      .482±.034 &      .407±.020 \\
        GLR & \bf{.761±.007} & \bf{.843±.008} &      .754±.002 &      .757±.008 &      .755±.004 & \bf{.522±.014} & \bf{.402±.005} \\
        BKT &      .713±.012 &      .789±.012 &      .738±.087 &      .684±.011 &      .708±.041 &      .423±.012 &      .436±.004 \\
       Mean &      .513±.013 &      .500±.000 &      .000±.000 &      .000±.000 &      .000±.000 &      .000±.000 &      .500±.000 \\
        NaP &      .716±.006 &      .716±.006 &      .708±.005 &      .710±.005 &      .709±.005 &      .431±.012 &      .533±.006 \\
 NaP 3 Mean &      .708±.008 &      .767±.009 &      .696±.005 &      .713±.006 &      .704±.006 &      .416±.017 &      .466±.006 \\
 NaP 5 Mean &      .707±.008 &      .772±.010 &      .693±.006 &      .717±.006 &      .704±.006 &      .414±.017 &      .456±.006 \\
 NaP 9 Mean &      .705±.009 &      .774±.012 &      .687±.006 &      .723±.006 &      .704±.006 &      .410±.018 &      .450±.007 \\
\bottomrule
\end{tabular}
\end{table}

\begin{table}[H]
\footnotesize\centering
\caption{Results for Statics dataset.}
\label{tbl:results-statics}
\begin{tabular}{llllllll}
\toprule
      Model &            Acc &            AUC &      Precision &          Recall &             F1 &            MCC &           RMSE \\
\midrule
Vanilla-DKT &      .804±.017 &      .815±.014 &      .828±.017 &       .938±.013 &      .880±.014 &      .372±.021 &      .369±.015 \\
   LSTM-DKT & \bf{.807±.016} &      .824±.013 &      .832±.014 &       .937±.017 & \bf{.881±.014} &      .383±.030 &      .365±.013 \\
LSTM-DKT-S+ &      .806±.015 &      .823±.013 &      .832±.014 &       .936±.018 & \bf{.881±.013} &      .383±.034 &      .365±.013 \\
      DKVMN & \bf{.807±.017} & \bf{.825±.012} & \bf{.835±.017} &       .932±.011 & \bf{.881±.014} & \bf{.393±.023} & \bf{.364±.014} \\
DKVMN-Paper &      .803±.016 &      .812±.013 &      .831±.015 &       .933±.013 &      .879±.013 &      .374±.028 &      .369±.013 \\
       SAKT &      .801±.018 &      .808±.017 &      .827±.019 &       .936±.012 &      .878±.014 &      .363±.027 &      .371±.015 \\
        GLR &      .805±.006 &      .815±.004 &      .826±.005 &       .943±.006 & \bf{.881±.005} &      .378±.007 &      .369±.004 \\
        BKT &      .786±.022 &      .774±.028 &      .803±.028 &       .953±.012 &      .872±.013 &      .305±.070 &      .393±.013 \\
       Mean &      .765±.007 &      .500±.000 &      .765±.007 & \bf{1.000±.000} &      .867±.005 &      .000±.000 &      .424±.004 \\
        NaP &      .705±.004 &      .589±.005 &      .807±.004 &       .808±.004 &      .808±.004 &      .179±.010 &      .543±.004 \\
 NaP 3 Mean &      .729±.005 &      .640±.007 &      .801±.004 &       .859±.005 &      .829±.004 &      .180±.015 &      .453±.003 \\
 NaP 5 Mean &      .740±.004 &      .652±.008 &      .797±.003 &       .886±.005 &      .839±.004 &      .176±.018 &      .435±.003 \\
 NaP 9 Mean &      .751±.005 &      .661±.009 &      .791±.004 &       .916±.007 &      .849±.005 &      .170±.019 &      .422±.003 \\
\bottomrule
\end{tabular}
\end{table}

\begin{table}[H]
\footnotesize\centering
\caption{Results for Synthetic-K2 dataset.}
\label{tbl:results-synthetic-k2}
\begin{tabular}{llllllll}
\toprule
      Model &            Acc &            AUC &      Precision &          Recall &             F1 &            MCC &           RMSE \\
\midrule
Vanilla-DKT &      .804±.003 &      .869±.003 &      .851±.004 &       .867±.007 &      .858±.003 &      .542±.007 &      .367±.002 \\
   LSTM-DKT &      .805±.003 &      .871±.002 &      .848±.006 &       .872±.004 & \bf{.860±.002} &      .541±.008 &      .366±.002 \\
LSTM-DKT-S+ & \bf{.806±.003} &      .871±.002 &      .851±.005 &       .868±.003 &      .859±.002 &      .545±.006 &      .366±.002 \\
      DKVMN & \bf{.806±.003} & \bf{.872±.002} &      .852±.003 &       .867±.003 & \bf{.860±.002} &      .546±.005 & \bf{.365±.002} \\
DKVMN-Paper & \bf{.806±.002} & \bf{.872±.002} &      .853±.003 &       .865±.002 &      .859±.002 &      .547±.005 & \bf{.365±.002} \\
       SAKT & \bf{.806±.002} & \bf{.872±.002} & \bf{.855±.006} &       .862±.005 &      .859±.001 & \bf{.548±.006} & \bf{.365±.002} \\
        GLR &      .729±.004 &      .786±.002 &      .750±.003 &       .906±.005 &      .821±.003 &      .308±.013 &      .418±.002 \\
        BKT &      .692±.004 &      .635±.005 &      .700±.005 &       .961±.000 &      .810±.003 &      .132±.002 &      .455±.002 \\
       Mean &      .685±.005 &      .500±.000 &      .685±.005 & \bf{1.000±.000} &      .813±.003 &      .000±.000 &      .465±.002 \\
        NaP &      .644±.001 &      .585±.003 &      .738±.002 &       .744±.002 &      .741±.002 &      .171±.006 &      .597±.001 \\
 NaP 3 Mean &      .680±.003 &      .654±.004 &      .753±.002 &       .794±.004 &      .773±.003 &      .235±.008 &      .489±.002 \\
 NaP 5 Mean &      .688±.003 &      .681±.004 &      .754±.002 &       .808±.004 &      .780±.003 &      .247±.010 &      .465±.001 \\
 NaP 9 Mean &      .707±.003 &      .709±.005 &      .764±.003 &       .827±.006 &      .794±.004 &      .289±.008 &      .448±.001 \\
\bottomrule
\end{tabular}
\end{table}

\begin{table}[H]
\footnotesize\centering
\caption{Results for Synthetic-K5 dataset.}
\label{tbl:results-synthetic-k5}
\begin{tabular}{llllllll}
\toprule
      Model &            Acc &            AUC &      Precision &          Recall &             F1 &            MCC &           RMSE \\
\midrule
Vanilla-DKT &      .750±.002 &      .822±.002 &      .805±.003 &       .778±.004 &      .791±.003 &      .481±.004 &      .409±.001 \\
   LSTM-DKT &      .752±.003 &      .827±.002 &      .807±.004 &       .779±.003 & \bf{.793±.003} &      .485±.005 &      .406±.002 \\
LSTM-DKT-S+ &      .751±.002 &      .826±.002 &      .805±.003 &       .779±.008 &      .792±.004 &      .483±.004 &      .406±.001 \\
      DKVMN & \bf{.754±.003} & \bf{.829±.002} & \bf{.812±.003} &       .775±.005 & \bf{.793±.003} & \bf{.491±.006} & \bf{.404±.002} \\
DKVMN-Paper &      .753±.003 & \bf{.829±.003} & \bf{.812±.003} &       .773±.007 &      .792±.004 &      .490±.006 &      .405±.002 \\
       SAKT &      .753±.003 &      .828±.002 &      .809±.008 &       .777±.004 & \bf{.793±.002} &      .488±.008 &      .405±.002 \\
        GLR &      .649±.004 &      .683±.006 &      .666±.004 &       .848±.005 &      .746±.002 &      .220±.011 &      .465±.001 \\
        BKT &      .633±.002 &      .633±.002 &      .641±.004 &       .899±.010 &      .748±.002 &      .164±.006 &      .476±.000 \\
       Mean &      .608±.003 &      .500±.000 &      .608±.003 & \bf{1.000±.000} &      .756±.002 &      .000±.000 &      .488±.001 \\
        NaP &      .565±.004 &      .543±.003 &      .641±.004 &       .647±.004 &      .644±.004 &      .086±.007 &      .659±.003 \\
 NaP 3 Mean &      .579±.003 &      .560±.004 &      .645±.004 &       .682±.006 &      .663±.005 &      .103±.005 &      .551±.002 \\
 NaP 5 Mean &      .596±.004 &      .581±.005 &      .655±.004 &       .710±.006 &      .681±.005 &      .133±.009 &      .522±.002 \\
 NaP 9 Mean &      .610±.005 &      .605±.006 &      .664±.004 &       .725±.007 &      .693±.005 &      .163±.010 &      .504±.002 \\
\bottomrule
\end{tabular}
\end{table}


\newpage

\section{Best Hyperparameters}
\label{appendix:best-hyperparameters}


The layer sizes in the tables are as follows: recurrent layer size (attention layer size for SAKT), key-embedding layer size, value-embedding layer size and summary layer size. Hyperparameters that are not used for a model are denoted by the dash symbol ``-''. As an example, if a model has no key-embeddings or a summary layer, recurrent layer size of 100 and value-embedding layer size of 20, the layer sizes are shown as 100,-,20,-.

\begin{table}[H]
\footnotesize\centering
\caption{Best hyperparameters for DLKT models in ASSISTments 2009 Updated dataset}
\label{tbl:best-hyperparams-assistments-2009-updated}
\begin{tabular}{lllllll}
\toprule
Model            &        DKVMN &   DKVMN-Paper &    LSTM-DKT &  LSTM-DKT-S+ &         SAKT &  Vanilla-DKT \\
\midrule
AUC              &        0.809 &         0.806 &       0.814 &        0.814 &        0.798 &        0.809 \\
Seed             &           13 &            13 &          42 &           42 &           13 &           13 \\
Init lr          &        0.001 &          0.01 &       0.001 &        0.001 &        0.001 &        0.001 \\
Layer sizes      &  50,20,50,-, &  50,20,20,100 &  50,-,20,-, &   50,-,20,-, &  50,50,50,50 &  100,-,50,-, \\
N heads          &            - &             - &           - &            - &            5 &            - \\
One-hot input     &         True &         False &       False &        False &        False &        False \\
Output per skill &         True &         False &        True &         True &        False &         True \\
\bottomrule
\end{tabular}
\end{table}
\begin{table}[H]
\footnotesize\centering
\caption{Best hyperparameters for DLKT models in ASSISTments 2015 dataset}
\label{tbl:best-hyperparams-assistments-2015}
\begin{tabular}{lllllll}
\toprule
Model            &        DKVMN &   DKVMN-Paper &     LSTM-DKT &  LSTM-DKT-S+ &        SAKT &  Vanilla-DKT \\
\midrule
AUC              &        0.723 &         0.718 &        0.725 &        0.725 &       0.714 &         0.72 \\
Seed             &           13 &            42 &           13 &           13 &          13 &           42 \\
Init lr          &        0.001 &          0.01 &        0.001 &        0.001 &       0.001 &        0.001 \\
Layer sizes      &  50,20,20,-, &  50,50,50,100 &  100,-,50,-, &  100,-,50,-, &  100,-,-,50 &   50,-,50,-, \\
N heads          &            - &             - &            - &            - &           5 &            - \\
One-hot input     &        False &         False &        False &        False &        True &        False \\
Output per skill &         True &         False &         True &         True &       False &         True \\
\bottomrule
\end{tabular}
\end{table}
\begin{table}[H]
\footnotesize\centering
\caption{Best hyperparameters for DLKT models in ASSISTments 2017 dataset}
\label{tbl:best-hyperparams-assistments-2017}
\begin{tabular}{lllllll}
\toprule
Model            &          DKVMN &   DKVMN-Paper &     LSTM-DKT &  LSTM-DKT-S+ &         SAKT &  Vanilla-DKT \\
\midrule
AUC              &          0.704 &         0.696 &        0.723 &        0.723 &        0.661 &        0.703 \\
Seed             &             13 &            42 &           42 &           42 &           42 &           42 \\
Init lr          &           0.01 &          0.01 &         0.01 &         0.01 &        0.001 &        0.001 \\
Layer sizes      &  100,20,50,100 &  50,50,50,100 &  100,-,20,-, &  100,-,20,-, &  50,20,20,-, &  100,-,50,-, \\
N heads          &              - &             - &            - &            - &            5 &            - \\
One-hot input     &           True &         False &        False &        False &        False &        False \\
Output per skill &          False &         False &         True &         True &         True &         True \\
\bottomrule
\end{tabular}
\end{table}
\begin{table}[H]
\footnotesize\centering
\caption{Best hyperparameters for DLKT models in IntroProg dataset}
\label{tbl:best-hyperparams-introprog}
\begin{tabular}{lllllll}
\toprule
Model            &        DKVMN &  DKVMN-Paper &    LSTM-DKT &  LSTM-DKT-S+ &        SAKT &  Vanilla-DKT \\
\midrule
AUC              &        0.827 &        0.825 &       0.827 &        0.826 &       0.825 &        0.821 \\
Seed             &           13 &           13 &          42 &           42 &          13 &           42 \\
Init lr          &        0.001 &         0.01 &       0.001 &         0.01 &       0.001 &        0.001 \\
Layer sizes      &  50,20,20,-, &  50,50,20,50 &  50,-,50,-, &   50,-,20,-, &  50,-,-,100 &   50,-,20,-, \\
N heads          &            - &            - &           - &            - &           1 &            - \\
One-hot input     &        False &        False &       False &        False &        True &        False \\
Output per skill &         True &        False &        True &         True &       False &         True \\
\bottomrule
\end{tabular}
\end{table}
\begin{table}[H]
\footnotesize\centering
\caption{Best hyperparameters for DLKT models in Statics dataset}
\label{tbl:best-hyperparams-statics}
\begin{tabular}{lllllll}
\toprule
Model            &        DKVMN &  DKVMN-Paper &   LSTM-DKT &  LSTM-DKT-S+ &       SAKT &  Vanilla-DKT \\
\midrule
AUC              &        0.825 &        0.812 &      0.824 &        0.823 &      0.808 &        0.815 \\
Seed             &           13 &           13 &         13 &           13 &         42 &           42 \\
Init lr          &         0.01 &         0.01 &      0.001 &        0.001 &      0.001 &        0.001 \\
Layer sizes      &  50,20,20,-, &  50,50,50,-, &  50,-,-,-, &    50,-,-,-, &  50,-,-,-, &  100,-,50,-, \\
N heads          &            - &            - &          - &            - &          1 &            - \\
One-hot input     &         True &         True &       True &         True &       True &        False \\
Output per skill &         True &         True &       True &         True &       True &         True \\
\bottomrule
\end{tabular}
\end{table}
\begin{table}[H]
\footnotesize\centering
\caption{Best hyperparameters for DLKT models in Synthetic-K2 dataset}
\label{tbl:best-hyperparams-synthetic-k2}
\begin{tabular}{lllllll}
\toprule
Model            &        DKVMN &  DKVMN-Paper &    LSTM-DKT &  LSTM-DKT-S+ &           SAKT &  Vanilla-DKT \\
\midrule
AUC              &        0.872 &        0.872 &       0.871 &        0.871 &          0.872 &        0.869 \\
Seed             &           42 &           42 &          42 &           42 &             42 &           42 \\
Init lr          &        0.001 &        0.001 &       0.001 &        0.001 &          0.001 &        0.001 \\
Layer sizes      &  50,50,20,50 &  50,50,20,50 &  50,-,-,100 &  50,-,20,100 &  100,20,20,100 &   100,-,-,-, \\
N heads          &            - &            - &           - &            - &              5 &            - \\
One-hot input     &         True &        False &        True &        False &          False &         True \\
Output per skill &        False &        False &       False &        False &          False &         True \\
\bottomrule
\end{tabular}
\end{table}
\begin{table}[H]
\footnotesize\centering
\caption{Best hyperparameters for DLKT models in Synthetic-K5 dataset}
\label{tbl:best-hyperparams-synthetic-k5}
\begin{tabular}{lllllll}
\toprule
Model            &        DKVMN &  DKVMN-Paper &   LSTM-DKT &  LSTM-DKT-S+ &           SAKT &  Vanilla-DKT \\
\midrule
AUC              &        0.829 &        0.829 &      0.827 &        0.826 &          0.828 &        0.822 \\
Seed             &           42 &           13 &         42 &           13 &             42 &           13 \\
Init lr          &        0.001 &        0.001 &      0.001 &        0.001 &          0.001 &        0.001 \\
Layer sizes      &  50,50,20,50 &  50,50,50,50 &  50,-,-,-, &    50,-,-,-, &  100,20,20,100 &  100,-,20,-, \\
N heads          &            - &            - &          - &            - &              5 &            - \\
One-hot input     &         True &        False &       True &         True &          False &        False \\
Output per skill &        False &        False &       True &         True &          False &         True \\
\bottomrule
\end{tabular}
\end{table}

\begin{table}[H]
\footnotesize\centering
\caption{Output-per-skill effect on AUC in ASSISTments 2017 dataset}
\label{tbl:assistments-2017-model-output-per-skill}
\begin{tabular}{llrrrr}
\toprule
Model & Output-per-skill & Max &  Max-Min &    Min &    Sd \\
\midrule
DKVMN & False &  0.704 &    0.038 &  0.666 &  0.012 \\
            & True &  0.685 &    0.011 &  0.674 &  0.002 \\
DKVMN-Paper & False &  0.696 &    0.058 &  0.638 &  0.019 \\
            & True &  0.658 &    0.012 &  0.646 &  0.002 \\
LSTM-DKT & False &  0.711 &    0.048 &  0.663 &  0.010 \\
            & True &  0.723 &    0.010 &  0.713 &  0.002 \\
LSTM-DKT-S+ & False &  0.718 &    0.032 &  0.686 &  0.005 \\
            & True &  0.723 &    0.007 &  0.716 &  0.002 \\
SAKT & False &  0.659 &    0.138 &  0.521 &  0.051 \\
            & True &  0.661 &    0.038 &  0.623 &  0.005 \\
Vanilla-DKT & False &  0.683 &    0.126 &  0.557 &  0.036 \\
            & True &  0.703 &    0.090 &  0.613 &  0.029 \\
\bottomrule
\end{tabular}
\end{table}

\begin{table}[H]
\footnotesize\centering
\caption{Output-per-skill effect on AUC in ASSISTments 2015 dataset}
\label{tbl:assistments-2015-model-output-per-skill}
\begin{tabular}{llrrrr}
\toprule
Model & Output-per-skill & Max &  Max-Min &    Min &    Sd \\
\midrule
DKVMN & False &  0.720 &    0.006 &  0.714 &  0.001 \\
            & True &  0.723 &    0.009 &  0.714 &  0.003 \\
DKVMN-Paper & False &  0.718 &    0.032 &  0.686 &  0.010 \\
            & True &  0.716 &    0.029 &  0.687 &  0.006 \\
LSTM-DKT & False &  0.703 &    0.020 &  0.683 &  0.003 \\
            & True &  0.725 &    0.005 &  0.720 &  0.002 \\
LSTM-DKT-S+ & False &  0.720 &    0.016 &  0.704 &  0.003 \\
            & True &  0.725 &    0.005 &  0.720 &  0.002 \\
SAKT & False &  0.714 &    0.194 &  0.520 &  0.062 \\
            & True &  0.713 &    0.060 &  0.653 &  0.013 \\
Vanilla-DKT & False &  0.699 &    0.076 &  0.623 &  0.021 \\
            & True &  0.720 &    0.056 &  0.664 &  0.019 \\
\bottomrule
\end{tabular}
\end{table}

\begin{table}[H]
\footnotesize\centering
\caption{One-hot-input effect on AUC in IntroProg dataset}
\label{tbl:introprog-model-one-hot-input}
\begin{tabular}{llrrrr}
\toprule
Model & One-hot-input & Max &  Max-Min &    Min &    Sd \\
\midrule
DKVMN & False &  0.827 &    0.003 &  0.824 &  0.001 \\
            & True &  0.827 &    0.003 &  0.824 &  0.001 \\
DKVMN-Paper & False &  0.825 &    0.024 &  0.801 &  0.005 \\
            & True &  0.818 &    0.012 &  0.806 &  0.003 \\
LSTM-DKT & False &  0.827 &    0.007 &  0.820 &  0.002 \\
            & True &  0.826 &    0.007 &  0.819 &  0.003 \\
LSTM-DKT-S+ & False &  0.826 &    0.002 &  0.824 &  0.001 \\
            & True &  0.826 &    0.003 &  0.823 &  0.001 \\
SAKT & False &  0.817 &    0.265 &  0.552 &  0.055 \\
            & True &  0.825 &    0.219 &  0.606 &  0.065 \\
Vanilla-DKT & False &  0.821 &    0.102 &  0.719 &  0.017 \\
            & True &  0.821 &    0.088 &  0.733 &  0.018 \\
\bottomrule
\end{tabular}
\end{table}

\begin{table}[H]
\footnotesize\centering
\caption{One-hot input effect on AUC in ASSISTments 2009 Updated dataset}
\label{tbl:assistments-2009-updated-model-one-hot-input}
\begin{tabular}{llrrrr}
\toprule
Model & One-hot-input & Max &  Max-Min &    Min &    Sd \\
\midrule
DKVMN & False &  0.809 &    0.010 &  0.799 &  0.003 \\
            & True &  0.809 &    0.010 &  0.799 &  0.004 \\
DKVMN-Paper & False &  0.806 &    0.094 &  0.712 &  0.035 \\
            & True &  0.773 &    0.062 &  0.711 &  0.017 \\
LSTM-DKT & False &  0.814 &    0.077 &  0.737 &  0.013 \\
            & True &  0.814 &    0.026 &  0.788 &  0.011 \\
LSTM-DKT-S+ & False &  0.814 &    0.072 &  0.742 &  0.010 \\
            & True &  0.814 &    0.014 &  0.800 &  0.005 \\
SAKT & False &  0.798 &    0.253 &  0.545 &  0.071 \\
            & True &  0.752 &    0.207 &  0.545 &  0.074 \\
Vanilla-DKT & False &  0.809 &    0.098 &  0.711 &  0.022 \\
            & True &  0.806 &    0.101 &  0.705 &  0.023 \\
\bottomrule
\end{tabular}
\end{table}

\begin{table}[H]
\footnotesize\centering
\caption{Random seed effect on AUC in ASSISTments 2015 dataset}
\label{tbl:assistments-2015-model-seed}
\begin{tabular}{llrrrr}
\toprule
Model & Random seed & Max &  Max-Min &    Min &    Sd \\
\midrule
DKVMN & 13 &  0.723 &    0.009 &  0.714 &  0.002 \\
            & 42 &  0.723 &    0.009 &  0.714 &  0.002 \\
DKVMN-Paper & 13 &  0.718 &    0.031 &  0.687 &  0.009 \\
            & 42 &  0.718 &    0.032 &  0.686 &  0.009 \\
LSTM-DKT & 13 &  0.725 &    0.042 &  0.683 &  0.012 \\
            & 42 &  0.724 &    0.027 &  0.697 &  0.011 \\
LSTM-DKT-S+ & 13 &  0.725 &    0.021 &  0.704 &  0.004 \\
            & 42 &  0.725 &    0.018 &  0.707 &  0.003 \\
SAKT & 13 &  0.714 &    0.176 &  0.538 &  0.049 \\
            & 42 &  0.713 &    0.193 &  0.520 &  0.062 \\
Vanilla-DKT & 13 &  0.719 &    0.096 &  0.623 &  0.024 \\
            & 42 &  0.720 &    0.091 &  0.629 &  0.022 \\
\bottomrule
\end{tabular}
\end{table}

\newpage

\section{Max attempt filter cut versus split}
\label{appendix:max-attempt-count}
\begin{footnotesize}
\centering
\begin{longtable}{lll|llllllll}\label{tbl:split-vs-cut}\\
\caption{Student attempt split effect}\\ 
\toprule
Dataset & Model & Max attempt filter & Acc &   AUC &  Prec & Recall &    F1 &   MCC &  RMSE \\
\midrule
\endhead
\multirow{2}{8em}{ASSISTments 2009 Updated} 
& DKVMN & - \& - &  .759 &  .812 &  .763 &   .890 &  .822 &  .454 &  .401 \\
             &             & Cut \& 200 &  .753 &  .809 &  .758 &   .880 &  .814 &  .452 &  .406 \\
             &             & Cut \& 500 &  .758 &  .811 &  .762 &   .889 &  .821 &  .452 &  .402 \\
             &             & Split \& 200 &  .758 &  .809 &  .765 &   .895 &  .825 &  .450 &  .403 \\
             &             & Split \& 500 &  .758 &  .811 &  .762 &   .891 &  .821 &  .452 &  .403 \\
             & DKVMN-Paper & - \& - &  .745 &  .788 &  .750 &   .889 &  .814 &  .413 &  .411 \\
             &             & Cut \& 200 &  .749 &  .802 &  .750 &   .886 &  .812 &  .441 &  .409 \\
             &             & Cut \& 500 &  .756 &  .806 &  .759 &   .891 &  .820 &  .446 &  .404 \\
             &             & Split \& 200 &  .755 &  .806 &  .760 &   .902 &  .825 &  .443 &  .405 \\
             &             & Split \& 500 &  .754 &  .804 &  .759 &   .888 &  .818 &  .441 &  .406 \\
             & LSTM-DKT & - \& - &  .763 &  .817 &  .766 &   .892 &  .824 &  .463 &  .398 \\
             &             & Cut \& 200 &  .756 &  .813 &  .761 &   .878 &  .815 &  .458 &  .404 \\
             &             & Cut \& 500 &  .763 &  .816 &  .768 &   .888 &  .823 &  .463 &  .399 \\
             &             & Split \& 200 &  .761 &  .814 &  .767 &   .898 &  .827 &  .455 &  .400 \\
             &             & Split \& 500 &  .759 &  .815 &  .764 &   .889 &  .822 &  .455 &  .401 \\
             & LSTM-DKT-S+ & - \& - &  .762 &  .817 &  .765 &   .893 &  .824 &  .462 &  .398 \\
             &             & Cut \& 200 &  .756 &  .813 &  .761 &   .878 &  .815 &  .457 &  .404 \\
             &             & Cut \& 500 &  .762 &  .815 &  .765 &   .892 &  .824 &  .460 &  .399 \\
             &             & Split \& 200 &  .761 &  .814 &  .765 &   .901 &  .827 &  .455 &  .400 \\
             &             & Split \& 500 &  .759 &  .815 &  .764 &   .888 &  .821 &  .455 &  .401 \\
             & SAKT & - \& - &  .743 &  .787 &  .756 &   .865 &  .807 &  .416 &  .415 \\
             &             & Cut \& 200 &  .744 &  .795 &  .748 &   .879 &  .808 &  .430 &  .412 \\
             &             & Cut \& 500 &  .746 &  .790 &  .756 &   .871 &  .809 &  .424 &  .413 \\
             &             & Split \& 200 &  .752 &  .798 &  .759 &   .895 &  .821 &  .434 &  .408 \\
             &             & Split \& 500 &  .746 &  .792 &  .754 &   .880 &  .812 &  .423 &  .413 \\
             & Vanilla-DKT & - \& - &  .758 &  .810 &  .761 &   .893 &  .822 &  .451 &  .403 \\
             &             & Cut \& 200 &  .752 &  .808 &  .758 &   .876 &  .813 &  .449 &  .407 \\
             &             & Cut \& 500 &  .757 &  .808 &  .759 &   .893 &  .821 &  .449 &  .404 \\
             &             & Split \& 200 &  .757 &  .809 &  .766 &   .892 &  .824 &  .448 &  .403 \\
             &             & Split \& 500 &  .753 &  .806 &  .761 &   .882 &  .817 &  .441 &  .406 \\
ASSISTments 2015 & DKVMN & - \& - &  .755 &  .725 &  .772 &   .945 &  .850 &  .240 &  .410 \\
             &             & Cut \& 200 &  .754 &  .725 &  .772 &   .945 &  .850 &  .237 &  .411 \\
             &             & Cut \& 500 &  .755 &  .725 &  .772 &   .945 &  .850 &  .240 &  .410 \\
             &             & Split \& 200 &  .750 &  .723 &  .769 &   .941 &  .846 &  .239 &  .413 \\
             &             & Split \& 500 &  .754 &  .725 &  .772 &   .945 &  .849 &  .238 &  .411 \\
             & DKVMN-Paper & - \& - &  .753 &  .719 &  .771 &   .946 &  .849 &  .233 &  .412 \\
             &             & Cut \& 200 &  .753 &  .718 &  .772 &   .943 &  .849 &  .236 &  .413 \\
             &             & Cut \& 500 &  .753 &  .719 &  .773 &   .941 &  .848 &  .238 &  .413 \\
             &             & Split \& 200 &  .750 &  .718 &  .769 &   .943 &  .847 &  .237 &  .414 \\
             &             & Split \& 500 &  .753 &  .719 &  .772 &   .943 &  .849 &  .238 &  .413 \\
             & LSTM-DKT & - \& - &  .755 &  .727 &  .772 &   .945 &  .850 &  .241 &  .410 \\
             &             & Cut \& 200 &  .755 &  .727 &  .772 &   .946 &  .850 &  .240 &  .410 \\
             &             & Cut \& 500 &  .754 &  .727 &  .772 &   .945 &  .850 &  .241 &  .410 \\
             &             & Split \& 200 &  .751 &  .725 &  .769 &   .943 &  .847 &  .239 &  .413 \\
             &             & Split \& 500 &  .754 &  .727 &  .772 &   .944 &  .850 &  .241 &  .410 \\
             & LSTM-DKT-S+ & - \& - &  .754 &  .727 &  .772 &   .946 &  .850 &  .239 &  .410 \\
             &             & Cut \& 200 &  .754 &  .727 &  .772 &   .946 &  .850 &  .238 &  .410 \\
             &             & Cut \& 500 &  .754 &  .727 &  .772 &   .945 &  .850 &  .240 &  .410 \\
ASSISTments 2015  &  LSTM-DKT-S+            & Split \& 200 &  .751 &  .725 &  .769 &   .943 &  .847 &  .238 &  .412 \\
             &             & Split \& 500 &  .755 &  .726 &  .772 &   .945 &  .850 &  .241 &  .410 \\
             & SAKT & - \& - &  .752 &  .715 &  .769 &   .948 &  .849 &  .225 &  .413 \\
             &             & Cut \& 200 &  .753 &  .715 &  .769 &   .949 &  .849 &  .227 &  .413 \\
             &             & Cut \& 500 &  .752 &  .715 &  .769 &   .948 &  .849 &  .227 &  .413 \\
             &             & Split \& 200 &  .748 &  .714 &  .767 &   .942 &  .846 &  .230 &  .415 \\
             &             & Split \& 500 &  .751 &  .714 &  .772 &   .939 &  .847 &  .234 &  .414 \\
             & Vanilla-DKT & - \& - &  .752 &  .722 &  .770 &   .946 &  .849 &  .229 &  .412 \\
             &             & Cut \& 200 &  .753 &  .721 &  .769 &   .948 &  .849 &  .227 &  .412 \\
             &             & Cut \& 500 &  .752 &  .721 &  .770 &   .947 &  .849 &  .228 &  .412 \\
             &             & Split \& 200 &  .749 &  .720 &  .767 &   .943 &  .846 &  .230 &  .414 \\
             &             & Split \& 500 &  .753 &  .721 &  .769 &   .949 &  .849 &  .228 &  .412 \\
ASSISTments 2017 & DKVMN & - \& - &  .684 &  .712 &  .626 &   .380 &  .472 &  .283 &  .450 \\
             &             & Cut \& 200 &  .666 &  .690 &  .612 &   .391 &  .477 &  .263 &  .460 \\
             &             & Cut \& 500 &  .676 &  .708 &  .627 &   .416 &  .500 &  .287 &  .455 \\
             &             & Split \& 200 &  .680 &  .704 &  .611 &   .377 &  .466 &  .270 &  .452 \\
             &             & Split \& 500 &  .680 &  .703 &  .615 &   .384 &  .471 &  .272 &  .453 \\
             & DKVMN-Paper & - \& - &  .681 &  .701 &  .630 &   .348 &  .448 &  .271 &  .453 \\
             &             & Cut \& 200 &  .667 &  .685 &  .638 &   .341 &  .444 &  .261 &  .461 \\
             &             & Cut \& 500 &  .673 &  .699 &  .642 &   .367 &  .467 &  .276 &  .457 \\
             &             & Split \& 200 &  .678 &  .696 &  .617 &   .342 &  .438 &  .257 &  .454 \\
             &             & Split \& 500 &  .676 &  .693 &  .624 &   .335 &  .433 &  .256 &  .456 \\
             & LSTM-DKT & - \& - &  .697 &  .734 &  .631 &   .450 &  .525 &  .321 &  .443 \\
             &             & Cut \& 200 &  .676 &  .707 &  .620 &   .435 &  .511 &  .289 &  .455 \\
             &             & Cut \& 500 &  .685 &  .724 &  .632 &   .461 &  .533 &  .312 &  .450 \\
             &             & Split \& 200 &  .692 &  .723 &  .630 &   .412 &  .498 &  .302 &  .446 \\
             &             & Split \& 500 &  .690 &  .722 &  .625 &   .433 &  .510 &  .303 &  .447 \\
             & LSTM-DKT-S+ & - \& - &  .697 &  .734 &  .630 &   .452 &  .526 &  .321 &  .444 \\
             &             & Cut \& 200 &  .676 &  .708 &  .622 &   .433 &  .511 &  .291 &  .455 \\
             &             & Cut \& 500 &  .686 &  .726 &  .633 &   .462 &  .534 &  .315 &  .449 \\
             &             & Split \& 200 &  .692 &  .723 &  .626 &   .420 &  .502 &  .302 &  .446 \\
             &             & Split \& 500 &  .690 &  .721 &  .626 &   .427 &  .507 &  .302 &  .447 \\
             & SAKT & - \& - &  .660 &  .650 &  .600 &   .261 &  .363 &  .207 &  .466 \\
             &             & Cut \& 200 &  .652 &  .659 &  .604 &   .312 &  .411 &  .221 &  .468 \\
             &             & Cut \& 500 &  .657 &  .664 &  .612 &   .327 &  .426 &  .234 &  .467 \\
             &             & Split \& 200 &  .672 &  .661 &  .617 &   .298 &  .397 &  .235 &  .462 \\
             &             & Split \& 500 &  .665 &  .651 &  .609 &   .287 &  .383 &  .219 &  .465 \\
             & Vanilla-DKT & - \& - &  .679 &  .698 &  .627 &   .342 &  .443 &  .266 &  .454 \\
             &             & Cut \& 200 &  .667 &  .687 &  .625 &   .365 &  .460 &  .261 &  .461 \\
             &             & Cut \& 500 &  .670 &  .692 &  .632 &   .367 &  .464 &  .269 &  .459 \\
             &             & Split \& 200 &  .681 &  .703 &  .615 &   .374 &  .464 &  .271 &  .453 \\
             &             & Split \& 500 &  .675 &  .691 &  .613 &   .350 &  .443 &  .254 &  .456 \\
IntroProg & DKVMN & - \& - &  .765 &  .847 &  .761 &   .754 &  .757 &  .529 &  .399 \\
             &             & Cut \& 200 &  .753 &  .833 &  .763 &   .763 &  .763 &  .504 &  .408 \\
             &             & Cut \& 500 &  .761 &  .843 &  .761 &   .756 &  .758 &  .522 &  .402 \\
             &             & Split \& 200 &  .756 &  .827 &  .757 &   .755 &  .756 &  .490 &  .406 \\
             &             & Split \& 500 &  .763 &  .842 &  .760 &   .755 &  .757 &  .521 &  .401 \\
             & DKVMN-Paper & - \& - &  .764 &  .845 &  .762 &   .749 &  .756 &  .527 &  .400 \\
             &             & Cut \& 200 &  .752 &  .832 &  .763 &   .759 &  .761 &  .502 &  .409 \\
             &             & Cut \& 500 &  .760 &  .842 &  .761 &   .752 &  .757 &  .520 &  .403 \\
             &             & Split \& 200 &  .754 &  .825 &  .752 &   .759 &  .755 &  .486 &  .407 \\
             IntroProg &  DKVMN-Paper  & Split \& 500 &  .761 &  .841 &  .762 &   .746 &  .754 &  .518 &  .402 \\
             & LSTM-DKT & - \& - &  .765 &  .846 &  .761 &   .755 &  .758 &  .530 &  .400 \\
             &             & Cut \& 200 &  .753 &  .833 &  .763 &   .764 &  .763 &  .504 &  .408 \\
             &             & Cut \& 500 &  .761 &  .843 &  .760 &   .757 &  .758 &  .522 &  .402 \\
             &             & Split \& 200 &  .757 &  .827 &  .754 &   .762 &  .758 &  .491 &  .406 \\
             &             & Split \& 500 &  .763 &  .842 &  .760 &   .756 &  .758 &  .521 &  .401 \\
             & LSTM-DKT-S+ & - \& - &  .764 &  .845 &  .760 &   .753 &  .756 &  .527 &  .400 \\
             &             & Cut \& 200 &  .751 &  .832 &  .759 &   .767 &  .763 &  .501 &  .409 \\
             &             & Cut \& 500 &  .759 &  .842 &  .758 &   .756 &  .757 &  .519 &  .403 \\
             &             & Split \& 200 &  .755 &  .826 &  .752 &   .761 &  .756 &  .487 &  .406 \\
             &             & Split \& 500 &  .761 &  .841 &  .757 &   .754 &  .756 &  .517 &  .402 \\
             & SAKT & - \& - &  .761 &  .843 &  .748 &   .768 &  .757 &  .521 &  .403 \\
             &             & Cut \& 200 &  .749 &  .830 &  .751 &   .778 &  .764 &  .497 &  .411 \\
             &             & Cut \& 500 &  .758 &  .839 &  .750 &   .767 &  .759 &  .516 &  .405 \\
             &             & Split \& 200 &  .754 &  .825 &  .743 &   .776 &  .759 &  .482 &  .407 \\
             &             & Split \& 500 &  .760 &  .839 &  .750 &   .767 &  .758 &  .515 &  .404 \\
             & Vanilla-DKT & - \& - &  .760 &  .840 &  .759 &   .744 &  .751 &  .519 &  .404 \\
             &             & Cut \& 200 &  .748 &  .827 &  .760 &   .756 &  .758 &  .495 &  .412 \\
             &             & Cut \& 500 &  .756 &  .836 &  .758 &   .746 &  .752 &  .511 &  .406 \\
             &             & Split \& 200 &  .752 &  .821 &  .756 &   .746 &  .751 &  .483 &  .410 \\
             &             & Split \& 500 &  .758 &  .835 &  .759 &   .742 &  .750 &  .511 &  .405 \\
Statics & DKVMN & - \& - &  .813 &  .836 &  .841 &   .931 &  .884 &  .422 &  .360 \\
             &             & Cut \& 200 &  .828 &  .830 &  .852 &   .950 &  .898 &  .379 &  .348 \\
             &             & Cut \& 500 &  .805 &  .823 &  .832 &   .933 &  .880 &  .394 &  .367 \\
             &             & Split \& 200 &  .807 &  .825 &  .835 &   .932 &  .881 &  .393 &  .364 \\
             &             & Split \& 500 &  .807 &  .828 &  .837 &   .927 &  .880 &  .407 &  .365 \\
             & DKVMN-Paper & - \& - &  .807 &  .823 &  .832 &   .938 &  .881 &  .393 &  .366 \\
             &             & Cut \& 200 &  .826 &  .827 &  .849 &   .951 &  .897 &  .368 &  .350 \\
             &             & Cut \& 500 &  .801 &  .812 &  .825 &   .938 &  .878 &  .369 &  .371 \\
             &             & Split \& 200 &  .803 &  .812 &  .831 &   .933 &  .879 &  .374 &  .369 \\
             &             & Split \& 500 &  .801 &  .813 &  .831 &   .928 &  .877 &  .384 &  .372 \\
             & LSTM-DKT & - \& - &  .805 &  .819 &  .824 &   .948 &  .881 &  .375 &  .367 \\
             &             & Cut \& 200 &  .822 &  .812 &  .840 &   .959 &  .896 &  .334 &  .355 \\
             &             & Cut \& 500 &  .798 &  .803 &  .817 &   .948 &  .878 &  .351 &  .374 \\
             &             & Split \& 200 &  .807 &  .824 &  .832 &   .937 &  .881 &  .383 &  .365 \\
             &             & Split \& 500 &  .799 &  .811 &  .822 &   .942 &  .877 &  .364 &  .373 \\
             & LSTM-DKT-S+ & - \& - &  .802 &  .807 &  .821 &   .947 &  .880 &  .362 &  .372 \\
             &             & Cut \& 200 &  .819 &  .804 &  .837 &   .961 &  .894 &  .315 &  .358 \\
             &             & Cut \& 500 &  .798 &  .806 &  .818 &   .946 &  .878 &  .353 &  .374 \\
             &             & Split \& 200 &  .806 &  .823 &  .832 &   .936 &  .881 &  .383 &  .365 \\
             &             & Split \& 500 &  .799 &  .813 &  .822 &   .941 &  .877 &  .364 &  .372 \\
             & SAKT & - \& - &  .802 &  .811 &  .826 &   .940 &  .879 &  .371 &  .370 \\
             &             & Cut \& 200 &  .828 &  .835 &  .853 &   .947 &  .898 &  .382 &  .348 \\
             &             & Cut \& 500 &  .798 &  .808 &  .824 &   .937 &  .876 &  .362 &  .373 \\
             &             & Split \& 200 &  .801 &  .808 &  .827 &   .936 &  .878 &  .363 &  .371 \\
             &             & Split \& 500 &  .800 &  .810 &  .829 &   .930 &  .877 &  .378 &  .372 \\
             & Vanilla-DKT & - \& - &  .805 &  .822 &  .829 &   .938 &  .880 &  .383 &  .367 \\
             &             & Cut \& 200 &  .817 &  .798 &  .835 &   .960 &  .893 &  .306 &  .360 \\
             &             & Cut \& 500 &  .794 &  .791 &  .815 &   .945 &  .875 &  .335 &  .379 \\
             &             & Split \& 200 &  .804 &  .815 &  .828 &   .938 &  .880 &  .372 &  .369 \\
             &             & Split \& 500 &  .800 &  .811 &  .824 &   .938 &  .877 &  .371 &  .373 \\
Synthetic-K2 & DKVMN & - \& - &  .806 &  .872 &  .852 &   .867 &  .860 &  .546 &  .365 \\
             &             & Cut \& 200 &  .806 &  .872 &  .852 &   .867 &  .860 &  .546 &  .365 \\
             &             & Cut \& 500 &  .806 &  .872 &  .852 &   .867 &  .860 &  .546 &  .365 \\
             &             & Split \& 200 &  .806 &  .872 &  .852 &   .867 &  .860 &  .546 &  .365 \\
             &             & Split \& 500 &  .806 &  .872 &  .852 &   .867 &  .860 &  .546 &  .365 \\
             & DKVMN-Paper & - \& - &  .806 &  .872 &  .853 &   .865 &  .859 &  .547 &  .365 \\
             &             & Cut \& 200 &  .806 &  .872 &  .853 &   .865 &  .859 &  .547 &  .365 \\
             &             & Cut \& 500 &  .806 &  .872 &  .853 &   .865 &  .859 &  .547 &  .365 \\
             &             & Split \& 200 &  .806 &  .872 &  .853 &   .865 &  .859 &  .547 &  .365 \\
             &             & Split \& 500 &  .806 &  .872 &  .853 &   .865 &  .859 &  .547 &  .365 \\
             & LSTM-DKT & - \& - &  .805 &  .871 &  .848 &   .872 &  .860 &  .541 &  .366 \\
             &             & Cut \& 200 &  .805 &  .871 &  .848 &   .872 &  .860 &  .541 &  .366 \\
             &             & Cut \& 500 &  .805 &  .871 &  .848 &   .872 &  .860 &  .541 &  .366 \\
             &             & Split \& 200 &  .805 &  .871 &  .848 &   .872 &  .860 &  .541 &  .366 \\
             &             & Split \& 500 &  .805 &  .871 &  .848 &   .872 &  .860 &  .541 &  .366 \\
             & LSTM-DKT-S+ & - \& - &  .806 &  .871 &  .851 &   .868 &  .859 &  .545 &  .366 \\
             &             & Cut \& 200 &  .806 &  .871 &  .851 &   .868 &  .859 &  .545 &  .366 \\
             &             & Cut \& 500 &  .806 &  .871 &  .851 &   .868 &  .859 &  .545 &  .366 \\
             &             & Split \& 200 &  .806 &  .871 &  .851 &   .868 &  .859 &  .545 &  .366 \\
             &             & Split \& 500 &  .806 &  .871 &  .851 &   .868 &  .859 &  .545 &  .366 \\
             & SAKT & - \& - &  .806 &  .872 &  .855 &   .862 &  .859 &  .548 &  .365 \\
             &             & Cut \& 200 &  .806 &  .872 &  .855 &   .862 &  .859 &  .548 &  .365 \\
             &             & Cut \& 500 &  .806 &  .872 &  .855 &   .862 &  .859 &  .548 &  .365 \\
             &             & Split \& 200 &  .806 &  .872 &  .855 &   .862 &  .859 &  .548 &  .365 \\
             &             & Split \& 500 &  .806 &  .872 &  .855 &   .862 &  .859 &  .548 &  .365 \\
             & Vanilla-DKT & - \& - &  .805 &  .869 &  .850 &   .867 &  .859 &  .542 &  .367 \\
             &             & Cut \& 200 &  .805 &  .869 &  .850 &   .867 &  .859 &  .542 &  .367 \\
             &             & Cut \& 500 &  .805 &  .869 &  .850 &   .867 &  .859 &  .542 &  .367 \\
             &             & Split \& 200 &  .804 &  .869 &  .851 &   .867 &  .858 &  .542 &  .367 \\
             &             & Split \& 500 &  .805 &  .869 &  .850 &   .867 &  .859 &  .542 &  .367 \\
Synthetic-K5 & DKVMN & - \& - &  .754 &  .829 &  .812 &   .775 &  .793 &  .491 &  .404 \\
             &             & Cut \& 200 &  .754 &  .829 &  .812 &   .775 &  .793 &  .491 &  .404 \\
             &             & Cut \& 500 &  .754 &  .829 &  .812 &   .775 &  .793 &  .491 &  .404 \\
             &             & Split \& 200 &  .754 &  .829 &  .812 &   .775 &  .793 &  .491 &  .404 \\
             &             & Split \& 500 &  .754 &  .829 &  .812 &   .775 &  .793 &  .491 &  .404 \\
             & DKVMN-Paper & - \& - &  .753 &  .829 &  .812 &   .773 &  .792 &  .490 &  .405 \\
             &             & Cut \& 200 &  .753 &  .829 &  .812 &   .773 &  .792 &  .490 &  .405 \\
             &             & Cut \& 500 &  .753 &  .829 &  .812 &   .773 &  .792 &  .490 &  .405 \\
             &             & Split \& 200 &  .753 &  .829 &  .812 &   .773 &  .792 &  .490 &  .405 \\
             &             & Split \& 500 &  .753 &  .829 &  .812 &   .773 &  .792 &  .490 &  .405 \\
             & LSTM-DKT & - \& - &  .752 &  .827 &  .807 &   .779 &  .793 &  .485 &  .406 \\
             &             & Cut \& 200 &  .752 &  .827 &  .807 &   .779 &  .793 &  .485 &  .406 \\
             &             & Cut \& 500 &  .752 &  .827 &  .807 &   .779 &  .793 &  .485 &  .406 \\
             &             & Split \& 200 &  .752 &  .827 &  .807 &   .779 &  .793 &  .485 &  .406 \\
             &             & Split \& 500 &  .752 &  .827 &  .807 &   .779 &  .793 &  .485 &  .406 \\
             & LSTM-DKT-S+ & - \& - &  .751 &  .826 &  .805 &   .779 &  .792 &  .483 &  .406 \\
             &             & Cut \& 200 &  .751 &  .826 &  .805 &   .779 &  .792 &  .483 &  .406 \\
             &             & Cut \& 500 &  .751 &  .826 &  .805 &   .779 &  .792 &  .483 &  .406 \\
             &             & Split \& 200 &  .751 &  .826 &  .805 &   .779 &  .792 &  .483 &  .406 \\
             &             & Split \& 500 &  .751 &  .826 &  .805 &   .779 &  .792 &  .483 &  .406 \\
             & SAKT & - \& - &  .753 &  .828 &  .808 &   .780 &  .794 &  .488 &  .405 \\
             Synthetic-K5 & SAKT & Cut \& 200 &  .753 &  .828 &  .808 &   .780 &  .794 &  .488 &  .405 \\
             &             & Cut \& 500 &  .753 &  .828 &  .808 &   .780 &  .794 &  .488 &  .405 \\
             &             & Split \& 200 &  .753 &  .828 &  .809 &   .777 &  .793 &  .488 &  .405 \\
             &             & Split \& 500 &  .753 &  .828 &  .808 &   .780 &  .794 &  .488 &  .405 \\
             & Vanilla-DKT & - \& - &  .749 &  .822 &  .803 &   .777 &  .790 &  .478 &  .409 \\
             &             & Cut \& 200 &  .749 &  .822 &  .803 &   .777 &  .790 &  .478 &  .409 \\
             &             & Cut \& 500 &  .749 &  .822 &  .803 &   .777 &  .790 &  .478 &  .409 \\
             &             & Split \& 200 &  .750 &  .822 &  .805 &   .778 &  .791 &  .481 &  .409 \\
             &             & Split \& 500 &  .749 &  .822 &  .803 &   .777 &  .790 &  .478 &  .409 \\
\bottomrule
\end{longtable}

\begin{table}[H]
\small\centering
\caption{Student attempt split effect averaged over datasets}
\label{tbl:split-vs-cut-avg}
\begin{tabular}{ll|lllllll}
\toprule
Model & Max attempt filter &   Acc &   AUC &  Prec & Recall &    F1 &   MCC &  RMSE \\
\midrule
DKVMN & - \& - &  .762 &  .805 &  .775 &   .792 &  .777 &  .424 &  .398 \\
            & Cut \& 200 &  .759 &  .798 &  .774 &   .796 &  .779 &  .410 &  .400 \\
            & Cut \& 500 &  .759 &  .802 &  .774 &   .797 &  .780 &  .419 &  .401 \\
            & Split \& 200 &  .759 &  .798 &  .772 &   .792 &  .775 &  .411 &  .401 \\
            & Split \& 500 &  .760 &  .801 &  .773 &   .792 &  .776 &  .418 &  .400 \\
DKVMN-Paper & - \& - &  .758 &  .797 &  .773 &   .787 &  .771 &  .411 &  .402 \\
            & Cut \& 200 &  .758 &  .795 &  .777 &   .788 &  .773 &  .406 &  .402 \\
            & Cut \& 500 &  .757 &  .797 &  .775 &   .790 &  .774 &  .412 &  .403 \\
            & Split \& 200 &  .757 &  .794 &  .771 &   .788 &  .771 &  .405 &  .403 \\
            & Split \& 500 &  .758 &  .796 &  .773 &   .783 &  .769 &  .411 &  .403 \\
LSTM-DKT & - \& - &  .763 &  .806 &  .773 &   .806 &  .784 &  .422 &  .399 \\
            & Cut \& 200 &  .760 &  .799 &  .773 &   .805 &  .784 &  .407 &  .401 \\
            & Cut \& 500 &  .760 &  .802 &  .772 &   .807 &  .785 &  .416 &  .401 \\
            & Split \& 200 &  .761 &  .802 &  .772 &   .800 &  .781 &  .414 &  .400 \\
            & Split \& 500 &  .760 &  .802 &  .771 &   .802 &  .781 &  .416 &  .401 \\
LSTM-DKT-S+ & - \& - &  .762 &  .804 &  .772 &   .805 &  .784 &  .420 &  .399 \\
            & Cut \& 200 &  .759 &  .797 &  .772 &   .805 &  .783 &  .404 &  .401 \\
            & Cut \& 500 &  .759 &  .802 &  .772 &   .807 &  .785 &  .416 &  .401 \\
            & Split \& 200 &  .760 &  .801 &  .771 &   .801 &  .781 &  .413 &  .400 \\
            & Split \& 500 &  .760 &  .802 &  .771 &   .800 &  .780 &  .415 &  .401 \\
SAKT & - \& - &  .754 &  .787 &  .766 &   .775 &  .758 &  .397 &  .405 \\
            & Cut \& 200 &  .755 &  .791 &  .770 &   .787 &  .769 &  .399 &  .403 \\
            & Cut \& 500 &  .753 &  .788 &  .768 &   .785 &  .767 &  .400 &  .406 \\
            & Split \& 200 &  .755 &  .787 &  .768 &   .784 &  .765 &  .397 &  .405 \\
            & Split \& 500 &  .754 &  .787 &  .768 &   .778 &  .761 &  .401 &  .405 \\
Vanilla-DKT & - \& - &  .758 &  .798 &  .771 &   .787 &  .771 &  .410 &  .402 \\
            & Cut \& 200 &  .756 &  .790 &  .771 &   .793 &  .775 &  .394 &  .404 \\
            & Cut \& 500 &  .755 &  .791 &  .770 &   .792 &  .773 &  .402 &  .405 \\
            & Split \& 200 &  .757 &  .794 &  .770 &   .791 &  .773 &  .404 &  .404 \\
            & Split \& 500 &  .756 &  .794 &  .768 &   .786 &  .769 &  .404 &  .404 \\
\bottomrule
\end{tabular}
\end{table}

\end{footnotesize}

\newpage

\section{Hardware comparison CPU versus GPU}


Both the CPU and GPU results here were obtained with a later 
TensorFlow version than the hyperparameter tuning results, wherefore 
the CPU results may not exactly match the results in other result tables.
Also the CPU hardware may differ from the other results as the models
were trained and evaluated in a computation cluster with multiple 
different CPU nodes with varying CPU types.

\label{appendix:cpu-vs-gpu}

\begin{footnotesize}
\begin{longtable}{lll|lllllll}
\label{tbl:cpu-vs-gpu}\\
\caption{Hardware result comparison}\\
\toprule
Dataset & Model & Hardware & Acc &   AUC &  Prec & Recall &    F1 &   MCC &  RMSE \\
\midrule
\endhead
\multirow{2}{8em}{ASSISTments 2009 Updated} 
& DKVMN & CPU-tf2.1.0 &  .758 &  .809 &  .765 &   .895 &  .825 &  .450 &  .403 \\
             &             & CPU-tf2.6.2 &  .757 &  .808 &  .765 &   .894 &  .825 &  .448 &  .404 \\
             &             & GPU &  .757 &  .808 &  .765 &   .894 &  .825 &  .448 &  .404 \\
             & DKVMN-Paper & CPU-tf2.1.0 &  .755 &  .806 &  .760 &   .902 &  .825 &  .443 &  .405 \\
             &             & CPU-tf2.6.2 &  .755 &  .806 &  .763 &   .893 &  .823 &  .442 &  .405 \\
             &             & GPU &  .754 &  .805 &  .762 &   .894 &  .823 &  .440 &  .405 \\
             & LSTM-DKT & CPU-tf2.1.0 &  .761 &  .814 &  .767 &   .898 &  .827 &  .455 &  .400 \\
             &             & CPU-tf2.6.2 &  .759 &  .811 &  .765 &   .898 &  .826 &  .451 &  .402 \\
             &             & GPU &  .759 &  .811 &  .765 &   .897 &  .826 &  .451 &  .402 \\
             & LSTM-DKT-S+ & CPU-tf2.1.0 &  .761 &  .814 &  .765 &   .901 &  .827 &  .455 &  .400 \\
             &             & CPU-tf2.6.2 &  .759 &  .810 &  .765 &   .898 &  .826 &  .451 &  .402 \\
             &             & GPU &  .759 &  .810 &  .765 &   .899 &  .826 &  .451 &  .402 \\
             & SAKT & CPU-tf2.1.0 &  .752 &  .798 &  .759 &   .895 &  .821 &  .434 &  .408 \\
             &             & CPU-tf2.6.2 &  .752 &  .799 &  .755 &   .904 &  .823 &  .433 &  .408 \\
             &             & GPU &  .752 &  .799 &  .756 &   .902 &  .823 &  .434 &  .408 \\
             & Vanilla-DKT & CPU-tf2.1.0 &  .757 &  .809 &  .766 &   .892 &  .824 &  .448 &  .403 \\
             &             & CPU-tf2.6.2 &  .758 &  .810 &  .768 &   .889 &  .824 &  .450 &  .403 \\
             &             & GPU &  .758 &  .811 &  .766 &   .894 &  .825 &  .449 &  .402 \\
ASSISTments 2015 & DKVMN & CPU-tf2.1.0 &  .750 &  .723 &  .769 &   .941 &  .846 &  .239 &  .413 \\
             &             & CPU-tf2.6.2 &  .750 &  .723 &  .769 &   .941 &  .846 &  .237 &  .413 \\
             &             & GPU &  .750 &  .722 &  .769 &   .941 &  .846 &  .239 &  .413 \\
             & DKVMN-Paper & CPU-tf2.1.0 &  .750 &  .718 &  .769 &   .943 &  .847 &  .237 &  .414 \\
             &             & CPU-tf2.6.2 &  .750 &  .722 &  .769 &   .941 &  .846 &  .238 &  .413 \\
             &             & GPU &  .750 &  .721 &  .769 &   .942 &  .846 &  .235 &  .414 \\
             & LSTM-DKT & CPU-tf2.1.0 &  .751 &  .725 &  .769 &   .943 &  .847 &  .239 &  .413 \\
             &             & CPU-tf2.6.2 &  .750 &  .724 &  .769 &   .943 &  .847 &  .237 &  .413 \\
             &             & GPU &  .750 &  .724 &  .769 &   .943 &  .847 &  .238 &  .413 \\
             & LSTM-DKT-S+ & CPU-tf2.1.0 &  .751 &  .725 &  .769 &   .943 &  .847 &  .238 &  .412 \\
             &             & CPU-tf2.6.2 &  .751 &  .724 &  .768 &   .944 &  .847 &  .236 &  .413 \\
             &             & GPU &  .750 &  .723 &  .770 &   .941 &  .846 &  .239 &  .413 \\
             & SAKT & CPU-tf2.1.0 &  .748 &  .714 &  .767 &   .942 &  .846 &  .230 &  .415 \\
             &             & CPU-tf2.6.2 &  .749 &  .714 &  .767 &   .944 &  .846 &  .230 &  .415 \\
             &             & GPU &  .749 &  .714 &  .767 &   .943 &  .846 &  .230 &  .416 \\
             & Vanilla-DKT & CPU-tf2.1.0 &  .749 &  .720 &  .767 &   .943 &  .846 &  .230 &  .414 \\
             &             & CPU-tf2.6.2 &  .749 &  .721 &  .770 &   .938 &  .845 &  .236 &  .414 \\
             &             & GPU &  .749 &  .721 &  .769 &   .939 &  .846 &  .236 &  .414 \\
ASSISTments 2017 & DKVMN & CPU-tf2.1.0 &  .680 &  .704 &  .611 &   .377 &  .466 &  .270 &  .452 \\
             &             & CPU-tf2.6.2 &  .680 &  .702 &  .614 &   .367 &  .459 &  .268 &  .453 \\
             &             & GPU &  .680 &  .702 &  .612 &   .372 &  .462 &  .268 &  .453 \\
             & DKVMN-Paper & CPU-tf2.1.0 &  .678 &  .696 &  .617 &   .342 &  .438 &  .257 &  .454 \\
             &             & CPU-tf2.6.2 &  .678 &  .695 &  .616 &   .344 &  .438 &  .257 &  .454 \\
             &             & GPU &  .678 &  .695 &  .616 &   .343 &  .438 &  .256 &  .454 \\
             ASSISTments 2017 & LSTM-DKT & CPU-tf2.1.0 &  .692 &  .723 &  .630 &   .412 &  .498 &  .302 &  .446 \\
             &             & CPU-tf2.6.2 &  .689 &  .719 &  .620 &   .418 &  .499 &  .297 &  .448 \\
             &             & GPU &  .689 &  .719 &  .619 &   .417 &  .497 &  .295 &  .448 \\
             & LSTM-DKT-S+ & CPU-tf2.1.0 &  .692 &  .723 &  .626 &   .420 &  .502 &  .302 &  .446 \\
             &             & CPU-tf2.6.2 &  .690 &  .719 &  .619 &   .422 &  .502 &  .298 &  .448 \\
             &             & GPU &  .689 &  .719 &  .621 &   .414 &  .496 &  .296 &  .448 \\
             & SAKT & CPU-tf2.1.0 &  .672 &  .661 &  .617 &   .298 &  .397 &  .235 &  .462 \\
             &             & CPU-tf2.6.2 &  .669 &  .655 &  .609 &   .292 &  .391 &  .227 &  .463 \\
             &             & GPU &  .669 &  .656 &  .610 &   .294 &  .393 &  .229 &  .463 \\
             & Vanilla-DKT & CPU-tf2.1.0 &  .681 &  .703 &  .615 &   .374 &  .464 &  .271 &  .453 \\
             &             & CPU-tf2.6.2 &  .683 &  .708 &  .612 &   .398 &  .481 &  .279 &  .451 \\
             &             & GPU &  .683 &  .709 &  .613 &   .397 &  .480 &  .280 &  .451 \\
IntroProg & DKVMN & CPU-tf2.1.0 &  .756 &  .827 &  .757 &   .755 &  .756 &  .490 &  .406 \\
             &             & CPU-tf2.6.2 &  .756 &  .826 &  .757 &   .753 &  .755 &  .490 &  .406 \\
             &             & GPU &  .756 &  .826 &  .757 &   .753 &  .755 &  .490 &  .406 \\
             & DKVMN-Paper & CPU-tf2.1.0 &  .754 &  .825 &  .752 &   .759 &  .755 &  .486 &  .407 \\
             &             & CPU-tf2.6.2 &  .755 &  .824 &  .757 &   .753 &  .754 &  .487 &  .407 \\
             &             & GPU &  .755 &  .824 &  .756 &   .755 &  .755 &  .487 &  .407 \\
             & LSTM-DKT & CPU-tf2.1.0 &  .757 &  .827 &  .754 &   .762 &  .758 &  .491 &  .406 \\
             &             & CPU-tf2.6.2 &  .755 &  .824 &  .751 &   .763 &  .757 &  .488 &  .407 \\
             &             & GPU &  .755 &  .824 &  .752 &   .761 &  .756 &  .488 &  .407 \\
             & LSTM-DKT-S+ & CPU-tf2.1.0 &  .755 &  .826 &  .752 &   .761 &  .756 &  .487 &  .406 \\
             &             & CPU-tf2.6.2 &  .753 &  .823 &  .753 &   .752 &  .752 &  .484 &  .408 \\
             &             & GPU &  .754 &  .824 &  .753 &   .755 &  .754 &  .485 &  .408 \\
             & SAKT & CPU-tf2.1.0 &  .754 &  .825 &  .743 &   .776 &  .759 &  .482 &  .407 \\
             &             & CPU-tf2.6.2 &  .754 &  .824 &  .755 &   .753 &  .753 &  .483 &  .408 \\
             &             & GPU &  .754 &  .824 &  .754 &   .753 &  .753 &  .483 &  .407 \\
             & Vanilla-DKT & CPU-tf2.1.0 &  .752 &  .821 &  .756 &   .746 &  .751 &  .483 &  .410 \\
             &             & CPU-tf2.6.2 &  .754 &  .823 &  .749 &   .764 &  .756 &  .485 &  .408 \\
             &             & GPU &  .753 &  .822 &  .746 &   .766 &  .756 &  .483 &  .409 \\
Statics & DKVMN & CPU-tf2.1.0 &  .807 &  .825 &  .835 &   .932 &  .881 &  .393 &  .364 \\
             &             & CPU-tf2.6.2 &  .807 &  .823 &  .837 &   .929 &  .880 &  .394 &  .365 \\
             &             & GPU &  .807 &  .823 &  .837 &   .929 &  .880 &  .394 &  .365 \\
             & DKVMN-Paper & CPU-tf2.1.0 &  .803 &  .812 &  .831 &   .933 &  .879 &  .374 &  .369 \\
             &             & CPU-tf2.6.2 &  .802 &  .809 &  .832 &   .930 &  .878 &  .373 &  .371 \\
             &             & GPU &  .802 &  .809 &  .832 &   .930 &  .878 &  .373 &  .371 \\
             & LSTM-DKT & CPU-tf2.1.0 &  .807 &  .824 &  .832 &   .937 &  .881 &  .383 &  .365 \\
             &             & CPU-tf2.6.2 &  .806 &  .823 &  .832 &   .936 &  .881 &  .382 &  .366 \\
             &             & GPU &  .806 &  .821 &  .832 &   .934 &  .880 &  .382 &  .366 \\
             & LSTM-DKT-S+ & CPU-tf2.1.0 &  .806 &  .823 &  .832 &   .936 &  .881 &  .383 &  .365 \\
             &             & CPU-tf2.6.2 &  .806 &  .822 &  .833 &   .934 &  .881 &  .384 &  .366 \\
             &             & GPU &  .807 &  .822 &  .835 &   .932 &  .881 &  .390 &  .366 \\
             & SAKT & CPU-tf2.1.0 &  .801 &  .808 &  .827 &   .936 &  .878 &  .363 &  .371 \\
             &             & CPU-tf2.6.2 &  .801 &  .808 &  .828 &   .935 &  .878 &  .364 &  .371 \\
             &             & GPU &  .801 &  .811 &  .830 &   .930 &  .878 &  .370 &  .371 \\
             & Vanilla-DKT & CPU-tf2.1.0 &  .804 &  .815 &  .828 &   .938 &  .880 &  .372 &  .369 \\
             &             & CPU-tf2.6.2 &  .804 &  .818 &  .830 &   .936 &  .880 &  .378 &  .368 \\
             &             & GPU &  .805 &  .819 &  .832 &   .934 &  .880 &  .380 &  .367 \\
Synthetic-K2 & DKVMN & CPU-tf2.1.0 &  .806 &  .872 &  .852 &   .867 &  .860 &  .546 &  .365 \\
             &             & CPU-tf2.6.2 &  .806 &  .872 &  .852 &   .868 &  .860 &  .546 &  .365 \\
             &             & GPU &  .806 &  .872 &  .852 &   .868 &  .860 &  .546 &  .365 \\
             Synthetic-K2 & DKVMN-Paper & CPU-tf2.1.0 &  .806 &  .872 &  .853 &   .865 &  .859 &  .547 &  .365 \\
             &             & CPU-tf2.6.2 &  .806 &  .872 &  .852 &   .867 &  .859 &  .546 &  .365 \\
             &             & GPU &  .806 &  .872 &  .852 &   .867 &  .859 &  .546 &  .365 \\
             & LSTM-DKT & CPU-tf2.1.0 &  .805 &  .871 &  .848 &   .872 &  .860 &  .541 &  .366 \\
             &             & CPU-tf2.6.2 &  .805 &  .871 &  .853 &   .865 &  .859 &  .545 &  .366 \\
             &             & GPU &  .805 &  .870 &  .852 &   .865 &  .859 &  .544 &  .366 \\
             & LSTM-DKT-S+ & CPU-tf2.1.0 &  .806 &  .871 &  .851 &   .868 &  .859 &  .545 &  .366 \\
             &             & CPU-tf2.6.2 &  .806 &  .871 &  .853 &   .865 &  .859 &  .546 &  .365 \\
             &             & GPU &  .806 &  .871 &  .853 &   .866 &  .859 &  .547 &  .365 \\
             & SAKT & CPU-tf2.1.0 &  .806 &  .872 &  .855 &   .862 &  .859 &  .548 &  .365 \\
             &             & CPU-tf2.6.2 &  .806 &  .872 &  .857 &   .860 &  .858 &  .549 &  .366 \\
             &             & GPU &  .806 &  .872 &  .857 &   .860 &  .858 &  .549 &  .366 \\
             & Vanilla-DKT & CPU-tf2.1.0 &  .804 &  .869 &  .851 &   .867 &  .858 &  .542 &  .367 \\
             &             & CPU-tf2.6.2 &  .804 &  .868 &  .854 &   .860 &  .857 &  .543 &  .368 \\
             &             & GPU &  .803 &  .868 &  .856 &   .857 &  .857 &  .544 &  .367 \\
Synthetic-K5 & DKVMN & CPU-tf2.1.0 &  .754 &  .829 &  .812 &   .775 &  .793 &  .491 &  .404 \\
             &             & CPU-tf2.6.2 &  .754 &  .829 &  .812 &   .775 &  .793 &  .491 &  .404 \\
             &             & GPU &  .754 &  .829 &  .812 &   .775 &  .793 &  .491 &  .404 \\
             & DKVMN-Paper & CPU-tf2.1.0 &  .753 &  .829 &  .812 &   .773 &  .792 &  .490 &  .405 \\
             &             & CPU-tf2.6.2 &  .753 &  .829 &  .813 &   .771 &  .791 &  .490 &  .405 \\
             &             & GPU &  .753 &  .829 &  .813 &   .771 &  .791 &  .490 &  .405 \\
             & LSTM-DKT & CPU-tf2.1.0 &  .752 &  .827 &  .807 &   .779 &  .793 &  .485 &  .406 \\
             &             & CPU-tf2.6.2 &  .751 &  .825 &  .803 &   .782 &  .792 &  .482 &  .407 \\
             &             & GPU &  .751 &  .825 &  .804 &   .780 &  .792 &  .483 &  .407 \\
             & LSTM-DKT-S+ & CPU-tf2.1.0 &  .751 &  .826 &  .805 &   .779 &  .792 &  .483 &  .406 \\
             &             & CPU-tf2.6.2 &  .751 &  .825 &  .804 &   .780 &  .792 &  .482 &  .407 \\
             &             & GPU &  .751 &  .825 &  .804 &   .780 &  .792 &  .482 &  .407 \\
             & SAKT & CPU-tf2.1.0 &  .753 &  .828 &  .809 &   .777 &  .793 &  .488 &  .405 \\
             &             & CPU-tf2.6.2 &  .752 &  .828 &  .817 &   .764 &  .789 &  .492 &  .406 \\
             &             & GPU &  .753 &  .828 &  .807 &   .779 &  .793 &  .486 &  .405 \\
             & Vanilla-DKT & CPU-tf2.1.0 &  .750 &  .822 &  .805 &   .778 &  .791 &  .481 &  .409 \\
             &             & CPU-tf2.6.2 &  .748 &  .822 &  .804 &   .774 &  .789 &  .478 &  .409 \\
             &             & GPU &  .749 &  .821 &  .804 &   .777 &  .790 &  .479 &  .409 \\
\bottomrule
\end{longtable}
\end{footnotesize}

\end{document}